\documentclass[runningheads]{llncs}

\usepackage{eccv}

\usepackage{eccvabbrv}

\usepackage{float}
\usepackage{algorithm}
\usepackage{algorithmic}
\usepackage{amsmath}
\usepackage{amssymb}
\usepackage{booktabs}
\usepackage{caption}
\usepackage{colortbl}
\usepackage{enumitem}
\usepackage{graphicx}
\usepackage{mathtools}
\usepackage{microtype}
\usepackage{multirow}
\usepackage{pifont}
\usepackage{placeins}
\usepackage{tabularx}
\usepackage{wrapfig}

\usepackage[accsupp]{axessibility}

\definecolor{BLACK}{RGB}{0,0,0}
\definecolor{BLUE}{RGB}{0,114,178}
\definecolor{GRAY}{RGB}{99,99,99}
\definecolor{GREEN}{RGB}{0,158,115}
\definecolor{ORANGE}{RGB}{230,159,0}
\definecolor{PURPLE}{RGB}{204,121,167}
\definecolor{VERMILION}{RGB}{213,94,0}

\newcommand{\gM}{{\mathcal{M}}}
\newcommand{\gN}{{\mathcal{N}}}
\newcommand{\gT}{{\mathcal{T}}}
\newcommand{\R}{\mathbb{R}}
\newcommand{\dee}{{\mathrm{d}}}
\newcommand{\fst}[1]{$\mathbf{#1}$}
\newcommand{\snd}[1]{\underline{#1}}
\newcommand{\sss}[1]{{\scriptscriptstyle (#1)}}
\newcommand{\sigone}{\makebox[0pt][l]{\smash{\textsuperscript{\tiny\!*}}}}
\newcommand{\sigtwo}{\makebox[0pt][l]{\smash{\textsuperscript{\tiny\!*\!*}}}}

\usepackage{hyperref}
\usepackage{orcidlink}

\begin{document}

\title{Be Tangential to Manifold: Discovering Riemannian Metric for Diffusion Models}

\titlerunning{Discovering Riemannian Metric for Diffusion Models}

\author{
    Shinnosuke Saito\inst{1}\orcidlink{0009-0001-1440-5485}
    \and
    Takashi Matsubara\inst{2,1}\orcidlink{0000-0003-0642-4800}
}

\authorrunning{S.~Saito and T.~Matsubara}

\institute{
    \begin{tabular}{@{}c@{\quad}c@{}}
        \inst{1}Hokkaido University & \inst{2}AI Lab, CyberAgent, Inc.\\
    \end{tabular}
    \email{saitou.shinnosuke.y0@elms.hokudai.ac.jp}
    \email{matsubara\_takashi@cyberagent.co.jp}
}

\maketitle

\begin{abstract}
    Diffusion models are powerful deep generative models, but unlike classical models, they lack an explicit low-dimensional latent space that parameterizes the data manifold.
    This absence makes it difficult to perform manifold-aware operations, such as geometrically faithful interpolation or conditional guidance that respects the learned manifold.
    We propose a training-free Riemannian metric on the noise space, derived from the Jacobian of the score function.
    The key insight is that the spectral structure of this Jacobian separates tangent and normal directions of the data manifold; our metric leverages this separation to encourage paths to stay tangential to the manifold rather than drift toward high-density regions.
    To validate that our metric faithfully captures the manifold geometry, we examine it from two complementary angles.
    First, geodesics under our metric yield perceptually more natural interpolations than existing methods on synthetic, image, and video frame datasets.
    Second, the tangent--normal decomposition induced by our metric prevents classifier-free guidance from deviating off the manifold, improving generation quality while preserving text-image alignment.
\end{abstract}

\section{Introduction}
\label{sec:introduction}
Diffusion models are a class of deep generative models (DGMs) that have shown a remarkable capability to generate high-fidelity, diverse content \cite{Ho2020,Song2021a,Rombach2022}.
A key theoretical lens for understanding and improving DGMs is the \emph{manifold hypothesis}, which states that real-world data (e.g., images) are concentrated around a low-dimensional manifold embedded in the high-dimensional data space \cite{Bengio2012,Fefferman2016}.
Under this hypothesis, DGMs are understood to learn not only the data distribution but also its underlying manifold, either explicitly or implicitly \cite{Loaiza-ganem2024}.

Among DGMs, variational autoencoders (VAEs) \cite{Kingma2014} and generative adversarial networks (GANs) \cite{Goodfellow2014} possess a low-dimensional latent space, which serves as an explicit parameterization of the data manifold \cite{Arjovsky2017}.
On this latent space, one can define a Riemannian metric by pulling back the data-space metric through the decoder~\cite{Gruffaz2025}, enabling geometrically meaningful operations.
For example, traversing the latent space along geodesics yields interpolations that are faithful to the intrinsic geometric structure of the data \cite{Shao2017,Arvanitidis2018,Chen2018,Arvanitidis2021}.
However, diffusion models lack such a low-dimensional latent space; their noise space has the same dimensionality as the data space, making it nontrivial to define a meaningful pullback metric.

\begin{figure}[t]
    \centering
    \includegraphics[width=0.315\textwidth]{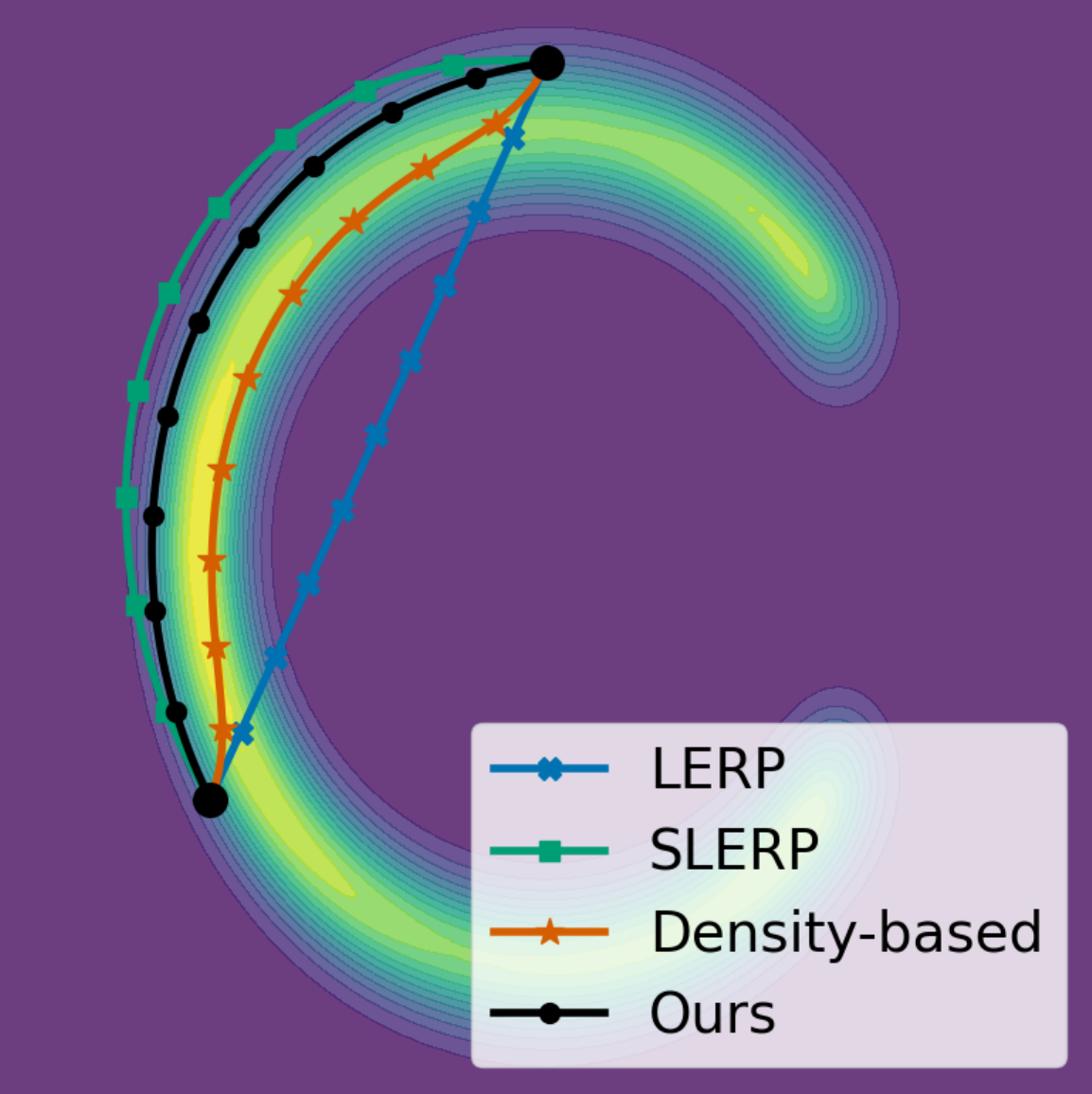}
    \includegraphics[width=0.4\textwidth]{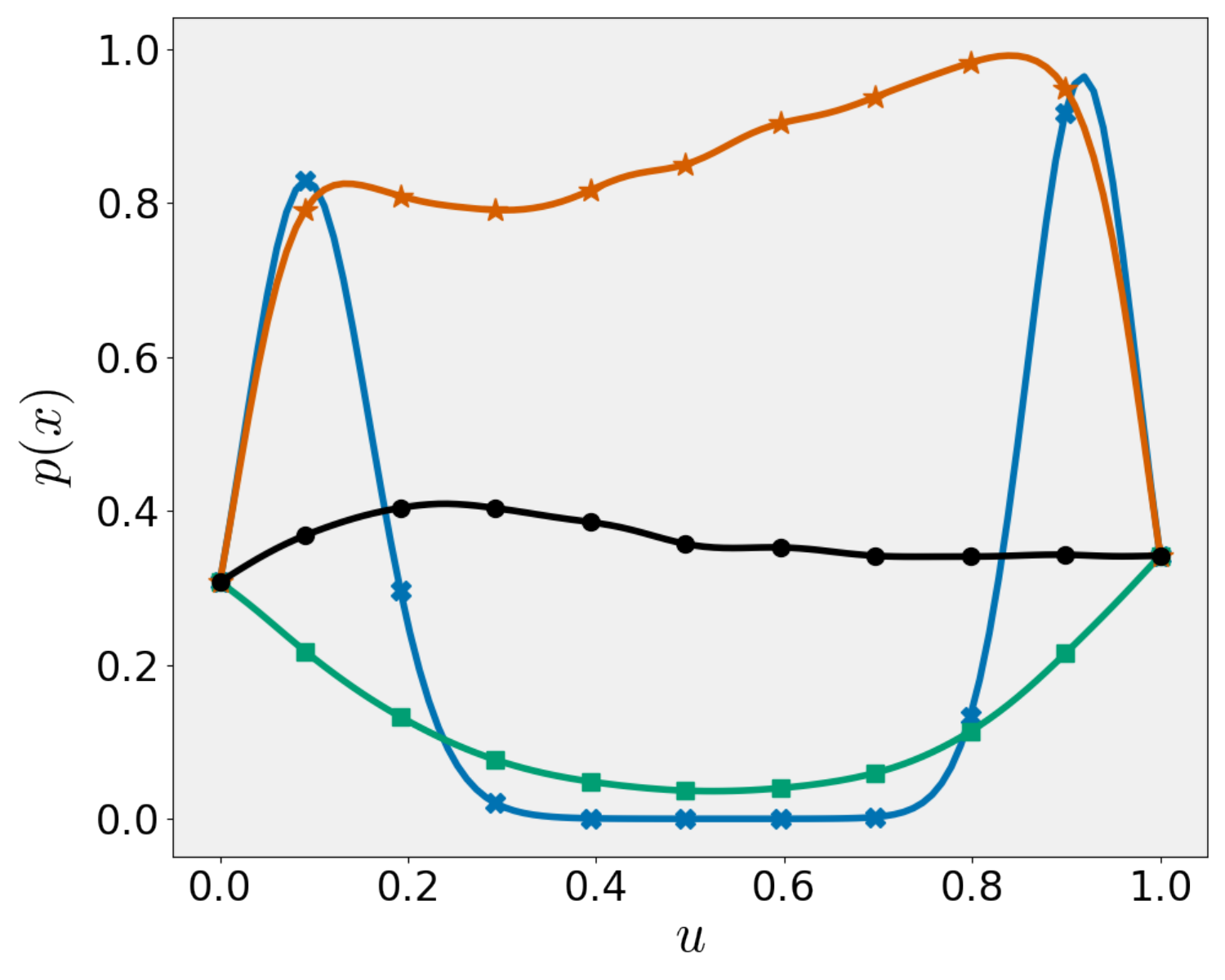}
    \caption{
        \textbf{A prominent example.}
        (\textit{left}) Interpolation paths on a C-shaped distribution.
        (\textit{right}) A plot of the probability density transitions for interpolation paths.
        \textcolor{BLUE}{LERP} cuts through a low-density region.
        \textcolor{GREEN}{SLERP} deviates from the manifold.
        \textcolor{VERMILION}{Density-based} interpolation approaches a high-density region and traverses it, not preserving the probabilities of the endpoints.
        \textbf{Ours} runs parallel to the manifold, yielding natural transitions with preserved endpoint probabilities.
    }
    \label{fig:illust}
\end{figure}

Some previous studies have attempted to define a Riemannian metric for diffusion models based on the density of noisy samples or the direction to the manifold (see \cref{fig:illust}) \cite{Yu2025,Azeglio2025}.
Geodesics under these metrics, however, may overly approach the center of the data distribution, resulting in over-smoothed images~\cite{Karczewski2025a}.
The fundamental issue is that density or direction only tell us \emph{where} samples concentrate, but not \emph{how} the manifold is oriented in the ambient space.
This raises a natural question: \emph{can we define a Riemannian metric for diffusion models that directly reflects the intrinsic geometry of the data manifold, rather than its density?}

\begin{wrapfigure}[14]{r}{0.45\textwidth}
    \vspace{-3pt}
    \includegraphics[width=0.447\textwidth,bb=0 50 900 450]{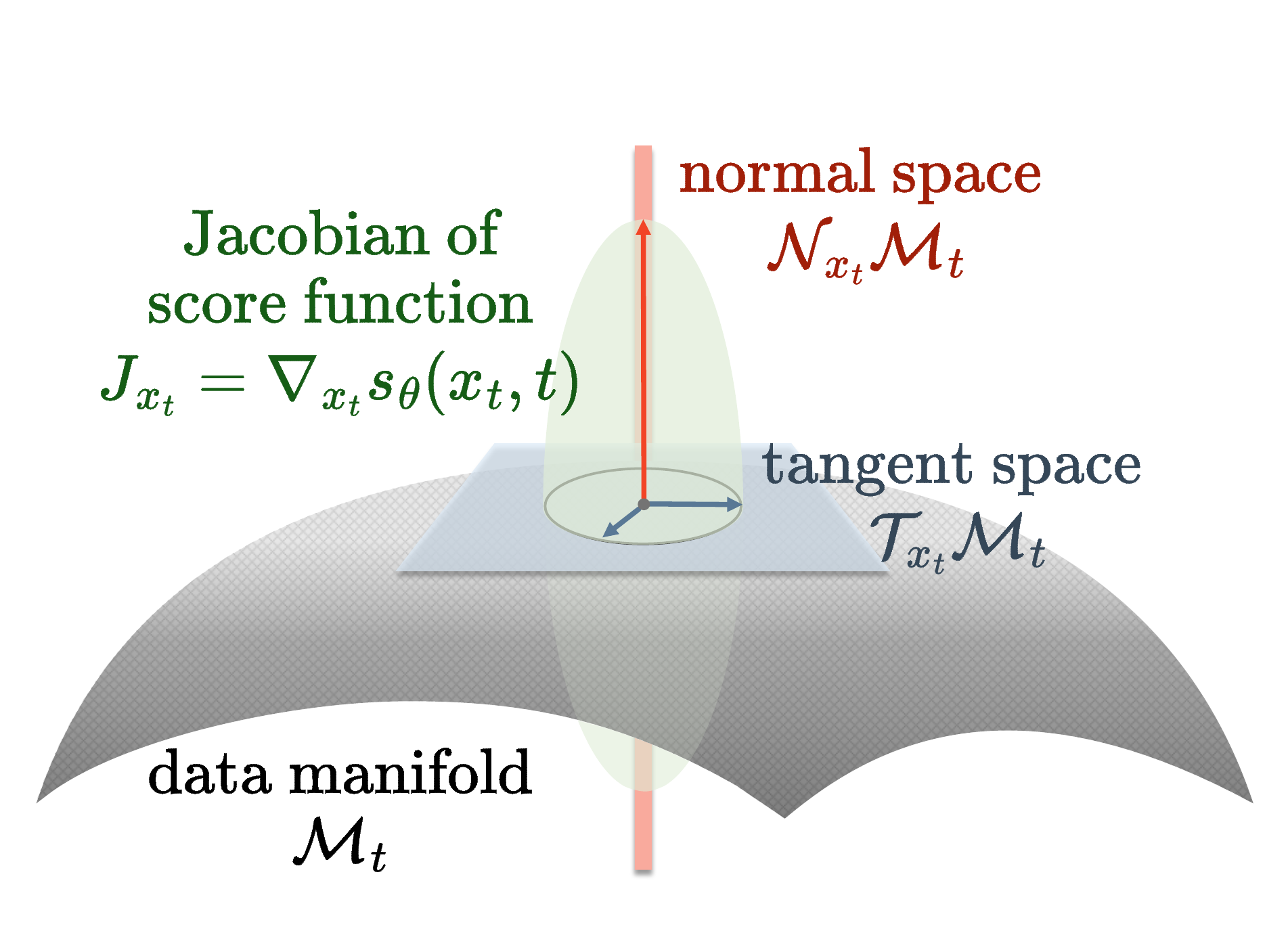}
    \caption{
        \textbf{A conceptual illustration.}
        Our metric is derived from the Jacobian $J_{x_t}$ of the score function $s_\theta$, whose spectral structure separates tangent and normal directions, $\gT_{x_t}\gM_t$ and $\gN_{x_t}\gM_t$, of the data manifold $\gM_t$.
    }
    \label{fig:concept}
\end{wrapfigure}

We answer this affirmatively.
We build on prior observations that the Jacobian of the score function already encodes this geometry: its spectral structure separates the tangent and normal directions of the data manifold (illustrated conceptually in \cref{fig:concept}) \cite{Stanczuk2024,Ventura2025}.
Building on this insight, we propose a Riemannian metric on the noise space, defined as the pullback of the Euclidean metric through the score function.
This metric assigns high cost to movement along normal directions and low cost along tangent directions, thereby encouraging paths to stay tangential to the data manifold.
Because it is constructed solely from the Jacobian of a pre-trained score function, it requires no additional training or architectural modifications.

The contribution of this work is the proposed Riemannian metric itself, which turns this known tangent--normal characterization into a concrete geometric foundation for understanding the data manifold in diffusion models and designing manifold-aware operations.
We examine the metric from two complementary angles.
Together, these angles serve to validate that the metric faithfully captures the geometry of the data manifold and to demonstrate its practical utility.
\textbf{(i)~Global geometry through interpolation} (\cref{sec:synthetic,sec:image_interp,sec:video_interp}):
Geodesics under our metric should stay on or run parallel to the manifold, producing perceptually natural transitions.
Experiments on synthetic, image, and video frame datasets confirm this.
\textbf{(ii)~Local geometry through guidance correction} (\cref{sec:guidance}):
The tangent--normal decomposition induced by our metric should prevent classifier-free guidance (CFG)~\cite{Ho2021} from pushing samples off the manifold.
The guidance correction consistently improves FID while maintaining CLIP Score, and the improved quality transfers to distilled models at no additional overhead.

\section{Related Work}
\label{sec:related-work}
\subsubsection{Riemannian Geometry of Deep Generative Models.}
In VAEs and GANs, the latent space parameterizes the data manifold, and linear traversals in this space are widely used for editing semantic attributes of generated images \cite{Goetschalckx2019,Harkonen2020,Plumerault2020,Shen2020,Voynov2020,Oldfield2021,Shen2021,Spingarn2021,Zhuang2021,Haas2022}.
However, as real-world data distributions are skewed and heterogeneous, linear manipulations often fall short, motivating nonlinear approaches~\cite{Ramesh2019,Jahanian2020,Tewari2020,Abdal2021,Khrulkov2021,Liang2021,Tzelepis2021,Chen2022,Choi2022,Aoshima2023}.
A principled way to handle such nonlinearity is to equip the latent space with a Riemannian metric \cite{Gruffaz2025}.
Some methods learn a metric through additional networks, which may overwrite the original manifold structure \cite{Yang2018,Arvanitidis2022,Lee2022,Sorrenson2025}.
Others construct the pullback metric from the Euclidean metric on the data space through the decoder (or the generator) of a pre-trained model without additional training \cite{Shao2017,Chen2018,Arvanitidis2018,Arvanitidis2021,Diepeveen2025}.
Our work follows the latter, training-free approach, but extends it to diffusion models, whose noise space is not low-dimensional.

\subsubsection{Data Manifold in Diffusion Models.}
Diffusion models learn a score function $s_\theta(x_t, t)$, which iteratively denoises noisy samples backward in time from $t=T$ to $t=0$ to form the data distribution \cite{Sohl-Dickstein2015,Ho2020,Song2021a,Song2021b,Rombach2022}.
A space of noisy samples at $t>0$ is often referred to as a \emph{noise space}.
Diffusion models have been shown to implicitly learn the data manifold through this denoising process \cite{Pidstrigach2022,Wenliang2022,Tang2024,Yun2024,George2025,Potaptchik2025}.
One common assumption is that high-density regions of the learned distribution correspond to the data manifold, but recent studies have shown that image likelihood is negatively correlated with perceptual detail: images in high-density regions are often over-smoothed, whereas those in lower-density regions may contain richer textures \cite{Karczewski2025a}.
From a different perspective, several studies have estimated the local intrinsic dimension of the data manifold \cite{Horvat2024,Kamkari2024,Stanczuk2024,Humayun2025,Ventura2025}.
Their key insight is that the Jacobian of the score function exhibits a rank deficiency corresponding to the codimension of the data manifold, or in practice a sharp spectral gap, which separates the normal directions of the data manifold from the tangent directions \cite{Stanczuk2024,Ventura2025}.
Rather than relying on density, we build upon this spectral characterization to define a Riemannian metric on the noise space of a pre-trained diffusion model.

\subsubsection{Interpolation in Diffusion Models.}
Although latent diffusion models \cite{Rombach2022} operate on the latent space of an autoencoder, this space is designed for perceptual compression rather than semantic representation, and simple interpolations therein do not yield coherent transitions.
Interpolation methods in the noise space can be broadly categorized into four groups.
(i)~Closed-form methods operate directly in the noise space.
LERP~\cite{Ho2020} treats it as a flat Euclidean space, leading to small norms and loss of detailed features.
SLERP~\cite{Shoemake1985,Song2021a} preserves norms along a hypersphere, and other variants adjust norms by other heuristics~\cite{Samuel2023,Bodin2025,Zheng2024}.
(ii)~Surrogate latent-space methods exploit intermediate layers of the diffusion network~\cite{He2024,Kwon2023,Park2023a,Park2023b}, but skip connections allow information to bypass these layers, limiting their ability to serve as self-contained representations.
(iii)~Training-dependent methods employ additional networks or modify the architecture for interpolation~\cite{Konpat2022,Kaiwen2023,Wang2023,Guo2024,Hahm2024,Lu2024,Shen2024,Yang2024,Kim2025}, but require additional training and cannot be applied to arbitrary pre-trained models.
(iv)~Riemannian metric-based methods define a geometry on the noise space without additional training.
GeodesicDiffusion~\cite{Yu2025} defines a conformal metric using the inverse density of noisy samples, and Azeglio and Bernardo~\cite{Azeglio2025} propose a metric inspired by the Fisher information matrix (FIM); both guide geodesics toward high-density regions, as examined for normalizing flows \cite{Sorrenson2025} and energy-based models \cite{Bethune2025}.
However, high-density regions do not necessarily correspond to perceptually detailed images~\cite{Karczewski2025a}, which can result in over-smoothed and cartoonish interpolations.
Although some studies draw inspiration from statistical manifolds, it remains unclear what geometric structures their methods leverage~\cite{Karczewski2025b,Lobashev2025}.
Our metric falls in this category but differs fundamentally: rather than favoring high-density regions, it leverages the spectral structure of the score Jacobian to keep geodesics tangential to the data manifold, keeping the density level.

\subsubsection{Guidance Methods.}
One can condition the denoising process on a text prompt \cite{Rombach2022}.
Classifier-free guidance (CFG) amplifies this condition to make generated images more faithful to text prompts \cite{Ho2021}, and the negative prompt suppresses concepts specified by a complementary prompt \cite{Rombach2022}.
Some guidance methods additionally utilize external information \cite{Chefer2023,Rassin2023,Sueyoshi2024}.
However, it is well known that these guidance methods can distort the learned data manifold, leading to visually unnatural outcomes.
Recent studies have attempted to mitigate this issue.
CFG++ \cite{Chung2025} uses the unconditional score instead of the CFG-based score for renoising steps, while TCFG \cite{Kwon2025}
projects the unconditional score onto the subspace shared with the conditional one via SVD.
Although these methods are motivated by the existence of the data manifold, they do not formally define or characterize its geometry, and their corrections remain heuristic.
Our metric provides such an explicit geometric characterization, which enables a geometrically grounded correction that decomposes the guidance term into tangent and normal components and suppresses the latter.

\section{Method}
\label{sec:method}
\subsection{Proposed Riemannian Metric}
\subsubsection{Definition of Proposed Metric.}
We define a Riemannian metric on the noise space that captures the local geometry of the data manifold through the score function.
See Appendix~\ref{appendix:riemannian_geometry} for background.
Let $x_t$ be a point in the noise space $\R^D$ at time $t$, and $v,w \in T_{x_t}\R^D$ be tangent vectors to $\R^D$ at $x_t$.
Let $s_\theta(x_t,t)$ be the score function of a pre-trained diffusion model parameterized by $\theta$.
We propose the following Riemannian metric at time $t$:
\begin{equation}
    \label{eq:jacobian_metric}
    \textstyle
    g_{x_t}(v, w) \coloneq \langle J_{x_t} v, J_{x_t} w \rangle = v^\top J_{x_t}^\top J_{x_t} w= v^\top G_{x_t} w,
\end{equation}
where $J_{x_t} = \nabla_{x_t} s_\theta(x_t, t)$ is the Jacobian of the score function $s_\theta(\cdot,t)$ at $x_t$, and $G_{x_t} = J_{x_t}^\top J_{x_t}$ is the matrix representation of the metric $g_{x_t}$.
This metric is the pullback $s_\theta^* I$ of the Euclidean metric $I$ on the score space $\R^D$ through the score function $s_\theta$, since $v^\top G_{x_t}w = (J_{x_t}v)^\top I (J_{x_t}w)$.
By construction, $G_{x_t}$ is symmetric and positive semi-definite; we show below that it is positive definite in practice.

\subsubsection{Interpretation.}
Stanczuk et al.~\cite{Stanczuk2024} found that the score function $s_\theta(x_t,t)$ points orthogonally towards the data manifold $\gM_t$ containing the data point $x_t$, and that this orthogonality becomes more prominent as $t$ approaches zero.
Ventura et al.~\cite{Ventura2025} further investigated the Jacobian $J_{x_t}$ and observed that its rank deficiency (for clean samples on a low-dimensional manifold) or spectral gap (for real-world data) reflects the intrinsic dimension of the data manifold.
Intuitively, $J_{x_t}$ shrinks along tangent directions and remains large along normal directions, as shown in \cref{fig:concept}.

More precisely, let $\gM_t$ be the data manifold at time $t$ learned by a diffusion model, and $x_t \in \gM_t$ be a point on the data manifold.
Define the tangent space $\gT_{x_t}\gM_t$ as the $d$-dimensional subspace ($d \ll D$) spanned by the right singular vectors of $J_{x_t}$ corresponding to small singular values; the normal space $\gN_{x_t}\gM_t$ is the orthogonal complement spanned by those corresponding to large singular values.
Then, $T_{x_t}\R^D=\gT_{x_t}\gM_t \oplus \gN_{x_t}\gM_t$.
From \cref{eq:jacobian_metric}, the squared metric norm of a tangent vector $v$ to the ambient noise space $\R^D$ is $g_{x_t}(v,v) = \|J_{x_t} v\|_2^2$.
\begin{proposition}
    \label{thm:equivalence}
    Among tangent vectors $v$ of a fixed Euclidean norm, those with the smallest squared metric norm $g_{x_t}(v,v)=\|J_{x_t} v\|_2^2$ lie in the tangent space $\gT_{x_t}\gM_t$ to the data manifold $\gM_t$.
\end{proposition}
Consequently, our metric assigns shorter lengths to paths that stay on or run parallel to the data manifold $\gM_t$.
See Appendix~\ref{appendix:proposition_1} for a detailed derivation.

\subsubsection{Geodesics.}
A geodesic is a shortest path under a given metric, obtained by minimizing the energy functional $E[\gamma]$ (see Appendix~\ref{appendix:riemannian_geometry} and Lee~\cite{Lee2019} for details).
Let $u \in [0,1]$ parameterize a curve $\gamma: u \mapsto \gamma(u)$.
The energy functional with our metric is:
\begin{equation}
    \label{eq:energy_jacobian_revisit}
    \begin{aligned}
        \textstyle E[\gamma]
         & \textstyle = \frac{1}{2} \int_0^1 g_{\gamma(u)}(\gamma'(u),\gamma'(u)) \dee u= \frac{1}{2} \int_0^1 \langle J_{\gamma(u)}\gamma'(u),J_{\gamma(u)}\gamma'(u)\rangle \dee u \\
         & \textstyle = \frac{1}{2}\int_0^1 \|J_{\gamma(u)} \gamma'(u) \|_2^2 \dee u
        \textstyle = \frac{1}{2}\int_0^1 \|\frac{\partial }{\partial u} s_\theta(\gamma(u),t) \|_2^2 \dee u.
    \end{aligned}
\end{equation}
This expression admits two complementary readings.
The second-last term shows that geodesics minimize the squared metric norm $\|J_{x_t} v\|_2^2$ of the velocity $v = \gamma'(u)$ at each point $x_t=\gamma(u)$ along the path, which by \cref{thm:equivalence} encourages the path to stay tangential to the data manifold.
The last term, obtained by the chain rule, shows that geodesics equivalently minimize the squared change in the score function $s_\theta$ along the path.
Since our metric is the pullback metric through $s_\theta$, geodesics are mapped to straight lines in the score space.
Because the score function encodes the gradient of the log-density, a path with minimal score variation maintains a consistent relationship with the density landscape: it neither approaches nor deviates from high-density regions, but runs parallel to the manifold at a consistent distance.
This provides a geometric explanation for why our geodesics preserve endpoint probabilities (\cref{fig:illust} (right)).
Moreover, considering that gradients of log-likelihoods with respect to model parameters have been used as semantic representations of inputs \cite{Yeh2018,Charpiat2019,Hanawa2021}, geodesics that minimize score variation can also be viewed as preserving semantic closeness between samples.

Given a geodesic, the distance between its endpoints is the length of the geodesic, computed as $d_g(\gamma(0), \gamma(1)) = \int_0^1 \sqrt{g_{\gamma(u)}(\gamma'(u),\gamma'(u))} \dee u$.

\subsubsection{Theoretical Remarks.}
Our metric $G_{x_t}=J_{x_t}^\top J_{x_t}$ captures the full spectral structure of the Jacobian, that is, all tangent and normal directions.
This contrasts with the FIM-based metric $\lambda s_\theta s_\theta^\top+I$~\cite{Azeglio2025}, which captures only the rank-1 direction of the score $s_\theta$ and thus respects only one normal direction.
The distinction is essential when the codimension of the data manifold is greater than one, which is the typical case for real-world data (e.g., images encoded in a representation space of tens of thousands of dimensions exhibit intrinsic dimensionality of a few hundred~\cite{Stanczuk2024,Ventura2025}).
Thus, practical diffusion models operate in a regime where the tangent--normal distinction is meaningful.

Diffusion models learn the score function $s_\theta$ directly rather than the log-density $\log p_t$, and consequently, $J_{x_t}$ is not exactly symmetric.
Even then, $J_{x_t}$ typically exhibits a spectral gap, and the decomposition into $\gT_{x_t}\gM_t \oplus \gN_{x_t}\gM_t$ in \cref{thm:equivalence} holds approximately.

The Jacobian $J_{x_t}$ is degenerate on ideally clean data lying on a low-dimensional manifold at $t=0$, but for real-world data it is full-rank with a sharp spectral gap.
In practice, operations are performed at an intermediate time $t>0$, where samples are corrupted by noise and the Jacobian $J_{x_t}$ is full-rank and exhibits a moderate spectral gap, ensuring that $g_{x_t}$ is positive definite and thus a proper Riemannian metric.

See Appendix~\ref{appendix:proposition_1} for further details, including a discussion of regularization and alternative construction.

\begin{figure}[t]
    \begin{minipage}[t]{0.48\textwidth}
        \vspace{0pt}
        \scriptsize
        \refstepcounter{algorithm}
        \label{alg:geodesic_interpolation}
        \hrule
        \vspace{2pt}
        \noindent{\small\textbf{Algorithm~\thealgorithm} Interpolation}
        \vspace{2pt}
        \hrule
        \vspace{2pt}
        \begin{algorithmic}[1]
                \REQUIRE Samples $x_0^\sss{0}$, $x_0^\sss{1}$, score $s_\theta$, time $\tau$, points $N$, iterations $K$
                \STATE $\{x_\tau^\sss{0}, x_\tau^\sss{1}\} \!\leftarrow\! \{\text{DDIM-Fwd}(x_0^\sss{i}, \tau)\}_{i\in\{0,1\}}$
                \STATE $\{x_\tau^\sss{u_i}\}_{i=1}^{N-1} \!\leftarrow\! \{\text{SLERP}(x_\tau^\sss{0}, x_\tau^\sss{1}, u_i)\}_{i=1}^{N-1}$
                \FOR{$k = 1$ \TO $K$}
                    \STATE Compute $E$ via \cref{eq:energy_jacobian_diff}
                    \STATE Update $\{x_\tau^\sss{u_i}\}_{i=1}^{N-1}$ to minimize $E$
                \ENDFOR
                \STATE $\{\hat{x}_0^\sss{u_i}\}_{i=0}^{N} \leftarrow \{\text{DDIM-Bwd}(x_\tau^\sss{u_i}, \tau)\}_{i=0}^{N}$
                \RETURN $\{\hat{x}_0^\sss{u_i}\}_{i=0}^{N}$
            \end{algorithmic}
        \vspace{2pt}
        \hrule
    \end{minipage}
    \hfill
    \begin{minipage}[t]{0.48\textwidth}
        \vspace{0pt}
        \centering
        \includegraphics[scale=0.174]{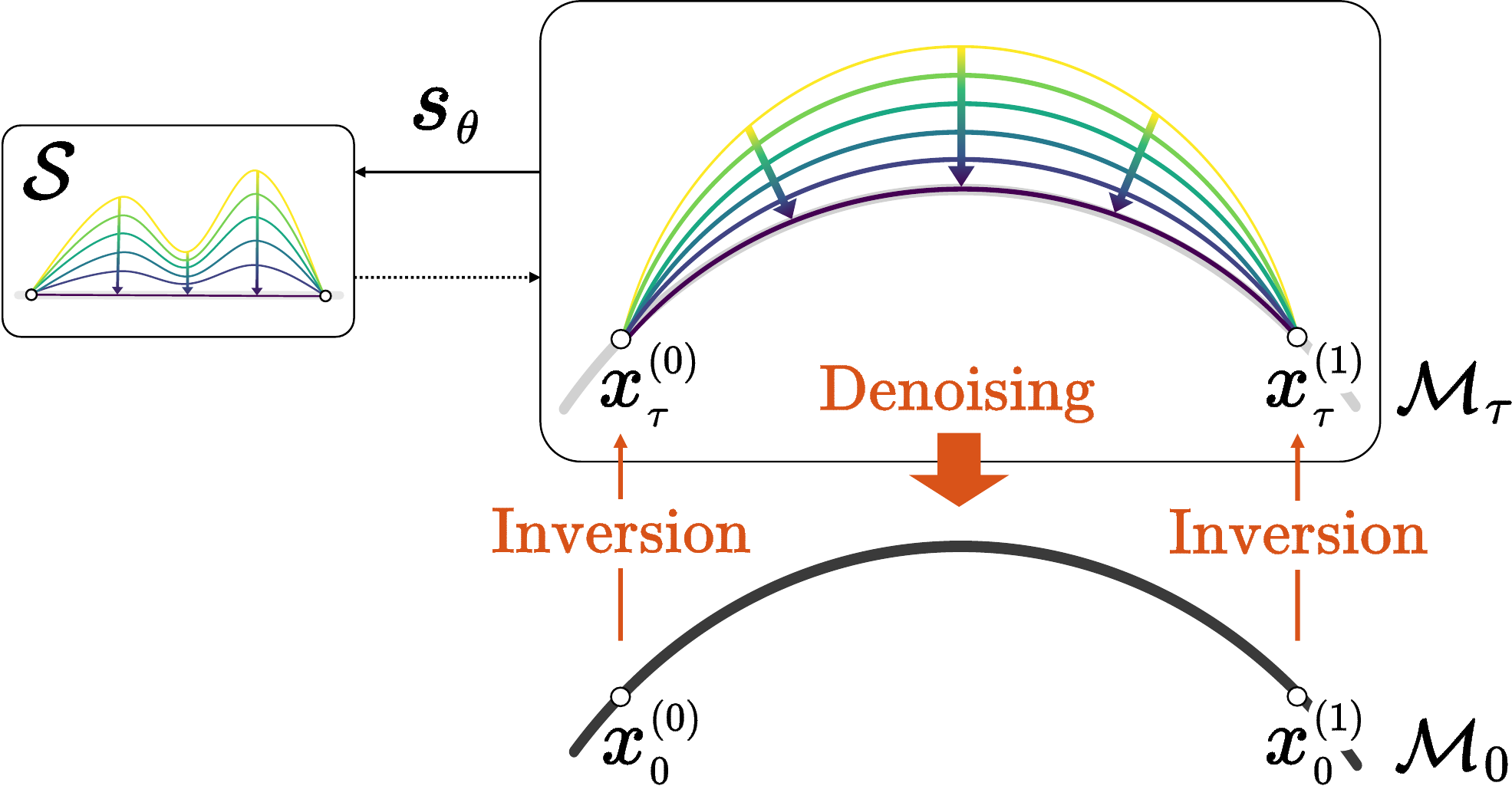}
        \caption{Illustration of geodesic-based interpolation}
        \label{fig:illustration_interp}
    \end{minipage}
\end{figure}

\subsection{From Metric to Algorithms}
\subsubsection{Geodesic-Based Interpolation.}
To use our metric for interpolation, we compute a geodesic in the noise space at a fixed time $t=\tau>0$ by discretizing the path and minimizing the resulting energy.
We approximate the path $\gamma$ as $N\!+\!1$ points $x_\tau^\sss{u_i}$ with $u_0=0$, $u_N=1$, $\Delta u=1/N$, and $x_\tau^\sss{u_i}=\gamma(u_i)$.
The energy functional $E[\gamma]$ in \cref{eq:energy_jacobian_revisit} is then approximated as:
\begin{equation}
    \label{eq:energy_jacobian_diff}
    \textstyle
    E[\gamma] \approx \frac{1}{2 \Delta u} \sum_{i=0}^{N-1} \|(s_\theta(x_\tau^\sss{u_{i+1}},\tau)-s_\theta(x_\tau^\sss{u_i},\tau))\|_2^2.
\end{equation}
Given two samples $x_\tau^\sss{0}$ and $x_\tau^\sss{1}$ at time $\tau$, the geodesic is obtained by minimizing \cref{eq:energy_jacobian_diff} with respect to the intermediate points $x_\tau^\sss{u_1}, \dots, x_\tau^\sss{u_{N-1}}$.
The distance is then $d_g(x_\tau^\sss{0},x_\tau^\sss{1})\approx\sum_{i=0}^{N-1} \|(s_\theta(x_\tau^\sss{u_{i+1}},\tau)-s_\theta(x_\tau^\sss{u_i},\tau))\|_2$.

This follows common practice in diffusion models, where interpolation is performed in the noise space rather than directly in the data space at $t=0$ (see \cref{alg:geodesic_interpolation} and \cref{fig:illustration_interp}).
Given a pair of clean samples $x_0^\sss{0}$ and $x_0^\sss{1}$, we first map them to noisy samples $x_\tau^\sss{0}$ and $x_\tau^\sss{1}$ at the fixed time $\tau$ using DDIM Inversion~\cite{Mokady2023}.
We then compute the geodesic path between them by minimizing \cref{eq:energy_jacobian_diff}, and map the interpolated noisy samples $x_\tau^\sss{u}$ for $u=u_0,\dots, u_{N}$ back to clean samples $x_0^\sss{u}$ using the deterministic denoising process (see Appendix~\ref{appendix:diffusion_models}).
Here, $x_0^\sss{u_1},\dots,x_0^\sss{u_{N-1}}$ serve as interpolated samples.
The optimization is more computationally expensive than closed-form methods such as LERP or SLERP, but comparable to other metric-based methods \cite{Azeglio2025,Yu2025} that also require solving an optimization problem, as summarized in Appendices~\ref{appendix:comparison_methods} and~\ref{appendix:comp_cost}.

\subsubsection{Metric-Based Guidance Correction.}
\begin{figure}[t]
    \begin{minipage}[t]{0.48\textwidth}
        \vspace{0pt}
        \scriptsize
        \refstepcounter{algorithm}
        \label{alg:guidance_correction}
        \hrule
        \vspace{2pt}
        \noindent{\small\textbf{Algorithm~\thealgorithm} Guidance Correction}
        \vspace{2pt}
        \hrule
        \vspace{2pt}
        \begin{algorithmic}[1]
                \REQUIRE $x_T \!\sim\! \mathcal{N}(0, I)$, condition $c$, CFG scale $w$, weight $\lambda$
                \FOR{$t = T$ \TO $1$}
                    \STATE $s \leftarrow s_\theta(x_t, t, c)$
                    \STATE $\tilde{s} \leftarrow \tilde{s}_\theta(x_t, t, c, w)$ \COMMENT{\cref{eq:cfg}}
                    \STATE $\Delta s \leftarrow \tilde{s} - s$
                    \STATE Compute $\Delta\hat{s}^*$ via \cref{eq:corrected_guidance}
                    \STATE $\hat{s} \leftarrow s + \Delta\hat{s}^*$
                    \STATE $x_{t-1} \leftarrow \text{DDIMStep}(x_t, \hat{s})$
                \ENDFOR
                \RETURN $x_0$
            \end{algorithmic}
        \vspace{2pt}
        \hrule
    \end{minipage}
    \hfill
    \begin{minipage}[t]{0.48\textwidth}
        \vspace{0pt}
        \centering
        \includegraphics[scale=0.13]{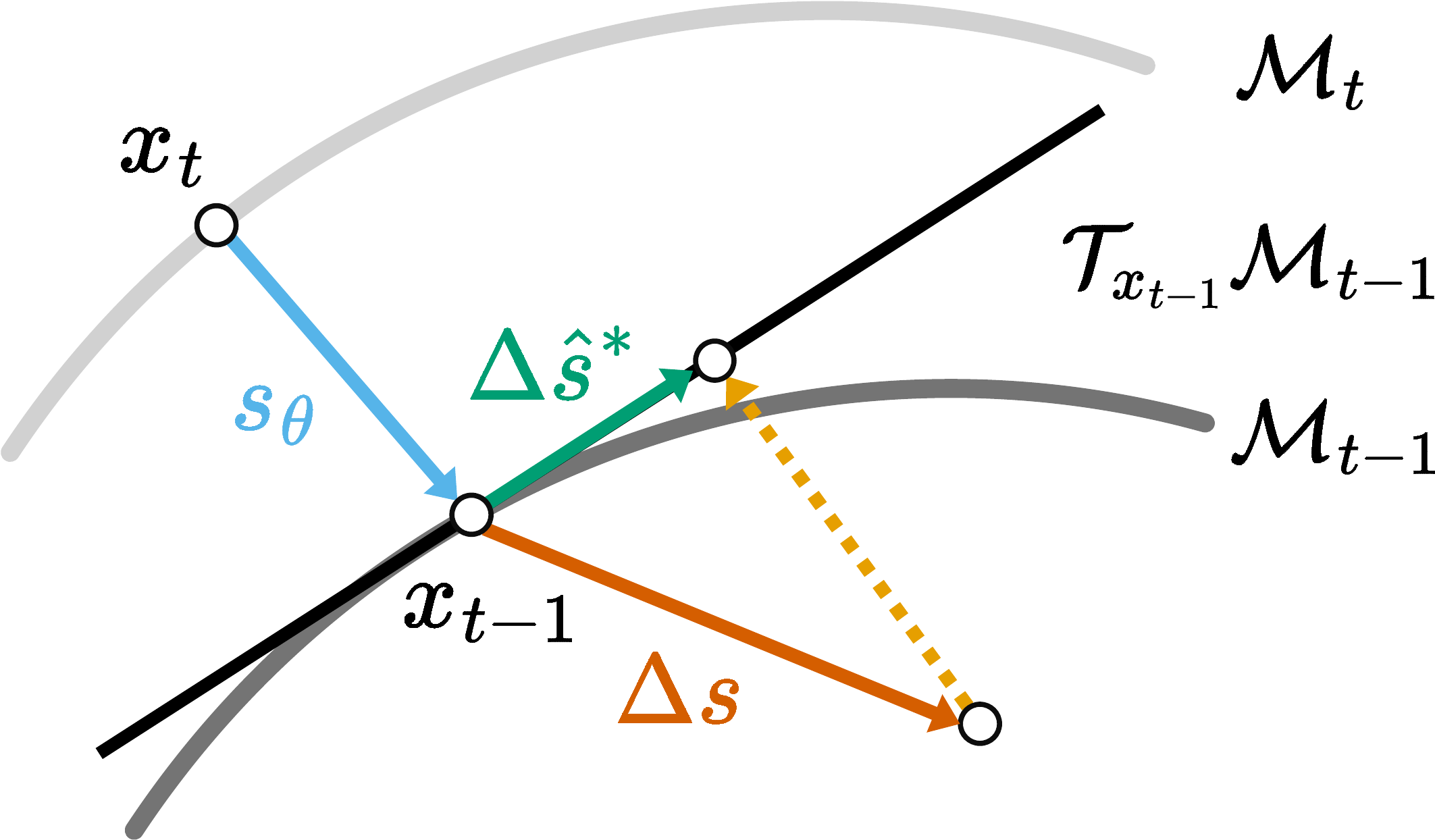}
        \caption{Illustration of metric-based guidance correction}
        \label{fig:illustration_guidance}
    \end{minipage}
\end{figure}

Complementary to global geodesics, our metric also provides a local tangent--normal decomposition at each point in the noise space.
This decomposition provides a metric-based way to correct perturbations that push a sample off the data manifold.
We apply this to classifier-free guidance (CFG), where the guidance term is known to distort the learned manifold~\cite{Chung2025,Kwon2025} (see \cref{alg:guidance_correction} and \cref{fig:illustration_guidance}).

Diffusion models generate samples by iteratively updating a noisy sample $x_t$ as $x_{t-1} = x_t + \eta_t s_\theta(x_t,t)$, where the noise term is omitted for simplicity, and $\eta_t$ is the update size depending on time $t$.
CFG modifies this by adding a guidance term $\Delta s(x_t,c)$ to $s_\theta(x_t,t)$, where $c$ is some condition.
Our correction adjusts this guidance term to balance two objectives: the corrected sample should remain close to the guided sample in Euclidean distance, while staying close to the unguided sample under our metric $g$, i.e., on the data manifold.
Formally, we solve:
\begin{equation}
    \Delta\hat{s}^*\!\!=\!\!\arg\min_{\Delta\hat{s}} d_E(x_t\!+\!\eta_t(s_\theta\!+\!\Delta s),\,x_t\!+\!\eta_t(s_\theta\!+\!\Delta\hat{s}))^2\!+\!\lambda d_g(x_t\!+\!\eta_t s_\theta,\,x_t\!+\!\eta_t(s_\theta\!+\!\Delta\hat{s}))^2,
\end{equation}
where $d_E$ is the Euclidean distance.

For tractability, we evaluate $d_g$ at time $t$ rather than $t-1$ (valid because the manifold changes smoothly between adjacent timesteps) and approximate the path integral with $N=1$ (sufficient for a local correction).
Absorbing $\eta_t$ into $\lambda$, the corrected guidance $\Delta\hat{s}^*$ is then given by $(I+\lambda G_{x_t})\Delta\hat{s}^* = \Delta s$.
We solve this by the conjugate gradient (CG) method with $\Delta s$ as the initial point, using only a single update step to limit computational overhead:
\begin{equation}\label{eq:corrected_guidance}\textstyle
    \Delta\hat{s}^*=\Delta s+\frac{\epsilon^\top\epsilon}{\epsilon^\top(I+\lambda G_{x_t})\epsilon}\epsilon \mbox{ for } \epsilon=\Delta s-(I+\lambda G_{x_t})\Delta s.
\end{equation}
Since $G_{x_t}=J_{x_t}^\top J_{x_t}$, multiplication by $G_{x_t}$ can be decomposed into a Jacobian-vector product followed by a vector-Jacobian product.
We approximate the Jacobian-vector product $J_{x_t}v$ via a finite difference $(s_\theta(x_t+h v,t)-s_\theta(x_t,t))/h$ for a small $h>0$, and approximate the vector-Jacobian product $J_{x_t}^\top v$ in the same way under the empirically supported assumption that $J_{x_t}$ is approximately symmetric.
In total, the correction requires four additional evaluations of $s_\theta$.
With CFG or negative prompt, each denoising step involves two evaluations of $s_\theta$, so the computational cost is approximately three times that without correction.
Since our correction modifies only the guidance term, it can be applied during distillation, so that the improved quality is inherited by the student model at no additional inference cost.

\section{Experiments}
\label{sec:experiments}
Our primary contribution is the Riemannian metric defined in \cref{eq:jacobian_metric}.
To validate that it faithfully captures the geometry of the data manifold, we examine its global properties through interpolation (\cref{sec:synthetic,sec:image_interp,sec:video_interp}) and its local properties through guidance correction (\cref{sec:guidance}).

\subsection{Synthetic 2D Data}
\label{sec:synthetic}
To illustrate the behavior of geodesics under our metric, we conducted experiments on a synthetic C-shaped distribution in 2D (see \cref{fig:illust} and Appendix~\ref{appendix:synthetic} for details).
We trained a DDPM \cite{Ho2020} with a SiLU MLP backbone~\cite{Elfwing2017} on 100{,}000 samples from this distribution with $T=50$ steps and obtained interpolations at $\tau=0.02T=1$ via DDIM Inversion.
As shown in \cref{fig:illust} (left), LERP traverses through low-density regions, SLERP slightly deviates from the manifold, and density-based interpolation \cite{Yu2025} approaches a high-density region, not preserving the endpoint probabilities.
In contrast, our geodesic runs parallel to the data manifold, preserving the endpoint probabilities.

\begin{wrapfigure}[9]{r}{0.31\textwidth}
        \centering
        \vspace{-12mm}
        \begin{minipage}{0.82\linewidth}
            \centering
            \footnotesize
            \captionsetup{type=table,width=0.89\linewidth}
            \captionof{table}{Results on synthetic 2D data}
            \label{tab:synthetic}
            \setlength{\tabcolsep}{6.5pt}
            \begin{tabular}{@{}l c@{}}
                \toprule
                \textbf{Method}                & \textbf{Std.~}$\downarrow$ \\
                \midrule
                \textcolor{BLUE}{LERP}         & 0.1606                     \\
                \textcolor{GREEN}{SLERP}       & \snd{0.0833}               \\
                \textcolor{VERMILION}{Density} & 0.1073                     \\
                \midrule
                \textcolor{BLACK}{Ours}        & \fst{0.0701}               \\
                \bottomrule
            \end{tabular}
        \end{minipage}
\end{wrapfigure}

To quantify this, we sampled 50 pairs of endpoints and evaluated the standard deviation of the density $p(x)$ along each interpolation path, summarized in \cref{tab:synthetic}.
A smaller value indicates a more consistent distance from the data manifold.
Our metric achieves the lowest standard deviation, which confirms that geodesics under our metric maintain a consistent relationship with the manifold, as predicted by the analysis in \cref{sec:method}.

\subsection{Image Interpolation}

\label{sec:image_interp}
\subsubsection{Experimental Setup.}
Following prior work~\cite{Samuel2023,Zheng2024,Yu2025}, we use Stable Diffusion v2.1-base \cite{Rombach2022} as the backbone with $T=50$ timesteps and $N-1=9$ interpolated images.
We do not use CFG or negative prompts for DDIM inversion and denoising.
For the geodesic computation, we use negative prompts and treat the resulting update $\tilde s_\theta$ (i.e., \cref{eq:negative_prompt} in Appendix~\ref{appendix:diffusion_models}) as the score function $s_\theta$ for computing the Jacobian $J_{x_t}$ and thus the metric $g_{x_t}$.
We evaluate on four benchmark datasets: the animation and metamorphosis subsets of MorphBench (MB(A) and MB(M)) \cite{Kaiwen2023}, Animal Faces-HQ (AF) \cite{Choi2022b}, and CelebA-HQ (CA) \cite{Karras2018}.
For AF and CA, we curate 50 image pairs with LPIPS \cite{Zhang2018} below 0.6 to ensure semantic similarity, following Yu et al.~\cite{Yu2025}.
Further details are provided in Appendix~\ref{appendix:image_interpolation_datasets}.

We compare against LERP \cite{Ho2020}, SLERP \cite{Song2021a}, NAO \cite{Samuel2023}, NoiseDiffusion (NoiseDiff) \cite{Zheng2024}, FIM-based metric \cite{Azeglio2025}, and GeodesicDiffusion (GeoDiff) \cite{Yu2025}.
We use default settings for NAO, NoiseDiff, and GeoDiff based on their official codes.
Due to the lack of official code, we re-implemented the FIM-based metric with the same backbone as ours.
For fair comparison, LERP, SLERP, FIM-based metric, and our proposed metric use the DDIM Scheduler \cite{Song2021a} and operate in the noise space at $\tau=0.6T$, following Yu et al.~\cite{Yu2025}.
For FIM-based metric and our proposed metric, each path was initialized with SLERP and updated for 500 iterations using Adam optimizer \cite{Kingma2015} with a learning rate of $10^{-3}$, decayed with cosine annealing to $10^{-4}$ \cite{Loshchilov2017}.
We also adopted the prompt adjustment of Yu et al.~\cite{Yu2025}, which optimizes the text embedding similarly to textual inversion~\cite{Gal2023}; see Appendices~\ref{appendix:comparison_methods}, \ref{appendix:comp_cost}, and~\ref{appendix:prompt_adjustment} for details on comparison methods, computational overhead, and prompt adjustment.

We evaluate interpolation quality using four measures.
(1)~Perceptual Path Length (PPL) \cite{Karras2018b} is the sum of LPIPS between adjacent images, assessing the directness of transitions.
(2)~Perceptual Distance Variance (PDV) \cite{Kaiwen2023} is the standard deviation of adjacent LPIPS, assessing the uniformity of transitions.
(3)~Fr\'{e}chet Inception Distance (FID) \cite{Martin2017} is a distributional distance between input and interpolated images via the features of Inception v3 \cite{Szegedy2015}.
(4)~Reconstruction Error (RE) is the mean squared error between the input endpoints and the reconstructed endpoints.
PPL, PDV, and FID are standard measures for image interpolation~\cite{Samuel2023,Zheng2024,Yu2025}; we additionally report RE to verify whether the endpoints are faithfully preserved.
We follow the evaluation protocol of prior work, including the sample sizes for FID computation (900 interpolated images vs.\ 200 reference images on AF and CA).
While none of these measures directly evaluates manifold faithfulness, consistent improvement across all four provides converging evidence.
We further complement these indirect measures with ground-truth evaluation in video frame interpolation (\cref{sec:video_interp}).

\begin{table}[!t]
    \centering
    \footnotesize
    \caption{Results on image interpolation}
    \label{tab:main_results}
    \setlength{\tabcolsep}{5.2pt}
    \begin{tabular}{@{}l rrrr rrrr@{}}
        \toprule
                        & \multicolumn{4}{c}{\textbf{PPL} $\downarrow$} & \multicolumn{4}{c}{\textbf{PDV} $\downarrow$} \\
        \cmidrule(lr){2-5} \cmidrule(lr){6-9}
        \textbf{Method} & \multicolumn{1}{c}{MB(A)} & \multicolumn{1}{c}{MB(M)} & \multicolumn{1}{c}{CA} & \multicolumn{1}{c}{AF} & \multicolumn{1}{c}{MB(A)} & \multicolumn{1}{c}{MB(M)} & \multicolumn{1}{c}{CA} & \multicolumn{1}{c}{AF} \\
        \midrule
        \textcolor{BLUE}{LERP}           & 0.848              & 1.787              & 1.420              & 1.859              & 0.055              & 0.128              & 0.091              & 0.154              \\
        \textcolor{GREEN}{SLERP}         & 0.644              & 1.065              & 0.707              & 0.871              & 0.030              & \fst{0.055\sigtwo} & \fst{0.033\sigone} & \fst{0.022\sigone} \\
        \textcolor{PURPLE}{NAO}          & 2.868              & 4.299              & 2.121              & 2.443              & 0.163              & 0.164              & 0.154              & 0.173              \\
        \textcolor{ORANGE}{NoiseDiff}    & 3.618              & 2.011              & 2.098              & 3.250              & 0.064              & 0.085              & 0.069              & 0.083              \\
        \textcolor{VERMILION}{GeoDiff}   & \snd{0.402}        & \snd{1.021}        & \snd{0.669}        & \snd{0.842}        & \snd{0.024}        & \snd{0.073}        & 0.044              & 0.027              \\
        \textcolor{GRAY}{FIM-based}      & 3.358              & 4.429              & 4.152              & 5.249              & 0.142              & 0.187              & 0.172              & 0.196              \\
        \midrule
        \textcolor{BLACK}{Ours}          & \fst{0.380\sigone} & \fst{0.977\sigone} & \fst{0.633\sigtwo} & \fst{0.767\sigtwo} & \fst{0.021\sigone} & \snd{0.073}        & \snd{0.036}        & \snd{0.023}        \\
        \bottomrule
        \toprule
                        & \multicolumn{4}{c}{\textbf{FID} $\downarrow$} & \multicolumn{4}{c}{\textbf{RE} $\downarrow$ $\scriptscriptstyle (\times 10^{-3})$} \\
        \cmidrule(lr){2-5} \cmidrule(lr){6-9}
        \textbf{Method} & \multicolumn{1}{c}{MB(A)} & \multicolumn{1}{c}{MB(M)} & \multicolumn{1}{c}{CA} & \multicolumn{1}{c}{AF} & \multicolumn{1}{c}{MB(A)} & \multicolumn{1}{c}{MB(M)} & \multicolumn{1}{c}{CA} & \multicolumn{1}{c}{AF} \\
        \midrule
        \textcolor{BLUE}{LERP}           & 84.20              & 118.90             & 95.68              & 119.58             & 0.401              & 0.397              & 1.010              & 2.049              \\
        \textcolor{GREEN}{SLERP}         & 62.81              & 48.99              & 37.84              & 26.07              & 0.401              & 0.397              & 1.010              & 2.049              \\
        \textcolor{PURPLE}{NAO}          & 130.54             & 102.64             & 83.05              & 71.47              & 39.244             & 44.302             & 27.623             & 40.178             \\
        \textcolor{ORANGE}{NoiseDiff}    & 119.47             & 74.03              & 65.04              & 68.87              & 15.096             & 7.835              & 8.618              & 19.628             \\
        \textcolor{VERMILION}{GeoDiff}   & \snd{28.70}        & \snd{38.12}        & \snd{35.98}        & \snd{25.80}        & \snd{0.188}        & \snd{0.272}        & \snd{0.891}        & \snd{1.969}        \\
        \textcolor{GRAY}{FIM-based}      & 92.09              & 78.80              & 70.95              & 59.11              & 0.401              & 0.397              & 1.010              & 2.049              \\
        \midrule
        \textcolor{BLACK}{Ours}          & \fst{27.44}        & \fst{36.00}        & \fst{32.54}        & \fst{21.01}        & \fst{0.177\sigone} & \fst{0.201\sigtwo} & \fst{0.888\sigone} & \fst{1.962\sigtwo} \\
        \bottomrule
    \end{tabular}\\
    \raggedright\textsuperscript{\tiny\!*} and \textsuperscript{\tiny\!*\!*} indicate that the improvement over the second-best method is statistically significant at the 0.01 and 0.001 levels, respectively, according to a one-sided exact binomial test ($H_0:\ p = 0.5$).
\end{table}

\subsubsection{Results.}
We summarize the quantitative results in \cref{tab:main_results}.
Our metric achieves the best scores on all datasets for PPL, FID, and RE, and records the best PDV on MB(A) with a close second-best on the others.

\begin{figure}[p]
    \centering
    \scriptsize
    \setlength{\tabcolsep}{2pt}
    \renewcommand{\arraystretch}{0}
    \begin{tabular}{rc}
        \raisebox{2.9mm}{\textcolor{BLUE}{LERP}}          & {\includegraphics[width=10.4cm]{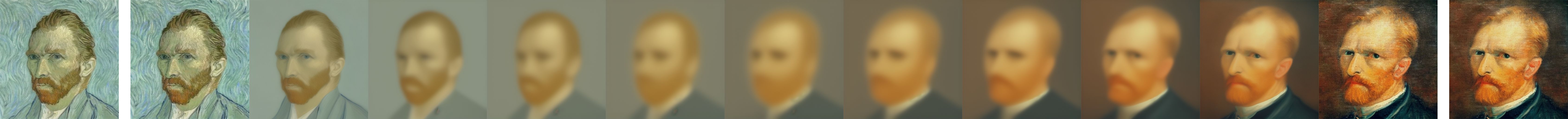}}      \\ \addlinespace[1.2pt]
        \raisebox{2.9mm}{\textcolor{GREEN}{SLERP}}        & {\includegraphics[width=10.4cm]{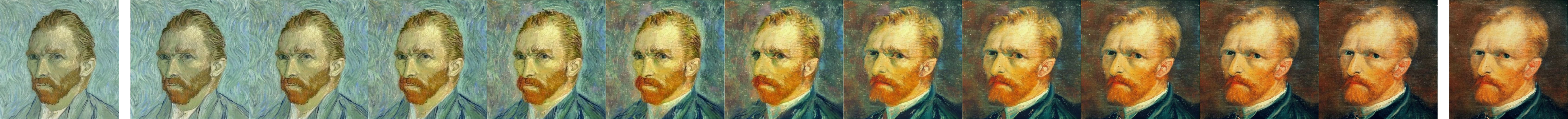}}     \\ \addlinespace[1.2pt]
        \raisebox{2.9mm}{\textcolor{PURPLE}{NAO}}         & {\includegraphics[width=10.4cm]{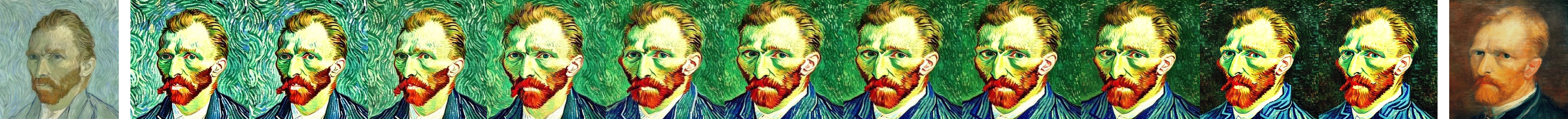}}       \\ \addlinespace[1.2pt]
        \raisebox{2.9mm}{\textcolor{ORANGE}{NoiseDiff}}   & {\includegraphics[width=10.4cm]{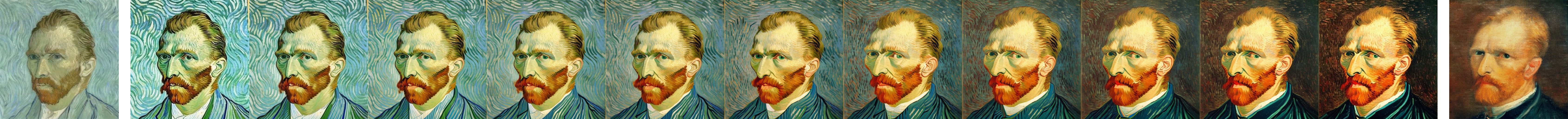}} \\ \addlinespace[1.2pt]
        \raisebox{2.9mm}{\textcolor{VERMILION}{GeoDiff}}  & {\includegraphics[width=10.4cm]{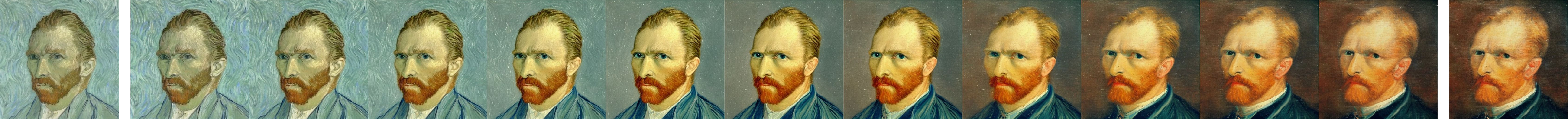}}   \\ \addlinespace[1.2pt]
        \raisebox{2.9mm}{\textcolor{GRAY}{FIM-based}}     & {\includegraphics[width=10.4cm]{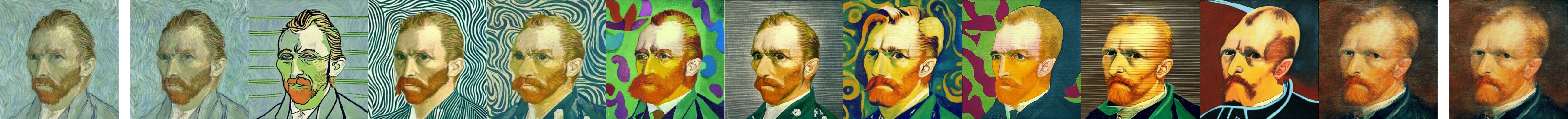}}       \\ \addlinespace[1.2pt]
        \raisebox{2.9mm}{\textcolor{BLACK}{Ours}}         & {\includegraphics[width=10.4cm]{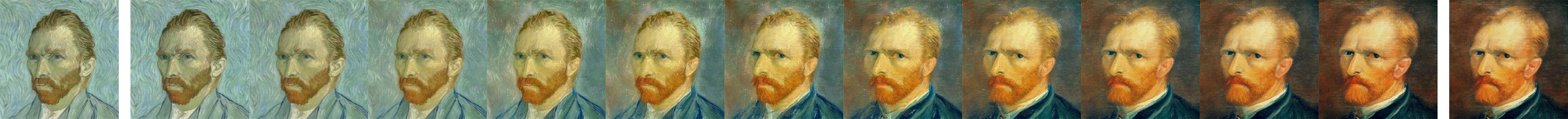}}      \\ \addlinespace[1.2pt]
                                                                  & \includegraphics[scale=0.946]{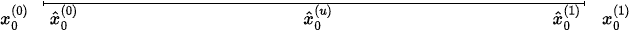}             \\ \addlinespace[1.2pt]
                                                                  & MorphBench (metamorphosis)                                                        \\ \addlinespace[6.0pt]
    \end{tabular}
    \begin{tabular}{rc}
        \raisebox{2.9mm}{\textcolor{BLUE}{LERP}}          & {\includegraphics[width=10.4cm]{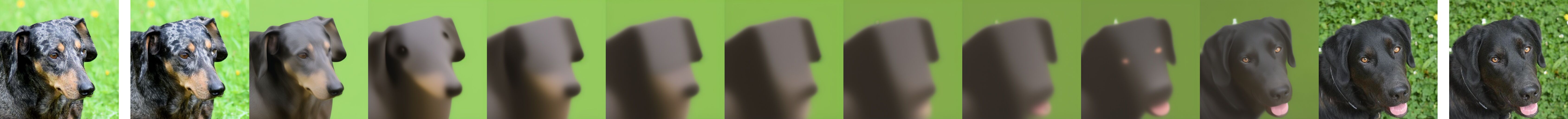}}         \\ \addlinespace[1.2pt]
        \raisebox{2.9mm}{\textcolor{GREEN}{SLERP}}        & {\includegraphics[width=10.4cm]{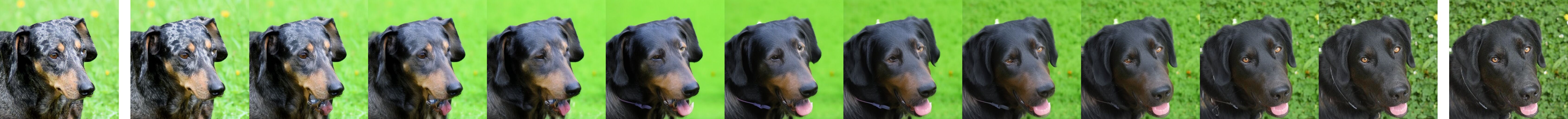}}        \\ \addlinespace[1.2pt]
        \raisebox{2.9mm}{\textcolor{PURPLE}{NAO}}         & {\includegraphics[width=10.4cm]{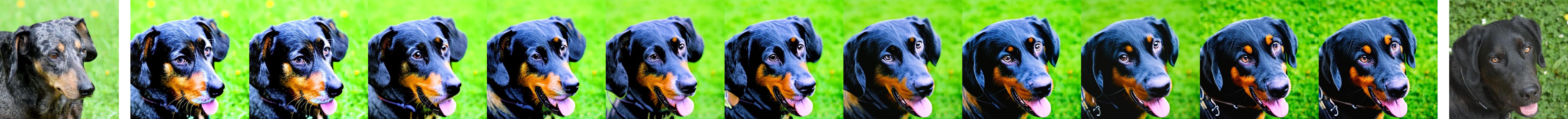}}          \\ \addlinespace[1.2pt]
        \raisebox{2.9mm}{\textcolor{ORANGE}{NoiseDiff}}   & {\includegraphics[width=10.4cm]{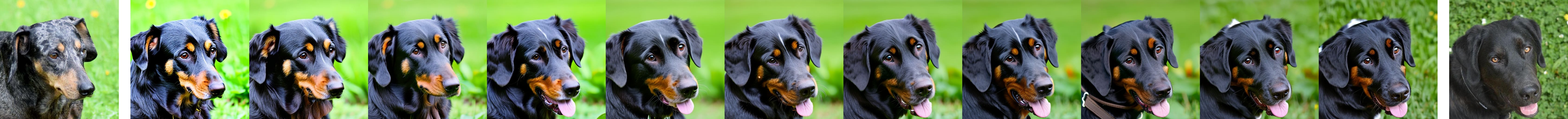}}    \\ \addlinespace[1.2pt]
        \raisebox{2.9mm}{\textcolor{VERMILION}{GeoDiff}}  & {\includegraphics[width=10.4cm]{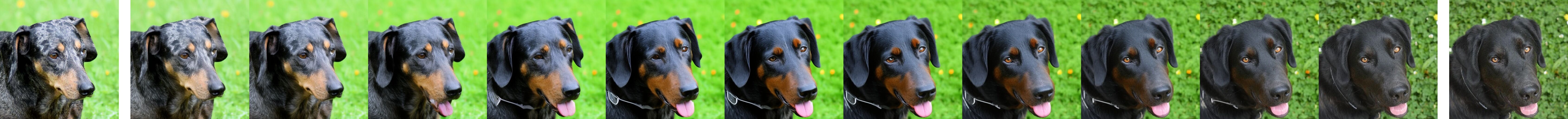}}      \\ \addlinespace[1.2pt]
        \raisebox{2.9mm}{\textcolor{GRAY}{FIM-based}}     & {\includegraphics[width=10.4cm]{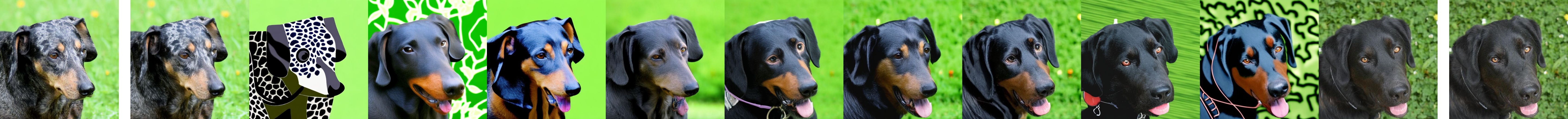}}          \\ \addlinespace[1.2pt]
        \raisebox{2.9mm}{\textcolor{BLACK}{Ours}}         & {\includegraphics[width=10.4cm]{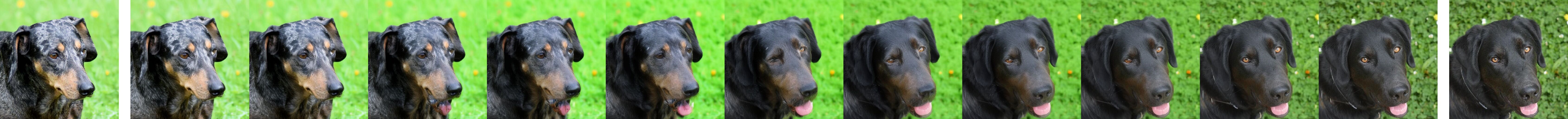}}         \\ \addlinespace[1.2pt]
                                                                  & \includegraphics[scale=0.946]{assets/utils/imagelabel.pdf}             \\ \addlinespace[3.0pt]
                                                                  & Animal Faces-HQ (Dog)                                                             \\ \addlinespace[1.5pt]
    \end{tabular}\\
    \caption{
        Qualitative examples of interpolated image sequences for MB(M) and AF (Dog).
        The images at both ends are the given endpoints $x_0^\sss{0}$ and $x_0^\sss{1}$, and the middle images are the interpolated results $\{\hat{x}_0^\sss{u}\}$ for $u\in[0,1]$.
        See also \cref{fig:results_qualitative_appendix} in Appendix~\ref{appendix:additional_results}.
    }
    \label{fig:results_qualitative}

    \vspace*{4.2mm}

    \centering
    \scriptsize
    \setlength{\tabcolsep}{2pt}
    \renewcommand{\arraystretch}{0}
    \begin{tabular}{rc}
        \raisebox{2.9mm}{MB (A)}    & {\includegraphics[width=10.4cm]{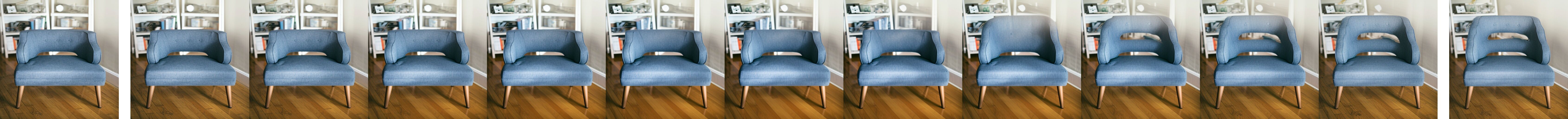}}    \\ \addlinespace[1.2pt]
        \raisebox{2.9mm}{CA}        & {\includegraphics[width=10.4cm]{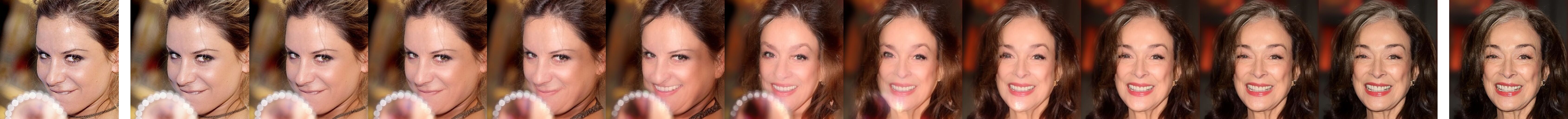}} \\ \addlinespace[1.2pt]
        \raisebox{2.9mm}{AF (Cat)}  & {\includegraphics[width=10.4cm]{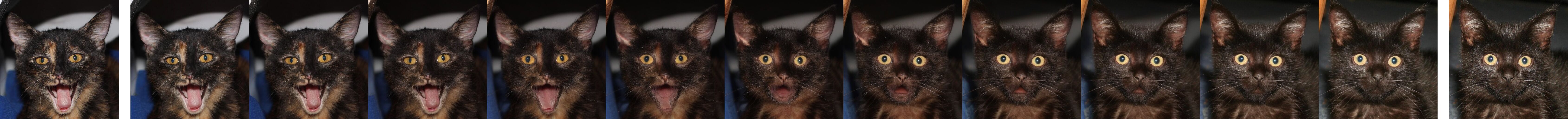}}     \\ \addlinespace[1.2pt]
                                    & \includegraphics[scale=0.946]{assets/utils/imagelabel.pdf}                 \\ \addlinespace[1.2pt]
    \end{tabular}
    \caption{
        Additional qualitative examples by our method.
        See also \cref{fig:results_qualitative_appendix} in Appendix~\ref{appendix:additional_results}.
    }
    \label{fig:results_qualitative2}
\end{figure}

Qualitative results in \cref{fig:results_qualitative,fig:results_qualitative2} reveal further differences.
LERP yields blurry interpolations.
NAO and NoiseDiff produce vivid textures absent in the originals and exhibit large reconstruction errors, as their norm adjustments alter the endpoints.
SLERP produces sharper results than LERP but lags behind geodesic-based methods.
The most competitive baseline is GeoDiff, which ranks second on most metrics but produces over-smoothed images that lack fine details.
This is consistent with the finding that sample density is negatively correlated with perceptual detail \cite{Karczewski2025a}: density-based metrics guide geodesics toward high-density regions, sacrificing texture.
FIM-based metric yields cartoonish results with non-smooth transitions, as it is nearly singular (rank-1) with the default $\lambda=1{,}000$.
In contrast, our metric preserves fine details of the input images throughout the interpolation.
We confirmed that these trends hold with a different backbone (Stable Diffusion v2.0-base); see Appendix~\ref{appendix:additional_results} for complementary results, including ablation studies and visualizations of the spectral gap.

\subsection{Video Frame Interpolation}
\label{sec:video_interp}

\subsubsection{Experimental Setup.}
To complement the indirect measures of the previous section, we evaluate methods on video frame interpolation using MSE and LPIPS against ground-truth middle frames.
Our goal is not to compete with dedicated video interpolation methods, but to use ground-truth frames as an objective proxy for manifold faithfulness.

We employ three benchmarks curated by Zhu et al.~\cite{Zhu2024}: 21 natural-scene clips from DAVIS~\cite{Perazzi2016}, 56 human-pose clips from Pexels (Human), and 26 indoor/outdoor clips from RealEstate10K (RE10K)~\cite{Zhou2018}.
From each clip, we use three consecutive frames to ensure a unique interpolation target: frames 1 and 3 serve as $x_0^\sss{0}$ and $x_0^\sss{1}$, and frame 2 is used as the ground-truth middle frame for evaluating $\hat x_0^\sss{0.5}$.
Unless otherwise specified, all methods and hyperparameters are identical to those used for image interpolation.
Each frame is resized to $512\times 512$ pixels, and a text prompt is generated from frame 1 using BLIP-2 \cite{Li2023}.

\begin{table}[t]
    \centering
    \footnotesize
    \setlength{\tabcolsep}{5.2pt}
    \caption{Results on video frame interpolation}
    \label{tab:video_results}
    \begin{tabular}{l rrr rrr}
        \toprule
                                      & \multicolumn{3}{c}{\textbf{MSE} $\downarrow$ $\scriptscriptstyle (\times 10^{-3})$} & \multicolumn{3}{c}{\textbf{LPIPS} $\downarrow$} \\
        \cmidrule(lr){2-4} \cmidrule(lr){5-7}
        \textbf{Method}               & \multicolumn{1}{c}{DAVIS} & \multicolumn{1}{c}{Human} & \multicolumn{1}{c}{RE10K} & \multicolumn{1}{c}{DAVIS} & \multicolumn{1}{c}{Human} & \multicolumn{1}{c}{RE10K} \\
        \midrule
        \textcolor{BLUE}{LERP}        & \snd{12.135}              & 4.566                     & 6.299                     & 0.590                     & 0.379                     & 0.377                     \\
        \textcolor{GREEN}{SLERP}      & 15.440                    & 6.080                     & 6.128                     & 0.487                     & 0.320                     & 0.301                     \\
        \textcolor{PURPLE}{NAO}       & 108.211                   & 99.867                    & 121.680                   & 0.679                     & 0.668                     & 0.664                     \\
        \textcolor{ORANGE}{NoiseDiff} & 46.881                    & 41.994                    & 28.867                    & 0.561                     & 0.552                     & 0.482                     \\
        \textcolor{VERMILION}{GeoDiff}  & 13.253                   & \snd{3.363}               & \snd{5.941}               & \snd{0.334}               & \snd{0.184}               & \snd{0.229}               \\
        \textcolor{GRAY}{FIM-based}   & 30.172                    & 11.638                    & 12.679                    & 0.535                     & 0.388                     & 0.373                     \\
        \midrule
        \textcolor{BLACK}{Ours}       & \fst{8.777\sigtwo}       & \fst{2.018\sigtwo}       & \fst{2.771\sigtwo}       & \fst{0.318\sigone}       & \fst{0.170\sigtwo}       & \fst{0.178\sigtwo}       \\
        \bottomrule
    \end{tabular}\\
    \textsuperscript{\tiny\!*} and \textsuperscript{\tiny\!*\!*} indicate the statistical significance in the same manner as \cref{tab:main_results}.
\end{table}

\begin{figure}[t]
    \centering
    \scriptsize
    \setlength{\tabcolsep}{0pt}
    \begin{tabular}{@{}>{\raggedleft\arraybackslash}m{0.90cm}@{\hspace{3pt}}*{10}{>{\centering\arraybackslash}m{1.10cm}@{\hspace{0pt}}}@{}}
        Human
            & \includegraphics[width=1.0cm]{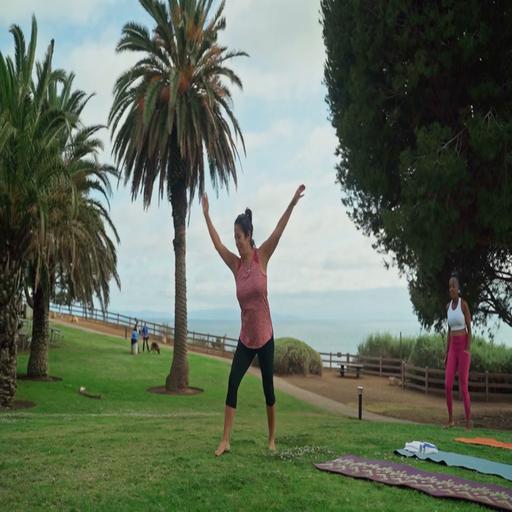}
            & \includegraphics[width=1.0cm]{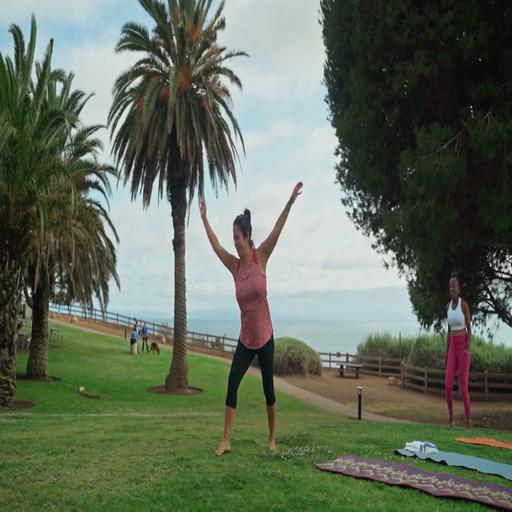}
            & \includegraphics[width=1.0cm]{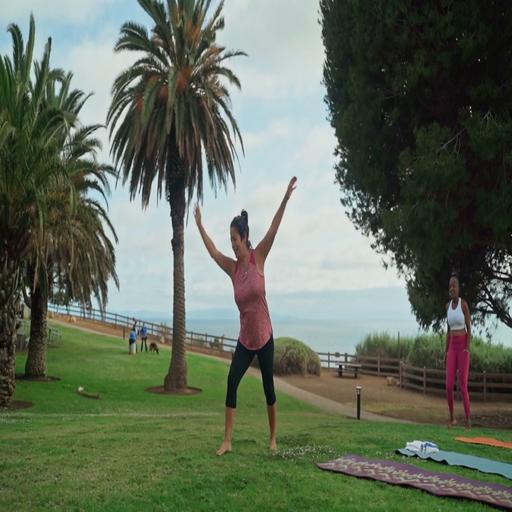}
            & \includegraphics[width=1.0cm]{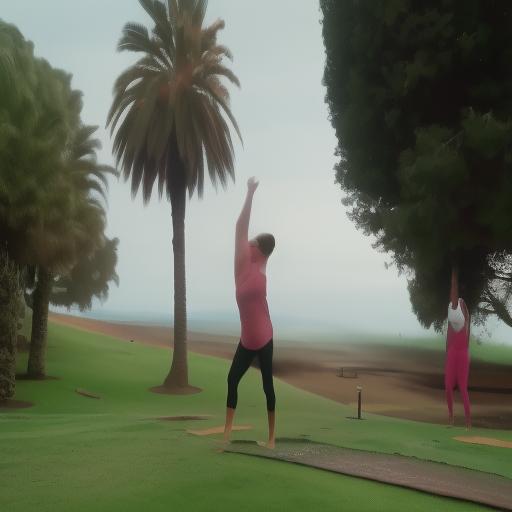}
            & \includegraphics[width=1.0cm]{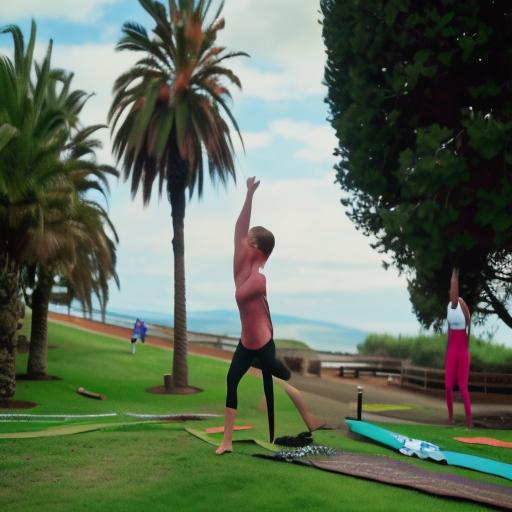}
            & \includegraphics[width=1.0cm]{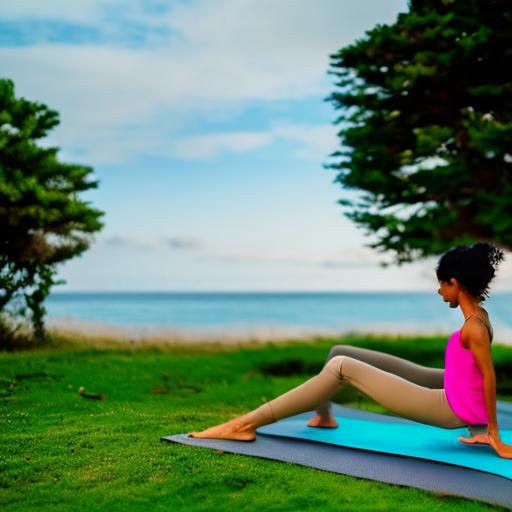}
            & \includegraphics[width=1.0cm]{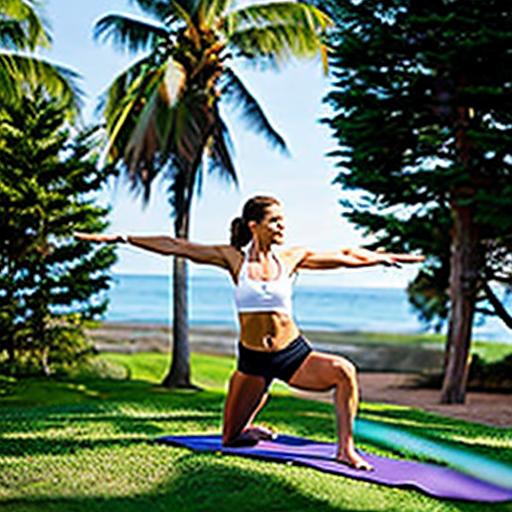}
            & \includegraphics[width=1.0cm]{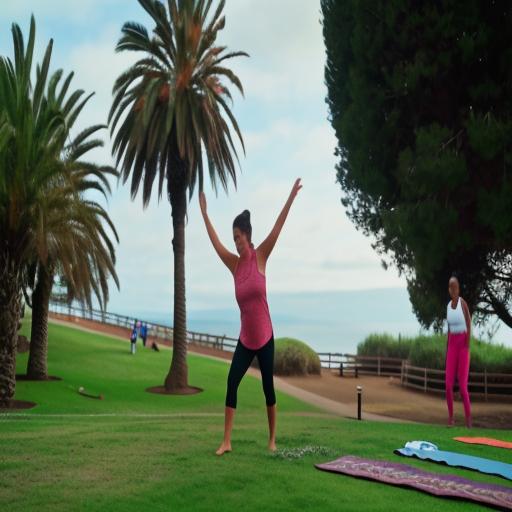}
            & \includegraphics[width=1.0cm]{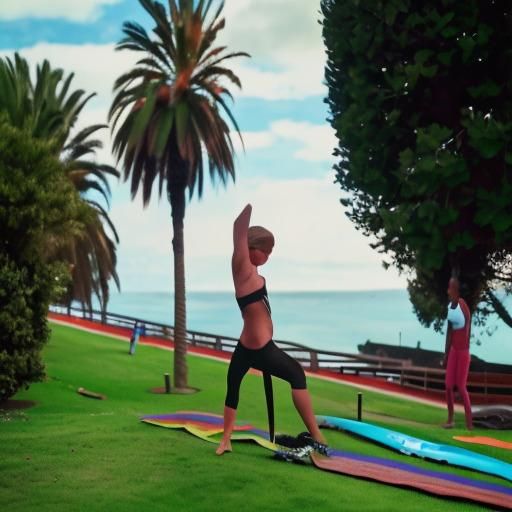}
            & \includegraphics[width=1.0cm]{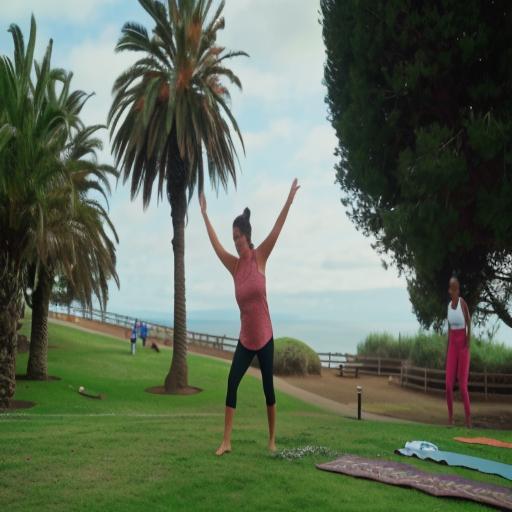} \\[-0.5mm]
        (Crop)
            & \includegraphics[width=1.0cm]{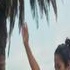}
            & \includegraphics[width=1.0cm]{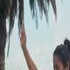}
            & \includegraphics[width=1.0cm]{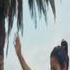}
            & \includegraphics[width=1.0cm]{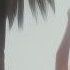}
            & \includegraphics[width=1.0cm]{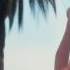}
            & \includegraphics[width=1.0cm]{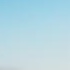}
            & \includegraphics[width=1.0cm]{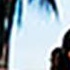}
            & \includegraphics[width=1.0cm]{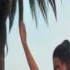}
            & \includegraphics[width=1.0cm]{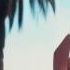}
            & \includegraphics[width=1.0cm]{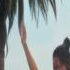} \\[-0.5mm]
        DAVIS
            & \includegraphics[width=1.0cm]{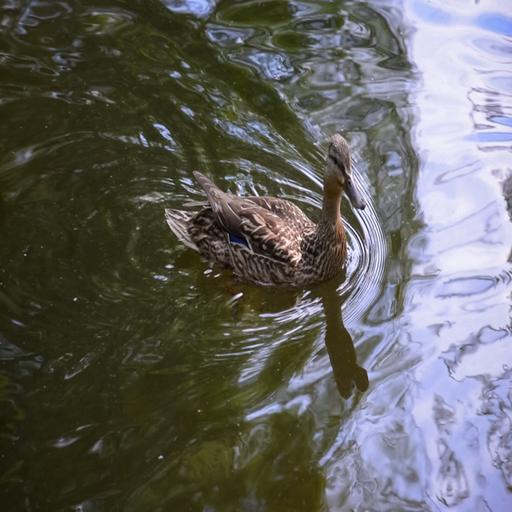}
            & \includegraphics[width=1.0cm]{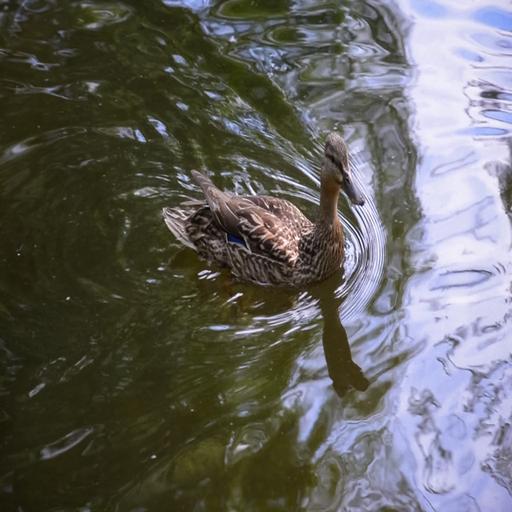}
            & \includegraphics[width=1.0cm]{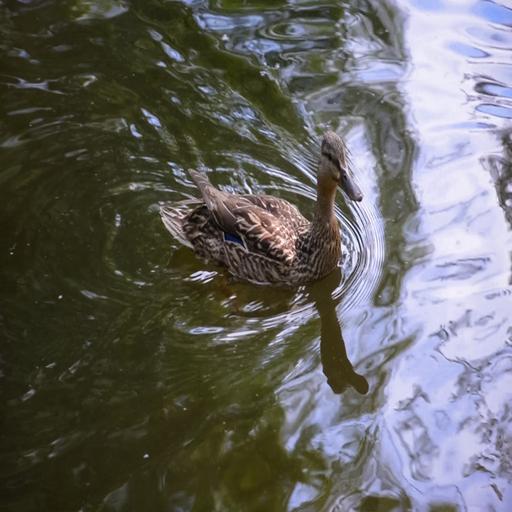}
            & \includegraphics[width=1.0cm]{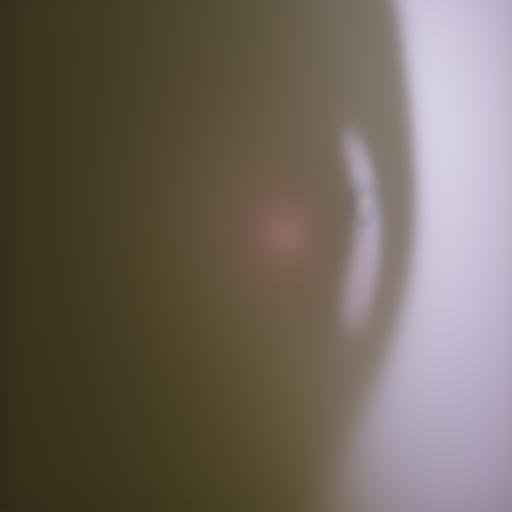}
            & \includegraphics[width=1.0cm]{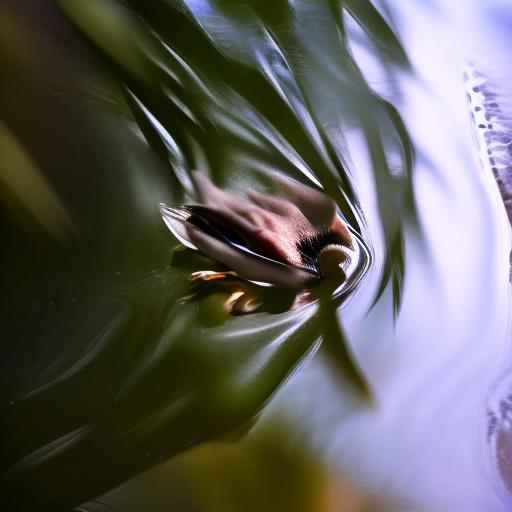}
            & \includegraphics[width=1.0cm]{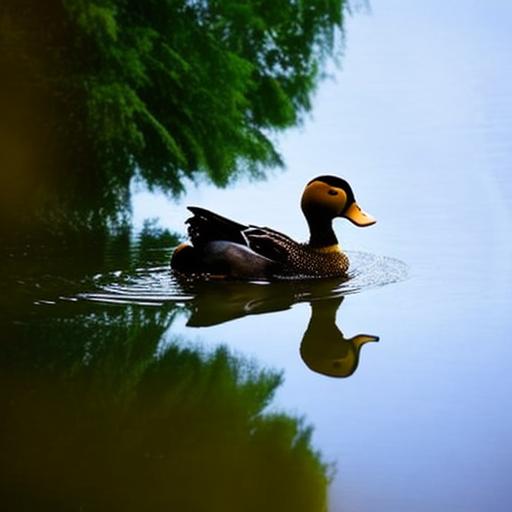}
            & \includegraphics[width=1.0cm]{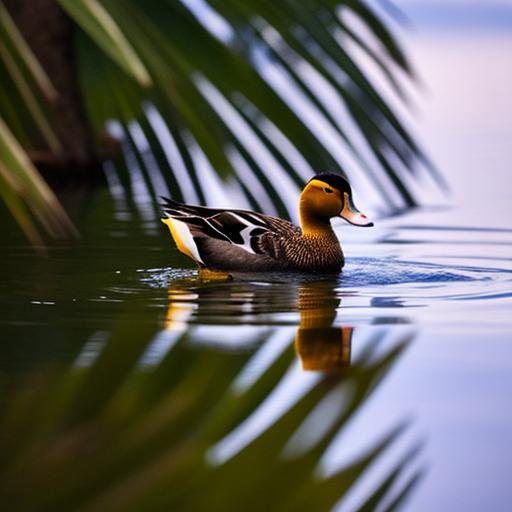}
            & \includegraphics[width=1.0cm]{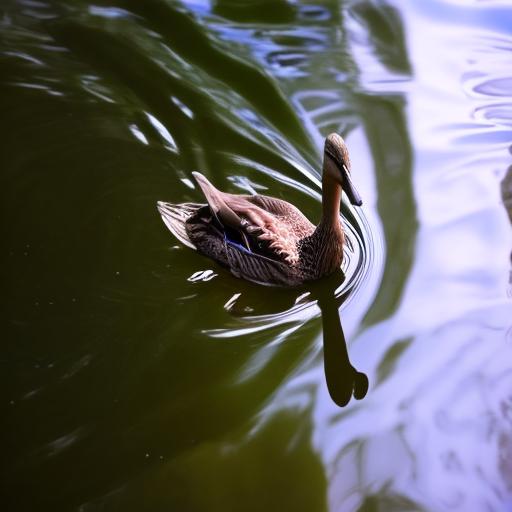}
            & \includegraphics[width=1.0cm]{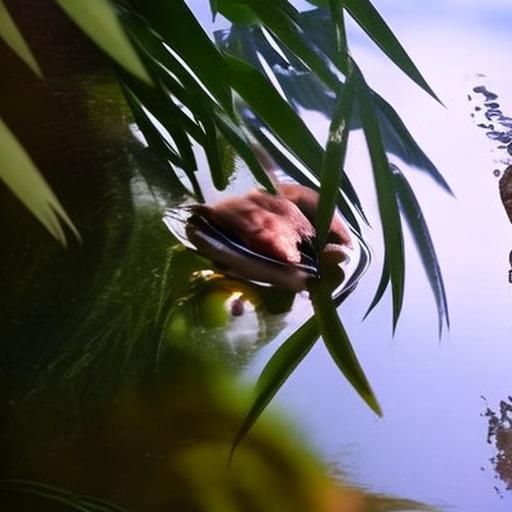}
            & \includegraphics[width=1.0cm]{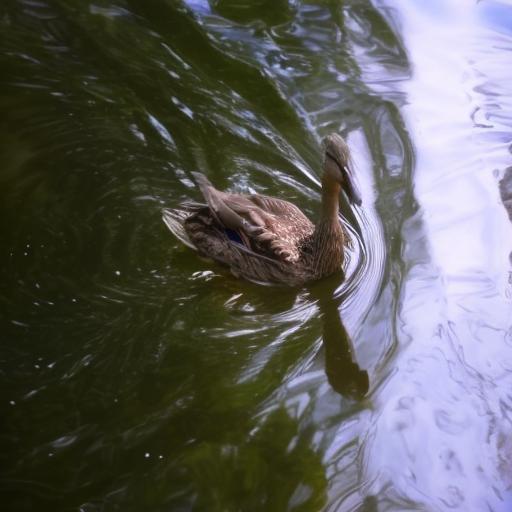} \\[-0.5mm]
        (Crop)
            & \includegraphics[width=1.0cm]{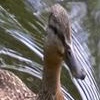}
            & \includegraphics[width=1.0cm]{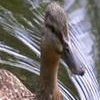}
            & \includegraphics[width=1.0cm]{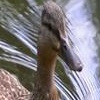}
            & \includegraphics[width=1.0cm]{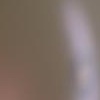}
            & \includegraphics[width=1.0cm]{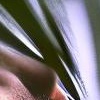}
            & \includegraphics[width=1.0cm]{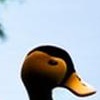}
            & \includegraphics[width=1.0cm]{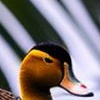}
            & \includegraphics[width=1.0cm]{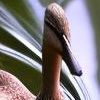}
            & \includegraphics[width=1.0cm]{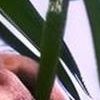}
            & \includegraphics[width=1.0cm]{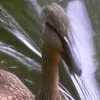} \\[-0.5mm]
        RE10K
            & \includegraphics[width=1.0cm]{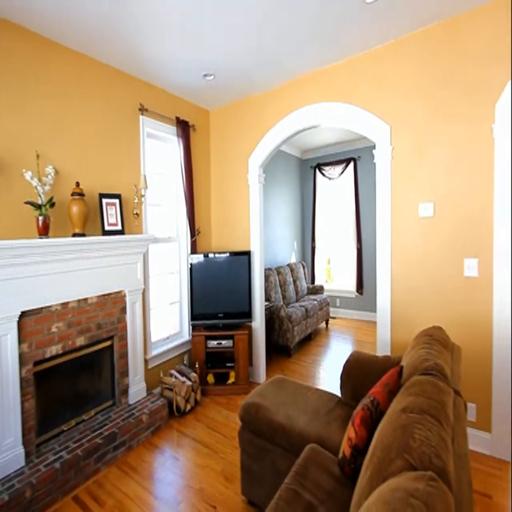}
            & \includegraphics[width=1.0cm]{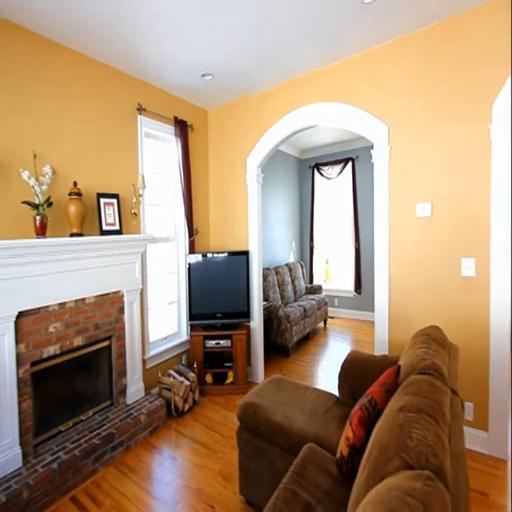}
            & \includegraphics[width=1.0cm]{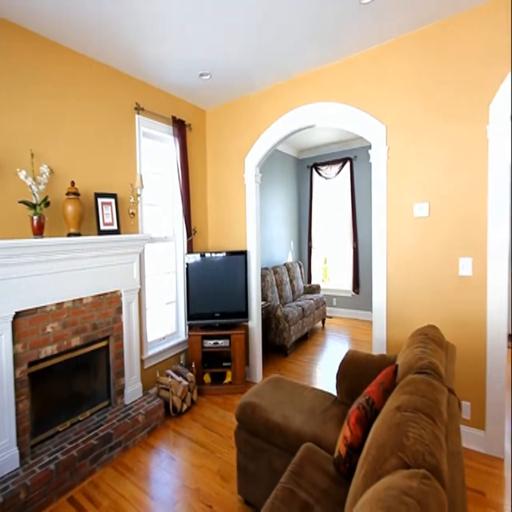}
            & \includegraphics[width=1.0cm]{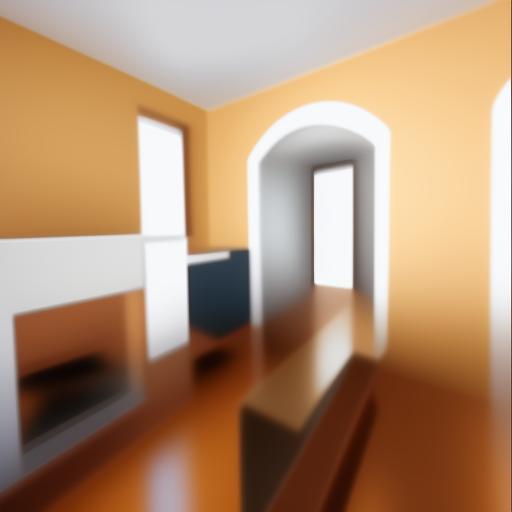}
            & \includegraphics[width=1.0cm]{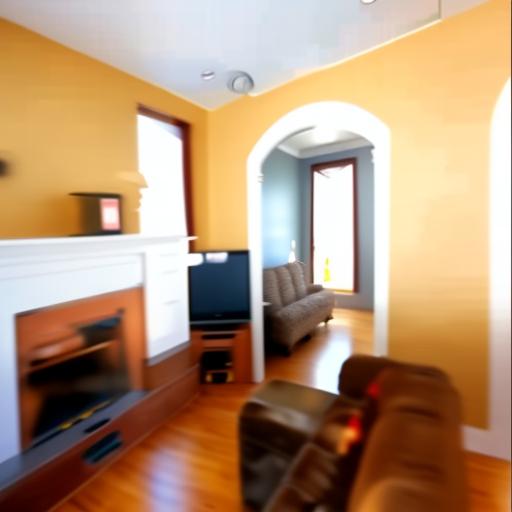}
            & \includegraphics[width=1.0cm]{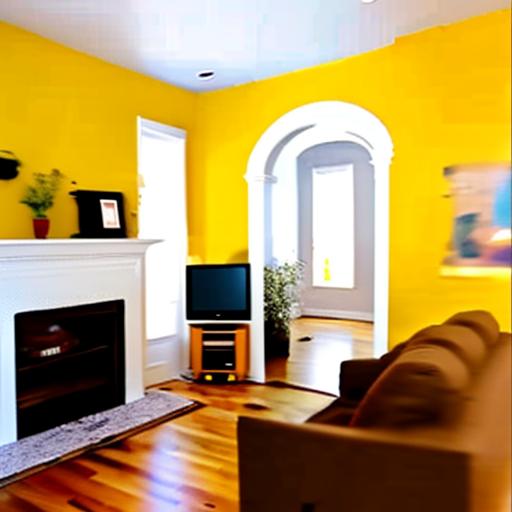}
            & \includegraphics[width=1.0cm]{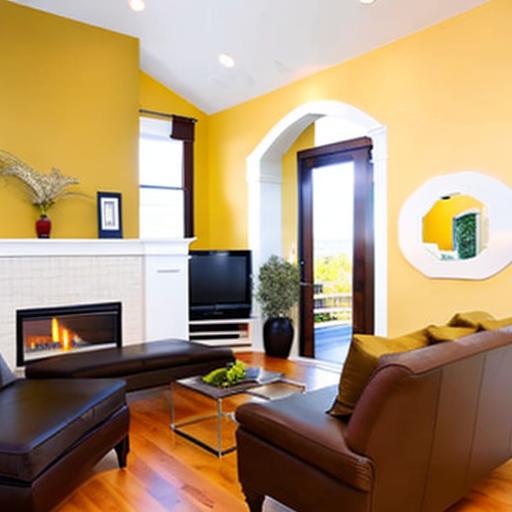}
            & \includegraphics[width=1.0cm]{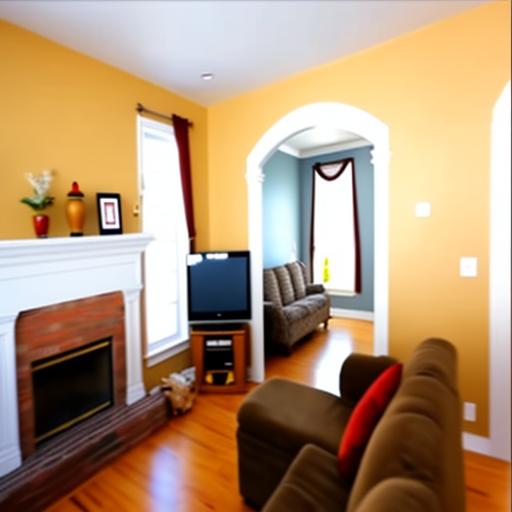}
            & \includegraphics[width=1.0cm]{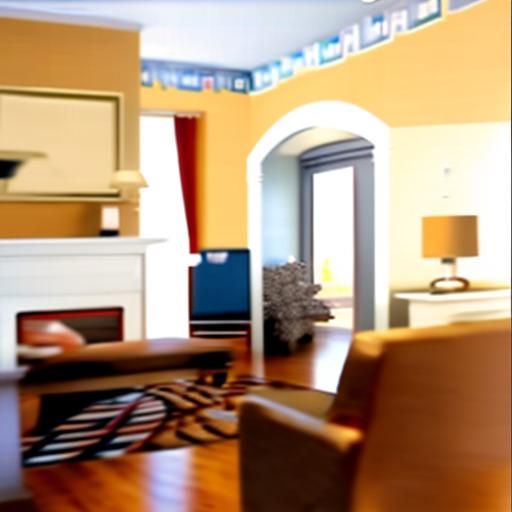}
            & \includegraphics[width=1.0cm]{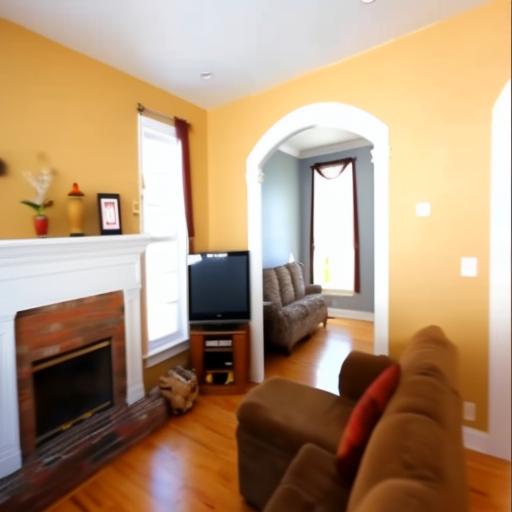} \\[-0.5mm]
        (Crop)
            & \includegraphics[width=1.0cm]{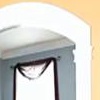}
            & \includegraphics[width=1.0cm]{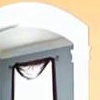}
            & \includegraphics[width=1.0cm]{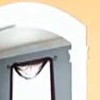}
            & \includegraphics[width=1.0cm]{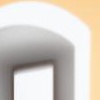}
            & \includegraphics[width=1.0cm]{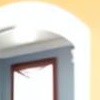}
            & \includegraphics[width=1.0cm]{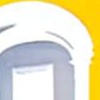}
            & \includegraphics[width=1.0cm]{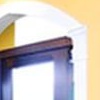}
            & \includegraphics[width=1.0cm]{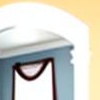}
            & \includegraphics[width=1.0cm]{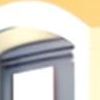}
            & \includegraphics[width=1.0cm]{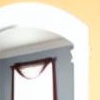} \\[-0.5mm]
            & frame 1
            & frame 2
            & frame 3
            & \textcolor{BLUE}{LERP}
            & \textcolor{GREEN}{SLERP}
            & \textcolor{PURPLE}{NAO}
            & \textcolor{ORANGE}{NoiseDiff}
            & \textcolor{VERMILION}{GeoDiff}
            & \textcolor{GRAY}{FIM-based}
            & \textcolor{BLACK}{Ours} \\
    \end{tabular}
    \caption{Qualitative examples for video frame interpolation}
    \label{fig:results_video}
\end{figure}

\subsubsection{Results.}
\Cref{tab:video_results} summarizes the quantitative results.
Our method achieves the lowest MSE and LPIPS on all datasets, with statistically significant improvements over the second-best method.
In \cref{fig:results_video}, the qualitative trends observed in image interpolation carry over: LERP produces blurry outputs, NAO and NoiseDiff deviate substantially from the ground truth, and GeoDiff over-smooths textures.
Notably, the zoomed-in comparison on Human shows that only our method and GeoDiff correctly interpolate the arm movement, but GeoDiff flattens water ripples on DAVIS, confirming that density-based geodesics sacrifice fine detail.
Our method preserves edges, object shapes, and textures most faithfully, as now confirmed by comparison against ground-truth frames.

\begin{table}[t]
    \footnotesize
    \begin{minipage}{0.72\textwidth}
        \centering
        \setlength{\tabcolsep}{2.5pt}
        \caption{Results on guidance correction}
        \label{tab:guidance_results}
        \begin{tabular}{l cccccc}
            \toprule
                                    & \multicolumn{2}{c}{$w=5.0$} & \multicolumn{2}{c}{$w=7.5$} & \multicolumn{2}{c}{$w=12.5$} \\
            \cmidrule(lr){2-3} \cmidrule(lr){4-5} \cmidrule(lr){6-7}
                                    & FID $\downarrow$            & CLIP $\uparrow$             & FID $\downarrow$            & CLIP $\uparrow$ & FID $\downarrow$ & CLIP $\uparrow$ \\
            \midrule
            CFG                     & \snd{11.69}                 & 0.313                       & 14.29                       & 0.314           & \snd{17.28}      & 0.315           \\
            CFG$++$                 & 11.87                       & 0.313                       & \snd{13.98}                 & 0.314           & 17.76            & 0.315           \\
            \midrule
            \textcolor{BLACK}{Ours} & \fst{11.53}                 & 0.313                       & \fst{13.81}                 & 0.314           & \fst{16.04}      & 0.315           \\
            \bottomrule
        \end{tabular}
    \end{minipage}
    \hfill
    \begin{minipage}{0.27\textwidth}
        \centering
        \setlength{\tabcolsep}{2.5pt}
        \caption{Results on distillation}
        \label{tab:distillation_results}
        \begin{tabular}{l cccccc}
            \toprule
                 & FID $\downarrow$ & CLIP $\uparrow$ \\
            \midrule
            CFG  & 17.91            & 0.306           \\
            Ours & \fst{16.98}      & 0.306           \\
            \bottomrule
        \end{tabular}
    \end{minipage}
\end{table}

\begin{figure}[t]
    \centering
    \scriptsize
    \setlength{\tabcolsep}{1.5pt}
    \renewcommand{\arraystretch}{0}
    \newlength{\guidimgw}
    \setlength{\guidimgw}{1.85cm}
    \begin{tabular}{r cccccc}
        \rotatebox{90}{\hspace{6.6mm}CFG}
            & \includegraphics[width=\guidimgw]{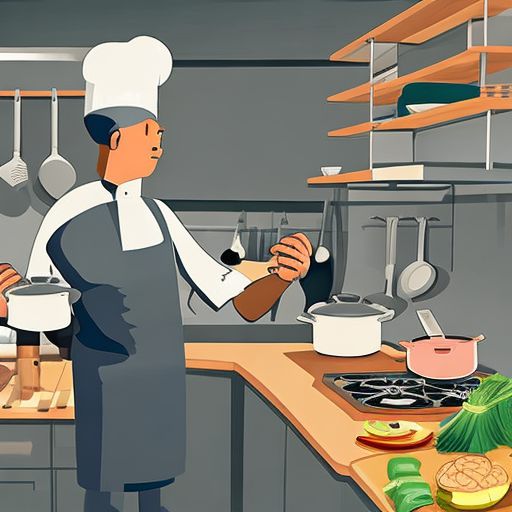}
            & \includegraphics[width=\guidimgw]{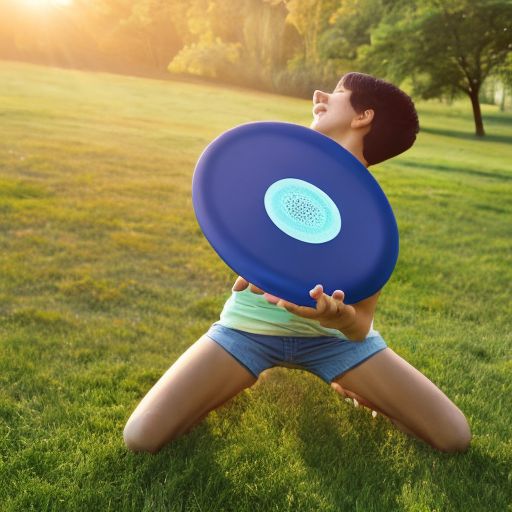}
            & \includegraphics[width=\guidimgw]{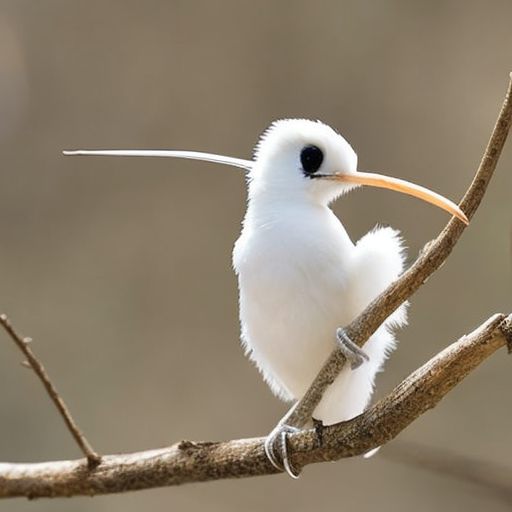}
            & \includegraphics[width=\guidimgw]{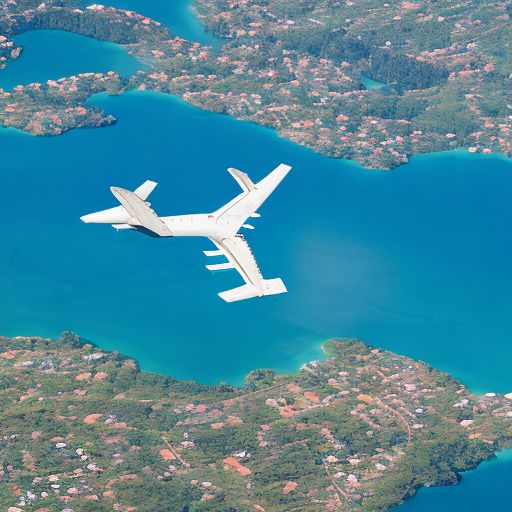}
            & \includegraphics[width=\guidimgw]{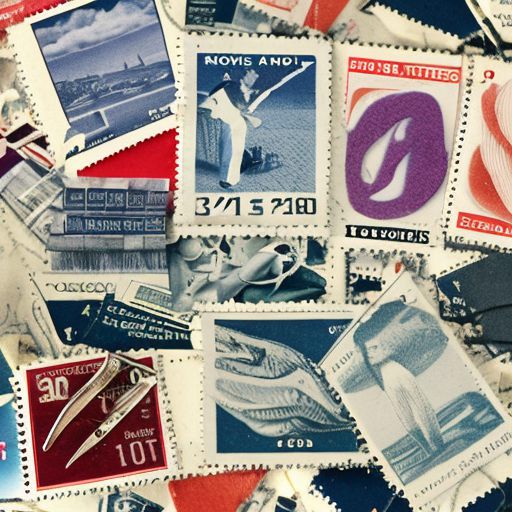}
            & \includegraphics[width=\guidimgw]{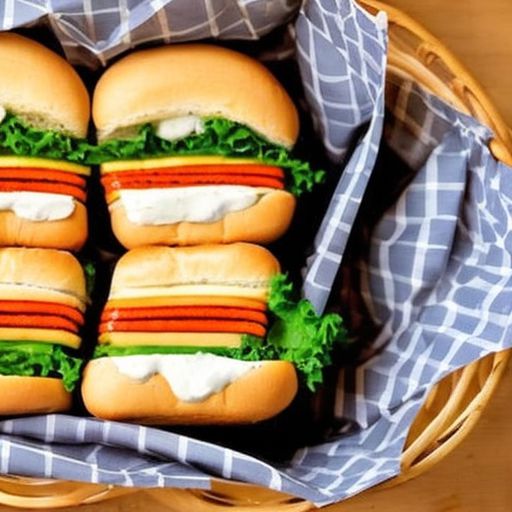} \\ \addlinespace[1.5pt]
        \rotatebox{90}{\hspace{4.4mm}CFG++}
            & \includegraphics[width=\guidimgw]{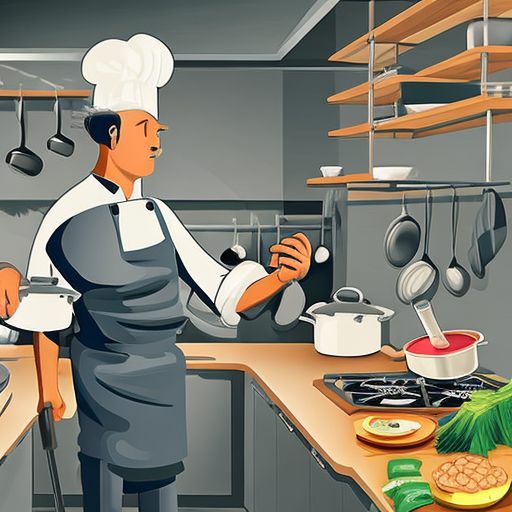}
            & \includegraphics[width=\guidimgw]{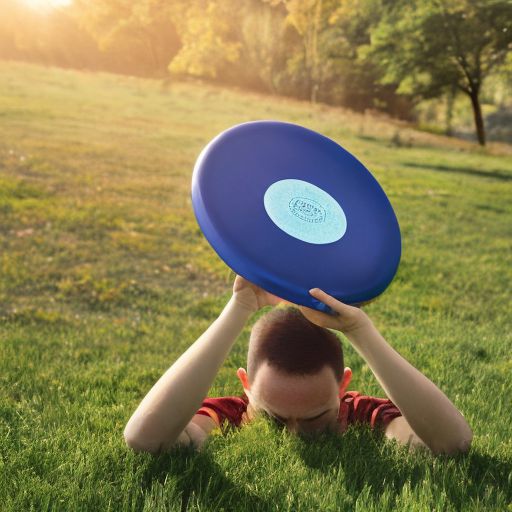}
            & \includegraphics[width=\guidimgw]{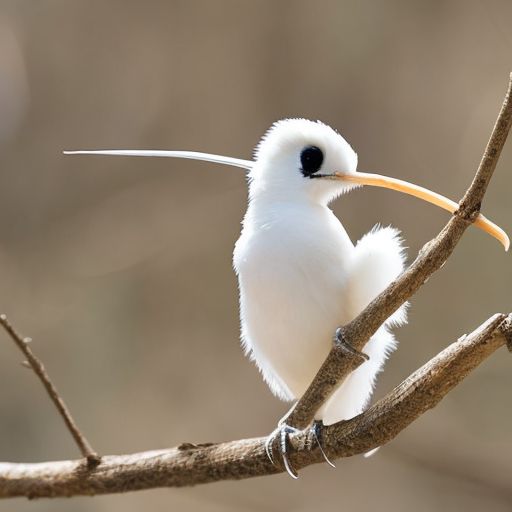}
            & \includegraphics[width=\guidimgw]{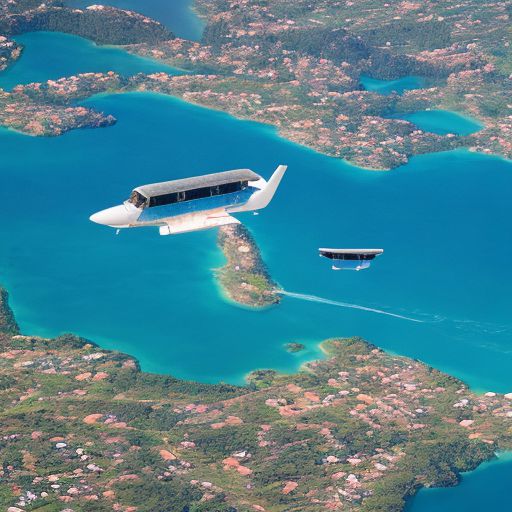}
            & \includegraphics[width=\guidimgw]{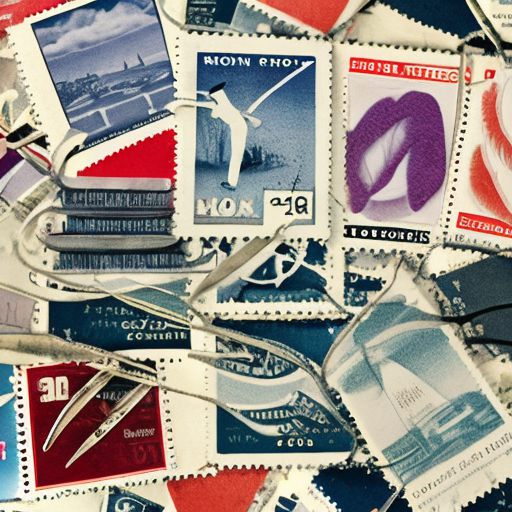}
            & \includegraphics[width=\guidimgw]{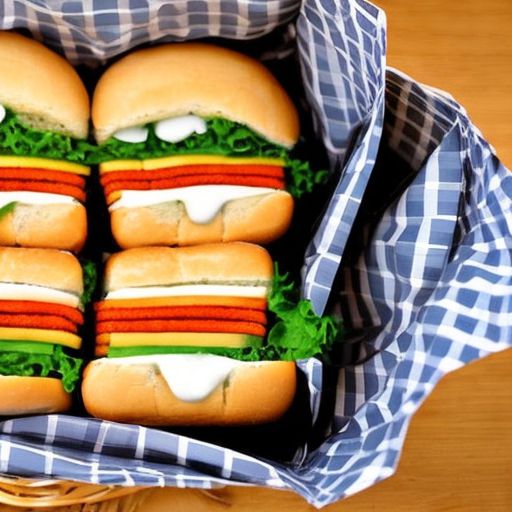} \\ \addlinespace[1.5pt]
        \rotatebox{90}{\hspace{6.5mm}Ours}
            & \includegraphics[width=\guidimgw]{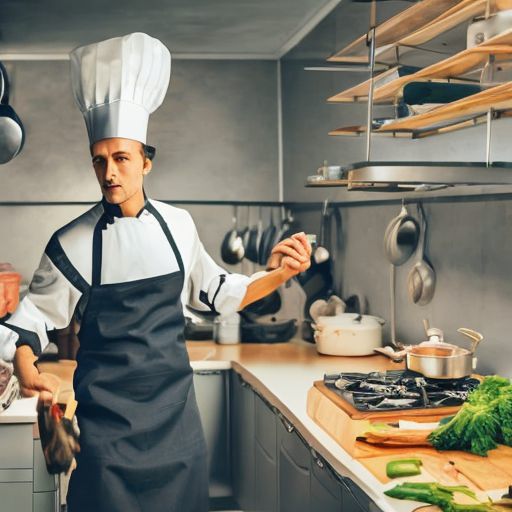}
            & \includegraphics[width=\guidimgw]{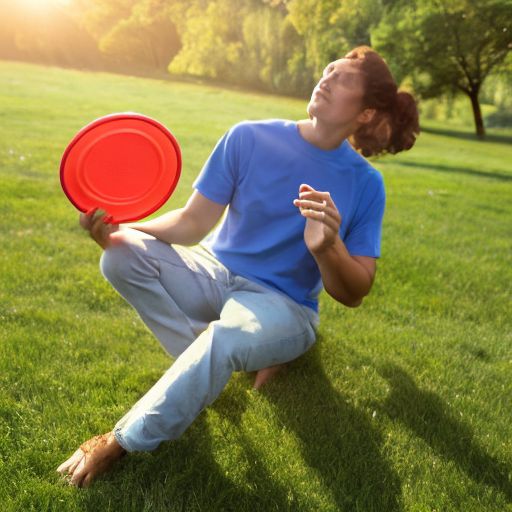}
            & \includegraphics[width=\guidimgw]{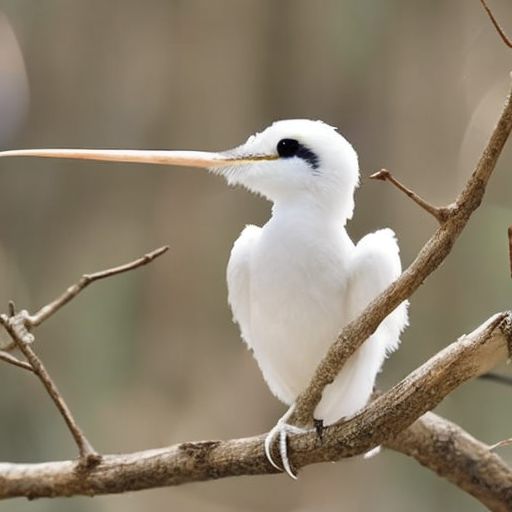}
            & \includegraphics[width=\guidimgw]{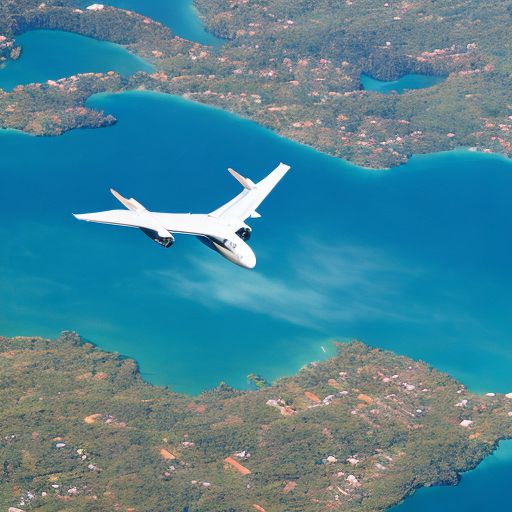}
            & \includegraphics[width=\guidimgw]{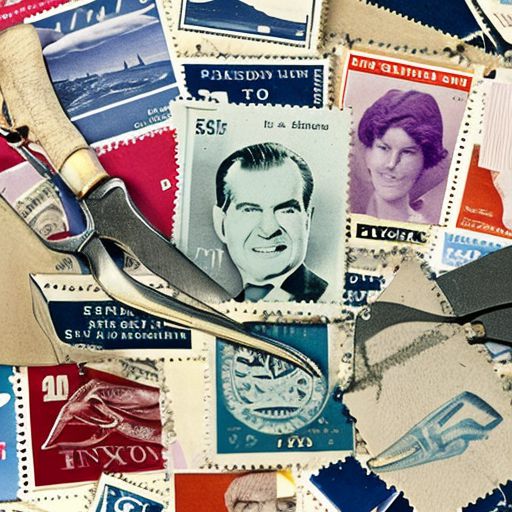}
            & \includegraphics[width=\guidimgw]{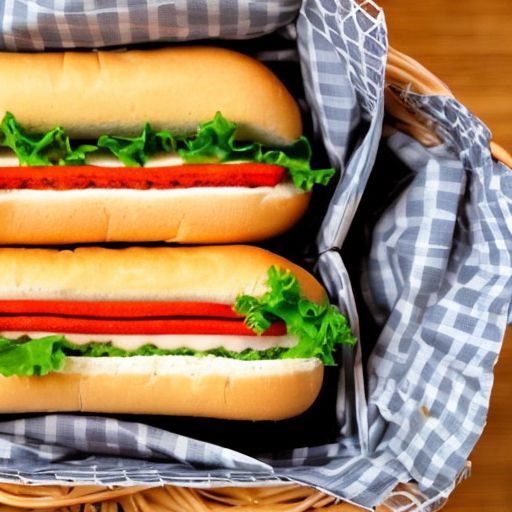} \\ \addlinespace[3pt]
            & \parbox{\guidimgw}{\centering ``A man is cooking in a crowded kitchen''}
            & \parbox{\guidimgw}{\centering ``A person is in the grass holding a frisbee''}
            & \parbox{\guidimgw}{\centering ``A small white bird with a long beak on a branch''}
            & \parbox{\guidimgw}{\centering ``A plane flies over water with two islands nearby''}
            & \parbox{\guidimgw}{\centering ``Scissors and a pile of postage stamps topped by one of President Nixon''}
            & \parbox{\guidimgw}{\centering ``A couple of sub sandwiches are in basket''} \\
    \end{tabular}
    \caption{Qualitative comparison between CFG, CFG++, and Ours}
    \label{fig:guidance_qualitative}
\end{figure}

\subsection{Generation with Guidance Correction}
\label{sec:guidance}
\subsubsection{Experimental Setup.}
As in the previous sections, we use Stable Diffusion v2.1-base \cite{Rombach2022} as the backbone with $T=50$ timesteps.
We generate 30k images using text prompts from the MS-COCO 2014 validation set \cite{Lin2014} and evaluate using FID \cite{Martin2017} against the corresponding images and CLIP Score \cite{Radford2021} against the text prompts.
The former evaluates image quality, while the latter evaluates text-image alignment.
We use CFG \cite{Ho2021} as the baseline and CFG++ \cite{Chung2025} as a comparison method.
We set the finite difference to $h=10^{-4}$ and the weight to $\lambda=0.1$ for our method and use the default hyperparameters of CFG++ from the original paper.

\subsubsection{Results.}
We summarize the results in \cref{tab:guidance_results}.
Regardless of the CFG scale $w$, our metric-based guidance correction consistently improves FID while maintaining comparable CLIP Score.
CFG++ does not consistently improve FID, suggesting sensitivity to hyperparameter settings.
The fact that CLIP Score is preserved across all methods indicates that our guidance correction improves image quality without sacrificing text-image alignment, consistent with the design of our objective in \cref{eq:corrected_guidance}.

\Cref{fig:guidance_qualitative} shows representative examples using prompts from the same dataset.
CFG and CFG++ produce cartoonish images in the leftmost column and overly emphasize some objects (i.e., frisbee and beak) in the second and third columns.
Our guidance correction produces more natural images with better details, which confirms that it effectively prevents off-manifold perturbations back onto the data manifold.

\subsubsection{Distillation with Guidance Correction.}
When our guidance correction modifies only the teacher's guidance term, it can be incorporated into distillation at no additional inference cost for the student.
We follow the protocol of the latent consistency model (LCM) \cite{Luo2023} with default settings in diffusers~\cite{von-platen-etal-2022-diffusers}, including AdamW optimization~\cite{Loshchilov2019}; see Appendix~\ref{appendix:hyperparameters} for details.
After distillation, we generate images with $T=4$ steps and evaluate them in the same way as above.

\Cref{tab:distillation_results} summarizes the results averaged over five runs.
For each run, we match the random seed between CFG and our method; our method improves FID in all five runs ($p < 0.05$, one-sided exact binomial test) while maintaining comparable CLIP Score.

\section{Conclusion}
\label{sec:conclusion}
We proposed a Riemannian metric on the noise space of diffusion models, derived from the Jacobian of the score function.
Building on the spectral structure of this Jacobian, which separates tangent and normal directions of the data manifold, our metric encourages geodesics to stay within or run parallel to the manifold without any additional training or architectural modifications.
We validated this metric from two complementary angles: geodesic-based interpolation confirmed that it captures global manifold geometry, and guidance correction confirmed that its local tangent--normal decomposition is accurate enough to project off-manifold perturbations back onto the manifold.
Beyond these two tasks, we believe this metric provides a geometric foundation for broader manifold-aware operations in diffusion models, such as image editing, noise-space clustering, and extension to flow-matching and energy-based models.
Validating the metric on more recent architectures remains important future work.

\section*{Acknowledgements}
This study was partly supported by JST BOOST (JPMJBY24H0), JST CREST (JPMJCR24Q5), and JST ASPIRE (JPMJAP2329), Japan, and achieved through the use of large-scale computer systems at the D3Center, Osaka University.

\clearpage

\appendix
\crefalias{section}{appendix}
\numberwithin{figure}{section}
\numberwithin{table}{section}

\section{Preliminaries}
\label{sec:preliminaries}

In this section, we briefly review the necessary background on Riemannian geometry and diffusion models.

\subsection{Riemannian Geometry}
\label{appendix:riemannian_geometry}

\subsubsection{Riemannian Metric.}
We adopt the notions in Lee \cite{Lee2019}.
Let $\gM$ be a smooth manifold.
A \emph{Riemannian metric} $g$ on $\gM$ is a smooth covariant 2-tensor field such that, at every point $p \in \gM$, the tensor $g_p$ defines an inner product on the tangent space $T_p\gM$.
In other words, $g$ is symmetric and positive-definite; at $p\in\gM$,
\begin{equation}\textstyle
    \begin{aligned}
        g_p(v,w) & =g_p(w,v),\quad                             \\
        g_p(v,v) & \ge 0 \ \text{for all } v\in \gT_p\gM,\quad \\
        g_p(v,v) & =0 \Leftrightarrow v=0.
    \end{aligned}
\end{equation}
By identifying $g_p$ with an inner product, we write
\begin{equation}\textstyle
    \langle v, w\rangle_g \coloneq g_p(v,w) \quad \text{for any } v,w\in T_p\gM.
\end{equation}
A \emph{Riemannian manifold} is the pair $(\gM, g)$.

Let $(x^1,\dots,x^D)$ be smooth local coordinates in a neighborhood of $p\in\gM$.
Then, the coordinate basis for $T_p\gM$ is $\bigl(\frac{\partial}{\partial x^1}|_p,\dots,\frac{\partial}{\partial x^D}|_p\bigr)$, where $\frac{\partial}{\partial x^i}$ is the $i$-th coordinate vector field.
Tangent vectors $v,w\in T_p{\gM}$ can be expressed as $v=\sum_{i=1}^D v^i\,\frac{\partial}{\partial x^i}|_p$ and $w=\sum_{i=1}^D w^i\,\frac{\partial}{\partial x^i}|_p$, respectively.
The matrix notation $G_p$ of $g$ at $p$ consists of $(i,j)$-elements
\begin{equation}
    \label{eq:metric_tensor}
    \textstyle
    g_{ij}(p)=g_p\left(\frac{\partial}{\partial x^i}|_p,\frac{\partial}{\partial x^j}|_p\right)=\left\langle \frac{\partial}{\partial x^i}|_p,\frac{\partial}{\partial x^j}|_p \right\rangle_g
\end{equation}
for $i,j=1,2,...,D$.
This is symmetric and positive definite.
The Euclidean metric is represented by the identity matrix $I$.
The inner product $\langle v,w\rangle_g $ of two tangent vectors $v,w\in T_p\gM$ at $p$ is given by
\begin{equation}
    \label{eq:inner_product_with_metric_tensor}
    \textstyle
    g_p(v,w)=\sum_{i=1}^D\sum_{j=1}^D g_{ij}(p)v^iw^j=v^TG_pw.
\end{equation}

\subsubsection{Geodesics.}
The length of a tangent vector $v\in T_p\gM$ is given by $|v|_g \coloneq \sqrt{\langle v,v\rangle_g}$.
For a smooth curve $\gamma:[0,1]\to \gM,\,u\mapsto \gamma(u)$, its length is
\begin{equation}
    \label{eq:length_functional}
    \begin{aligned}
        \textstyle L[\gamma]
         & \textstyle \coloneq \int_0^1 |\gamma'(u)|_g \,\dee u = \int_0^1 \sqrt{\langle \gamma'(u), \gamma'(u)\rangle_g}\,\dee u \\
         & \textstyle = \int_0^1 \sqrt{\gamma'(u)^\top G_{\gamma(u)} \gamma'(u)} \,\dee u.
    \end{aligned}
\end{equation}
A \emph{geodesic} is a curve that locally minimizes length; intuitively, it is a locally shortest path between two points.
It is often more convenient to work with the energy functional:
\begin{equation}
    \label{eq:energy_functional}
    \textstyle
    E[\gamma] = \frac{1}{2} \int_0^1 |\gamma'(u)|_g^2 \,\dee u
    = \frac{1}{2} \int_0^1 \langle \gamma'(u), \gamma'(u) \rangle_g \,\dee u.
\end{equation}
Any constant-speed geodesic is a critical point of the energy functional.

A geodesic can be obtained by solving the geodesic equation, a second-order ODE \cite{Lee2019}, which requires computation of $O(D^3)$ in general and is not feasible in high-dimensional spaces.

\subsection{Diffusion Models}
\label{appendix:diffusion_models}

\subsubsection{Forward Process.}
Let $x_0 \in \mathbb{R}^D$ be a data sample.
The forward process is defined as a Markov chain which adds Gaussian noise at each timestep $t=1,\dots,T$ recursively:
\begin{equation}
    \label{eq:forward}
    \begin{aligned}
        \textstyle q({x_t}|{x}_{t-1})
         & \textstyle =\mathcal{N}\left({x}_t;\sqrt{1-\beta_t}{x}_{t-1},\beta_t I\right)                                                     \\
         & \textstyle =\mathcal{N}\Big({x}_t;\sqrt{\frac{\alpha_t}{\alpha_{t-1}}}{x}_{t-1}, \Big(1-\frac{\alpha_t}{\alpha_{t-1}}\Big)I\Big),
    \end{aligned}
\end{equation}
where $\{\beta_t\}_{t=1}^T$ is a scheduled variance, $I$ is the identity matrix in $\mathbb{R}^D$, and $\alpha_t=\prod_{s=1}^t(1-\beta_s)$.
$x_t$ becomes progressively more corrupted by noise as $t$ increases, and $x_T$ is nearly an isotropic Gaussian distribution.

\subsubsection{Denoising Process.}
The generation process of diffusion models is referred to as the denoising process (or reverse process), which inverts the forward process by iteratively denoising a noisy sample ${x}_T\sim\mathcal{N}(0,I)$ backward in time from $t=T$ to $t=0$ and obtaining a clean sample $x_0$.
Namely, a reverse Markov chain $p_t({x}_{t-1}|{x}_t;\theta)$ is constructed as
\begin{equation}
    \label{eq:backward}
    \textstyle
    {x}_{t-1}=\frac{1}{\sqrt{1-\beta_t}}\left({x_t}-\frac{\beta_t}{\sqrt{1-\alpha_t}}{\epsilon}_\theta({x}_t, t)\right)+\sigma_t{z}_t,
\end{equation}
with a trainable noise predictor ${\epsilon}_\theta$, where ${z}_t \sim \mathcal{N}(0,I)$, and $\sigma_t^2 = \beta_t$ is a variance at timestep $t$.
The noise predictor ${\epsilon}_\theta({x}_t,t)$ is trained by minimizing the objective:
\begin{equation}
    \label{eq:ddpm_obj}
    \textstyle
    \mathcal{L}(\theta)=\mathbb{E}_{{x},{\epsilon}_t, t}\big[\|{\epsilon}_t-{\epsilon}_\theta({x}_t, t)\|_2^2\big],
\end{equation}
where ${\epsilon}_t\sim\mathcal{N}(0,I)$ is the noise added during the forward process at timestep $t$.

\subsubsection{Denoising Diffusion Implicit Models and Inversion.}
Denoising diffusion implicit models (DDIMs) \cite{Song2021a} modify \cref{eq:forward} to be a non-Markovian process $q(x_{t-1}|x_t, x_0) = \mathcal{N}\left(x_{t-1}; \sqrt{\alpha_{t-1}}x_0 + \sqrt{1 - \alpha_{t-1} - \sigma_t^2}\frac{x_t - \sqrt{\alpha_t}x_0}{\sqrt{1-\alpha_t}}, \sigma_t^2 I\right)$.
Then, the denoising process becomes
\begin{equation}
    \label{eq:backward_ddim}
    \begin{aligned}
        \textstyle {x}_{t-1} = \sqrt{\alpha_{t-1}}\left(\frac{{x_t}-\sqrt{1-\alpha_{t}}\,{\epsilon}_\theta({x}_t,t)}{\sqrt{\alpha_t}}\right) + \sqrt{1-\alpha_{t-1}-\sigma_t^2}\cdot{\epsilon}_\theta({x}_t,t) + \sigma_t{z}_t,
    \end{aligned}
\end{equation}
where $\sigma_t=\eta\sqrt{(1-\alpha_{t-1})/(1-\alpha_t)}\sqrt{1-\alpha_t/\alpha_{t-1}}$.
$\eta\in[0,1]$ controls the stochasticity: $\eta=1$ recovers DDPM, while $\eta=0$ yields a deterministic update.
The forward process in \cref{eq:forward} can also be modified accordingly.
Then, we can deterministically map a clean sample $x_0$ to a noisy sample $x_t$, operate interpolations in the noise space at timestep $t$, and then map it back to a clean sample $x_0$; this procedure is often referred to as DDIM Inversion (see below).

\subsubsection{Formulation as Stochastic Differential Equations.}
As the timestep size approaches zero, the forward process can also be formulated as a stochastic differential equation (SDE) \cite{Song2021b}.
The denoising process is the corresponding reverse-time SDE that depends on the score function $s_\theta(x_t,t)\coloneq\nabla_{x_t}\log p_t(x_t;\theta)$, where $p_t(x_t;\theta)$ denotes the density of $x_t$ at time $t$.
Notably, the noise predictor $\epsilon_\theta$ is closely tied to the score function as:
\begin{equation}
    \label{eq:score_noise_relation}
    \textstyle
    s_\theta(x_t,t)=\nabla_{{x}_t} \log p_t({x}_t;\theta) \approx -{\epsilon}_\theta({x}_t, t)/\sqrt{1-\alpha_t}.
\end{equation}
Thus, learning the noise predictor $\epsilon_\theta$ is essentially learning the score function $s_\theta$.
The following discussion about the score function $s_\theta$ applies to the noise predictor $\epsilon_\theta$ as well, up to a known scale.

\subsubsection{Conditioning and Guidance.}
We can condition the score function $s_\theta$ on a text prompt $c$, writing $s_\theta({x}_t, t, c)$, to guide the generation process \cite{Rombach2022}.
The actual implementation depends on the architecture of the score function $s_\theta$.
Classifier-free guidance (CFG) amplifies this condition to make generated images more faithful to text prompts \cite{Ho2021}, by replacing the score function as:
\begin{equation}
    \label{eq:cfg}
    \begin{aligned}
        \textstyle \tilde{s}_\theta({x}_t, t, c) = (w+1)\,s_\theta({x}_t, t, c) - w\,s_\theta({x}_t, t, \varnothing),
    \end{aligned}
\end{equation}
where $s_\theta({x}_t, t, c)$ and $s_\theta({x}_t, t, \varnothing)$ are conditional and unconditional score functions, respectively.
The negative prompt suppresses concepts specified by a complementary prompt \cite{Rombach2022} as:
\begin{equation}
    \label{eq:negative_prompt}
    \begin{aligned}
        \textstyle \tilde{s}_\theta({x}_t, t, c, c_\text{neg}) = s_\theta({x}_t, t, c) - w_\text{neg}\,s_\theta({x}_t, t, c_\text{neg}),
    \end{aligned}
\end{equation}
where $c_\text{neg}$ is a complementary prompt describing the concepts to be suppressed.

It is worth noting that when CFG and negative prompts are used simultaneously, the unconditional score function in \cref{eq:cfg} is often replaced with the score function $s_\theta({x}_t, t, c_\text{neg})$ conditioned on the negative prompt $c_\text{neg}$.

\subsubsection{DDIM Inversion.}
Naive encoding of an original image is to add Gaussian noise as in the forward process $q({x_t}\mid{x}_{t-1})$, which is stochastic and often yields poor reconstructions.
To accurately invert the denoising process and recover the specific noise map associated with a given image, \emph{DDIM Inversion} \cite{Mokady2023} is widely used.
The deterministic version ($\eta=0$) of DDIM can be regarded as an ordinary differential equation (ODE) solved by the Euler method \cite{Song2021a,Song2021b}.
In the limit of infinitesimally small timesteps, the ODE is invertible.

Concretely, setting $\sigma_t=0$ in \cref{eq:backward_ddim} gives
\begin{equation}
    \label{ddim_update_abb}
    \begin{aligned}
        x_{t-1} & = a_t x_t + b_t \epsilon_\theta(x_t, t)       \\
                & = x_t+(a_t-1)x_t+b_t \epsilon_\theta(x_t, t),
    \end{aligned}
\end{equation}
where $a_t=\sqrt{\alpha_{t-1}/\alpha_t}$ and $b_t=-\sqrt{\alpha_{t-1}(1-\alpha_t)/\alpha_t}+\sqrt{\vphantom{/}1-\alpha_{t-1}}$.
This can be viewed as an ODE with the time derivative $(a_t-1)x_t+b_t \epsilon_\theta(x_t, t)$ solved by the Euler method with the unit step size.
With a sufficiently small timestep size,
\begin{equation}
    \label{inversion}
    x_t = \frac{x_{t-1} - b_t \epsilon_\theta(x_t, t)}{a_t} \approx \frac{x_{t-1} - b_t \epsilon_\theta(x_{t-1}, t)}{a_t},
\end{equation}
since $\epsilon_\theta(x_{t},t)\approx\epsilon_\theta(x_{t-1},t)$.
The deterministic forward process iteratively applies the update rule in \cref{inversion} to a sample $x_0$ from $t=0$ to $t$ and obtains the noisy image $x_t$, from which the deterministic denoising process reconstructs the original $x_0$ up to numerical errors.
This inversion procedure substantially improves the fidelity of reconstructions and subsequent interpolations.

\section{Implications of Proposed Method}
\label{appendix:proposition}

\subsection{Implication of \Cref{thm:equivalence}}
\label{appendix:proposition_1}

When the score function $s_\theta$ is exact, it is the gradient $\nabla_{x_t}\log p_t(x_t;\theta)$ of the log-density $\log p_t(x_t;\theta)$, and its Jacobian $J_{x_t}$ equals the Hessian, given by $J_{x_t}=\nabla_{x_t}\nabla_{x_t}\log p_t(x_t;\theta)$, which is symmetric.
In this idealized case, its eigenvectors form an orthonormal basis of the noise space $\R^D$.
We divide these eigenvectors into a basis for the tangent space $\gT_{x_t}\gM_t$, $\{ v_i \}_{i=1}^d$ (with small eigenvalues $\lambda_i$), and a basis for the normal space $\gN_{x_t}\gM_t$, $\{ v_j \}_{j=d+1}^D$ (with large eigenvalues $\lambda_j$).
These spaces are orthogonal complements of each other, and the tangent space $\gT_{x_t}\R^D$ to the noise space $\R^D$ at $x_t$ can be decomposed into their direct sum, $\gT_{x_t}\R^D = \gT_{x_t}\gM_t \oplus \gN_{x_t}\gM_t$.
Any tangent vector $v \in \gT_{x_t}\R^D$ is uniquely decomposed as $v = v_\gT + v_\gN$, where $v_\gT \in \gT_{x_t}\gM_t$ and $v_\gN \in \gN_{x_t}\gM_t$.
The squared Jacobian-vector product $\|J_{x_t} v\|_2^2$ can be expanded as:
\begin{equation}
    \textstyle
    \begin{aligned}
        \|J_{x_t} v\|_2^2 & = \|J_{x_t}(v_\gT + v_\gN)\|_2^2                                                                 \\
                          & = \|J_{x_t} v_\gT\|_2^2 + \|J_{x_t} v_\gN\|_2^2 + 2\langle J_{x_t} v_\gT, J_{x_t} v_\gN \rangle.
    \end{aligned}
\end{equation}
Due to the orthogonality of the eigenspaces, the cross term $\langle J_{x_t} v_\gT, J_{x_t} v_\gN \rangle$ vanishes, and we have
\begin{equation}
    \begin{aligned}
        \textstyle \|J_{x_t} v_\gT\|_2^2 & = \textstyle \sum_{i=1}^d \lambda_i^2 \langle v, v_i \rangle^2 \approx 0,                               \\
        \textstyle \|J_{x_t} v_\gN\|_2^2 & = \textstyle \sum_{j=d+1}^D \lambda_j^2 \langle v, v_j \rangle^2 \gg 0 \quad (\text{if } v_\gN \neq 0).
    \end{aligned}
\end{equation}
Hence, minimizing the squared Jacobian–vector product $\|J_{x_t} v\|_2^2$ (under a fixed Euclidean norm of $v$) is dominated by minimizing the normal-space component $\|J_{x_t} v_\gN\|_2^2$, and essentially encourages the vector $v$ to lie in the tangent space $\gT_{x_t}\gM_t$.

In practice, diffusion models learn the score function $s_\theta$ directly, so its Jacobian $J_{x_t}$ is not exactly symmetric, and its eigenvectors need not be exactly orthogonal to each other.
Even then, minimizing $\|J_{x_t} v\|_2^2$ still suppresses the component in the subspace spanned by the large right singular vectors and amplifies the component spanned by the small right singular vectors; \Cref{thm:equivalence} continues to hold in this generalized sense.

\subsubsection{Regularization and Alternative Construction.}
To ensure positive definiteness, one can also consider a regularized metric $G_{x_t} = J_{x_t}^\top J_{x_t}+\lambda I$ for a small $\lambda>0$.
However, preliminary experiments using Stable Diffusion v2.1-base \cite{Rombach2022} showed that this does not significantly affect the results, so we use the simpler form in \cref{eq:jacobian_metric}.

One might instead use $J_{x_t}$ directly as a metric, rather than $J_{x_t}^\top J_{x_t}$, creating a Hessian manifold, as it corresponds to the approximated Hessian of the log-probability.
However, due to the non-convexity of the log-probability, the metric may be indefinite and thus pseudo-Riemannian, with which geodesics are no longer characterized as shortest paths.

Note that, at $t=0$, the score function $s_\theta$ is typically not well trained outside the data manifold, making it nontrivial to find meaningful paths.

\subsection{Intuitive Explanation}
\label{appendix:intuitive_explanation}

\subsubsection{Geodesic-Based Interpolation.}
\Cref{fig:illustration_interp} illustrates the geodesic-based interpolation pipeline.
Given two clean samples $x_0^\sss{0}$ and $x_0^\sss{1}$, we map them via DDIM Inversion to the noisy manifold $\gM_\tau$ embedded in the noise space $\R^D$ at $t=\tau$.
In this space, we initialize a discrete path by SLERP and optimize the intermediate points to minimize the energy $E$ (\cref{eq:energy_jacobian_diff}), which measures the total squared change in the score function along the path.
Minimizing $E$ makes the path geodesic, i.e., straight in the score space $\mathcal{S}$.
The optimized noisy samples are then denoised back to $\gM_0$, yielding the interpolated clean samples.

\subsubsection{Metric-Based Guidance Correction.}
\Cref{fig:illustration_guidance} illustrates the metric-based guidance correction mechanism.
Omitting the step size $\eta_t$ for brevity, the denoising update maps a noisy sample $x_t$ on $\gM_t$ to $x_{t-1}$ on $\gM_{t-1}$ via the score function $s_\theta$.
CFG adds a term $\Delta s$ to steer the generation, but $\Delta s$ can push the sample away from $\gM_{t-1}$, leading to artifacts such as oversaturation.
Our correction finds $\Delta\hat{s}^*$ (\cref{eq:corrected_guidance}) by balancing two objectives: staying close to the guided sample in Euclidean distance and to the unguided sample $x_{t-1}$ under our metric $g$.
This is equivalent to projecting $\Delta s$ onto the tangent space $\gT_{x_t}\gM_t$ induced by the score Jacobian $J_{x_t}$.
In practice, we use $\gT_{x_t}\gM_t$ instead of $\gT_{x_{t-1}}\gM_{t-1}$ for efficiency, since the manifold changes smoothly between adjacent timesteps.
The corrected guidance keeps the sample closer to $\gM_{t-1}$, improving image quality.

\section{Details of Methods}
\label{appendix:comparison}

\subsection{Comparison Methods}
\label{appendix:comparison_methods}

\subsubsection{Linear Interpolation.}
Once samples are noised via DDIM Inversion, one can perform straightforward linear interpolation (LERP) \cite{Ho2020}, by treating the noise space at fixed time $\tau>0$ as a linear latent space.
Given samples $x_0^\sss{0}$ and $x_0^\sss{1}$ in the data space, the deterministic forward process obtains their noised versions $x_\tau^\sss{0}$ and $x_\tau^\sss{1}$ at $\tau$, respectively.
A linear interpolation in that space is given by
\begin{equation}
    {x}_\tau^\sss{u}=(1-u)x_\tau^\sss{0}+u x_\tau^\sss{1},
\end{equation}
where $u\in[0,1]$ is the interpolation parameter.
Then, one applies the deterministic denoising process from $t=\tau$ back to $t=0$ to obtain a sequence of interpolated images $x_0^\sss{u}$ in the data space.

\subsubsection{Spherical Linear Interpolation.}
An alternative is spherical linear interpolation (SLERP) \cite{Song2021a}, which finds the shortest path on a sphere in the noise space:
\begin{equation}
    x_\tau^\sss{u}=\frac{\sin((1-u)\theta)}{\sin{(\theta)}}x_\tau^\sss{0}+\frac{\sin(u\theta)}{\sin{(\theta)}}x_\tau^\sss{1}
\end{equation}
where $\theta=\arccos\left({\frac{(x_\tau^\sss{0})^\top x_\tau^\sss{1}}{\|x_\tau^\sss{0}\| \|x_\tau^\sss{1}\|}}\right)$.
This procedure preserves the norms of the noisy samples $x_\tau^\sss{u}$, yielding more natural interpolations than LERP.
Note that SLERP assumes that $x_\tau^\sss{0}$ and $x_\tau^\sss{1}$ are drawn from a normal distribution, which holds only for a sufficiently large $\tau$ (typically, $\tau=T$).
Nonetheless, SLERP is often applied at moderate $\tau$.

\subsubsection{FIM-based Riemannian Metric.}
Information geometry shows that the Fisher score can induce a Fisher Information Matrix (FIM) on a statistical manifold.
Inspired by this, Azeglio and Bernardo~\cite{Azeglio2025} constructed an FIM-like metric for diffusion models using their score function $s_\theta(x_t,t)$ (also called Stein score).
\begin{equation}
    \label{eq:FIM-based_metric}
    \textstyle
    g_{x_t}(v, w) \coloneq v^\top (\lambda s_{\theta}(x_t,t) s_\theta(x_t,t)^\top + I) w,
\end{equation}
where $\lambda>0$ balances the metric based solely on the FIM ($\lambda \to \infty$) to the Euclidean metric ($\lambda \to 0$).
In our experiments, we set $\lambda=1{,}000$ following the original work.
The FIM-based term $s_{\theta}(x_t,t) s_\theta(x_t,t)^\top$ is rank-1 and encourages geodesics to be orthogonal to the direction of the score function $s_\theta$.
However, even with this metric, the geodesics lie on the tangent space $\gT_{x_t}\gM_t$ of the data manifold $\gM_t$ only when the codimension of the data manifold $\gM_t$ is one.
Otherwise, the geodesics may still deviate from the data manifold $\gM_t$, which is the typical case for real-world data.

The remaining comparison methods (NAO \cite{Samuel2023}, NoiseDiffusion \cite{Zheng2024}, and GeodesicDiffusion \cite{Yu2025}) are used with their default settings from official implementations.

\begin{table}[t]
    \centering
    \scriptsize
    \caption{Comparison of computational costs for interpolation methods}
    \label{tab:comp_cost}
    \begin{tabular}{@{}l@{\hspace{1.2em}}l@{\hspace{1.2em}}c@{}}
        \toprule
        \textbf{Method}                                             & \textbf{Type} & \textbf{Computational Cost}              \\
        \midrule
        LERP / SLERP / NoiseDiff \cite{Zheng2024}                   & closed-form   & $2IS + (N-1) L + (N+1)GS$                \\
        NAO \cite{Samuel2023}                                       & iterative     & $2IS + K (N+1) L + (N+1)GS$              \\
        GeoDiff \cite{Yu2025} / FIM-based \cite{Azeglio2025} / Ours & iterative     & $2IS + K (N-1) S + (N+1)GS$              \\
        \bottomrule
                                                                    &               & (Inversion + Interpolation + Generation) \\
    \end{tabular}
\end{table}

\subsection{Computational Cost of Interpolation Methods}\label{appendix:comp_cost}

The computational cost varies across interpolation methods.
Some methods obtain interpolated images as closed-form solutions, whereas others obtain them through an iterative optimization process.
We summarize the computational costs of different methods in \cref{tab:comp_cost}, where each variable is defined as follows.
\begin{itemize}
    \item $N$: The number of discretization points for the interpolation path, resulting in $N-1$ interpolated images.
    \item $S$: The cost of one evaluation of the score function $s_\theta$ (or noise predictor $\epsilon_\theta$), which is the most computationally intensive part of diffusion models.
    \item $I$: The number of evaluations of the score function $s_\theta$ during a single DDIM inversion, which maps a clean image to a noise image.
    \item $G$: The number of evaluations of the score function $s_\theta$ for generation (i.e., during a denoising process), which maps a noise image to a clean image.
    \item $K$: The number of optimization iterations required for iterative methods to converge.
    \item $L$: The cost of a simple latent space operation (e.g., vector addition and scaling for LERP), where $L \ll S$.
\end{itemize}
The total cost is broken down into three main stages: (1) DDIM inversion by mapping the two endpoints in the data space (at $t=0$) to the noisy samples in the noise space at $\tau>0$, (2) interpolation of those endpoints with $N-1$ intermediate points, and (3) generation by mapping all $N+1$ points from the noise space to the data space.

LERP, SLERP and NoiseDiffusion \cite{Zheng2024} are closed-form methods, which use simple, non-iterative computation for the interpolation.
NAO \cite{Samuel2023} is an iterative method that does not use the score function $s_\theta$ for interpolation; thus, its cost is relatively low.
GeodesicDiffusion \cite{Yu2025}, FIM-based metric \cite{Azeglio2025} and ours are iterative methods that use the score function $s_\theta$; thus, they have a higher computational cost than closed-form methods.

\subsection{Prompt Adjustment}
\label{appendix:prompt_adjustment}
To improve the quality of interpolations, we adopt the prompt adjustment proposed by Yu et al.~\cite{Yu2025}.
Internally in Stable Diffusion v2.1-base \cite{Rombach2022}, a text prompt $c$ is first encoded into a text embedding $z$ using CLIP \cite{Radford2021}.
To better align the text embedding $z$ with a given pair of images $x_0^\sss{0}$ and $x_0^\sss{1}$, we adjust the text embedding $z$ in a similar way to textual inversion \cite{Gal2023}.
Namely, the text embedding $z$ is updated to minimize the DDPM loss in \cref{eq:ddpm_obj} for 500 iterations for image interpolation and 1{,}000 iterations for video frame interpolation.
We use AdamW optimizer \cite{Loshchilov2019} with a learning rate of 0.005.

Also following Yu et al.~\cite{Yu2025}, we do not use CFG (i.e., set $w=0$ in \cref{eq:cfg}) but use the following negative prompt $c_\text{neg}$ with $w_\text{neg}=1$: ``A doubling image, unrealistic, artifacts, distortions, unnatural blending, ghosting effects, overlapping edges, harsh transitions, motion blur, poor resolution, low detail.''

See Appendix~\ref{appendix:ablation} for an ablation study on the effect of this adjustment.

\section{Experimental Setup}
\label{appendix:experimental}

This section provides details of the experimental setup in \cref{sec:experiments}.
Image and video interpolations were conducted on a single NVIDIA RTX A6000 GPU.
Metric-based guidance corrections and distillations were conducted on NVIDIA A100 and H200 GPUs, respectively.

\subsection{Synthetic 2D Dataset}
\label{appendix:synthetic}

\subsubsection{Dataset.}
We construct a two-dimensional C-shaped distribution as follows.
We start with an axis-aligned ellipse with semi-axes $1.0$ (along $x_1$) and $1.2$ (along $x_2$).
To open the ``C'', we remove all points in a $\pm 30^\circ$ wedge centered on the positive $x_1$-axis.
We then add isotropic Gaussian perturbations with standard deviation $0.001$ per coordinate to each point.
From the resulting distribution, we draw 100{,}000 samples.

\subsubsection{Network.}
The noise predictor $\epsilon_\theta$ is composed of three linear layers of hidden width 512 with SiLU activation functions \cite{Elfwing2017}.
The network takes a tuple of a data point $x$ and a normalized time $t$ as input.
We set the number of steps to $T=1{,}000$ for training and $T=50$ for generation.
We trained this network for 1{,}000 epochs using the AdamW optimizer \cite{Loshchilov2019} with a batch size of 512.
The learning rate follows cosine annealing \cite{Loshchilov2017}, decaying from $10^{-3}$ to $0$ without restarts.
For stability, we apply gradient-norm clipping with a threshold of $1.0$.

\subsubsection{Implementation Details.}
In \cref{fig:illust} (left), we visualize the interpolation between $x_0^\sss{0} = (0.0, 1.15)$ and $x_0^\sss{1} = (-0.8, -0.6)$ with $N=100$ discretization points.
Comparison methods include Linear Interpolation (LERP) \cite{Ho2020}, Spherical Linear Interpolation (SLERP) \cite{Song2021a}, and density-based interpolation based on the metric proposed in Yu et al.~\cite{Yu2025}.
We used the DDIM Scheduler \cite{Song2021a} and operated in the noise space at $\tau=0.02T=1$.
For our method and the density-based interpolation, we find the geodesic paths by minimizing the energy functional $E[\gamma]$.
Both paths are initialized using SLERP and updated using Adam optimizer \cite{Kingma2015} for 1{,}000 iterations with a learning rate of $10^{-4}$.

\subsection{Datasets for Image Interpolation}
\label{appendix:image_interpolation_datasets}

MorphBench \cite{Kaiwen2023} consists of pairs of images obtained via image editing, with 24 pairs in the animation subset (MB(A)) and 66 pairs in the metamorphosis subset (MB(M)).
For MB(A), both endpoints $x_0^\sss{0}$ and $x_0^\sss{1}$ share the same text prompt, which we used as the condition $c$.
For MB(M), each endpoint has a distinct prompt; following DiffMorpher~\cite{Kaiwen2023} and GeoDiff~\cite{Yu2025}, we linearly interpolate the text embeddings of the two prompts to obtain the condition $c$ at each interpolation step.
\Cref{tab:morphbench_prompts_a,tab:morphbench_prompts_m} list all prompts for the 90 pairs used in our experiments, covering a diverse range of subjects across both within-category and cross-category scenarios.

Animal Faces-HQ \cite{Choi2022b} is a dataset of high-resolution images of animal faces.
From this dataset, we randomly selected 50 pairs of dog images and 50 pairs of cat images with LPIPS below 0.6 to ensure semantic similarity.
We used the text prompts ``a photo of a dog'' for dog images and ``a photo of a cat'' for cat images.

CelebA-HQ \cite{Karras2018} is a high-resolution dataset of celebrity faces.
We randomly sampled 50 male pairs and 50 female pairs, again with LPIPS less than 0.6, and conditioned on ``a photo of a man'' and ``a photo of a woman,'' respectively.

\subsection{Hyperparameters for Distillation}
\label{appendix:hyperparameters}

In the protocol of the latent consistency model (LCM) \cite{Luo2023}, the teacher performs a single DDIM step from $x_t$ to $x_{t-1}$, and the student is trained to minimize the discrepancy between its own prediction of $x_0$ from $x_t$ and the target prediction of $x_0$ from $x_{t-1}$.
The teacher's DDIM step uses CFG at a guidance scale sampled uniformly from $[5, 15]$.
The student is parameterized with low-rank adaptation (LoRA) of rank 64.
It is trained for 1{,}000 iterations on the Conceptual 12M dataset with a batch size of 96, using the Huber loss ($c=0.001$) and AdamW optimizer~\cite{Loshchilov2019} at a learning rate of $10^{-4}$ with gradient clipping at 1.0.
These are all default settings in diffusers~\cite{von-platen-etal-2022-diffusers}.
After distillation, we generate images with $T=4$ steps and evaluate them.

\begin{table}[t]
    \centering
    \fontsize{6}{7.2}\selectfont
    \caption{Text prompts used for MorphBench Animation (MB(A), 24 pairs)}
    \label{tab:morphbench_prompts_a}
    \setlength{\tabcolsep}{4pt}
    \begin{tabular}{p{2.2cm} p{6.7cm}}
        \toprule
        Name & Prompt \\
        \midrule
        \rowcolor{gray!10} alpaca & an alpaca \\
        boy\_and\_girl & boy and girl holding hands \\
        \rowcolor{gray!10} bulb & light bulb \\
        cake & pancakes with berries \\
        \rowcolor{gray!10} chair & a blue mid-century armchair \\
        cocktail & a blue frozen cocktail with lime garnish \\
        \rowcolor{gray!10} counterfeit & an anime girl in kimono, forest \\
        david & Michelangelo's David, marble bust \\
        \rowcolor{gray!10} dog\_jump & German Shepherd jumping on grass \\
        dog\_sit & German Shepherd sitting on grass \\
        \rowcolor{gray!10} drag\_dog & fluffy dog wearing round sunglasses \\
        drag\_girl & an anime girl eating in a cafe \\
        \rowcolor{gray!10} drag\_realgirl & a girl with curly hair \\
        drag\_sculp & an ancient greek marble head sculpture \\
        \rowcolor{gray!10} duck & a goose head, open beak \\
        fuji & Mount Fuji with snow \\
        \rowcolor{gray!10} imagic\_cake & a cake on a wooden stand \\
        laser\_sculp & a vaporwave bust of David, iridescent, neon \\
        \rowcolor{gray!10} masa\_bird & a bird on a branch \\
        masa\_boy & a boy with glasses \\
        \rowcolor{gray!10} masa\_oldman & a wrinkled old man \\
        mushroom & a mushroom on mossy forest floor \\
        \rowcolor{gray!10} scene & an aerial view of a coastline, water, forest \\
        woman & a classical painting of a woman with a turban \\
        \bottomrule
    \end{tabular}
    \\
    \scriptsize
    Both endpoints share the same prompt. All prompts are prefixed with ``a photo of.''
\end{table}

\begin{table}[p]
    \centering
    \fontsize{6}{7.2}\selectfont
    \caption{Text prompts used for MorphBench Metamorphosis (MB(M), 66 pairs)}
    \label{tab:morphbench_prompts_m}
    \setlength{\tabcolsep}{4pt}
    \begin{tabular}{p{2.2cm} p{6.7cm}}
        \toprule
        Name & Prompt \\
        \midrule
        \rowcolor{gray!10} anakin & a young man with blue eyes \\
        \rowcolor{gray!10}  & darth vader from star wars \\
        arya & a woman holding a sword \\
         & a girl with long hair and a sword \\
        \rowcolor{gray!10} boy01 & a young boy in a hoodie \\
        \rowcolor{gray!10}  & a young boy with curly hair \\
        boy02 & a young boy in a hoodie \\
         & a young boy with blonde hair \\
        \rowcolor{gray!10} boy03 & a young boy in a hoodie \\
        \rowcolor{gray!10}  & a young boy in the rain \\
        boy12 & a young boy with curly hair \\
         & a young boy with blonde hair \\
        \rowcolor{gray!10} boy13 & a young boy with curly hair \\
        \rowcolor{gray!10}  & a young boy in the rain \\
        boy23 & a young boy with blonde hair \\
         & a young boy in the rain \\
        \rowcolor{gray!10} boy\_girl\_0 & a young boy looking at the camera \\
        \rowcolor{gray!10}  & a young girl with blue eyes \\
        boy\_girl\_1 & a young boy with curly hair \\
         & a young woman with green eyes \\
        \rowcolor{gray!10} boy\_girl\_2 & a young boy with blonde hair \\
        \rowcolor{gray!10}  & a young girl with blonde hair \\
        boy\_girl\_3 & a young boy in the rain \\
         & a young girl with her hair blowing in the wind \\
        \rowcolor{gray!10} cake\_burger & a stack of pancakes with berries on top \\
        \rowcolor{gray!10}  & a hamburger on a black plate \\
        carr\_cars & a red maserati convertible parked on a pier \\
         & the mercedes amg gt3 race car \\
        \rowcolor{gray!10} cars\_van & the mercedes amg gt3 race car \\
        \rowcolor{gray!10}  & a silver mini cooper on a gray floor \\
        castle & mont saint michel in normandy, france \\
         & mont saint michel at dusk \\
        \rowcolor{gray!10} cat\_rabbit & an orange and white cat looking up at the window \\
        \rowcolor{gray!10}  & a rabbit sitting on the grass \\
        chair01 & a blue chair sitting on a hardwood floor \\
         & a green couch in front of a white wall \\
        \rowcolor{gray!10} chair02 & a blue chair sitting on a hardwood floor \\
        \rowcolor{gray!10}  & an office chair with wooden legs \\
        chair03 & a blue chair sitting on a hardwood floor \\
         & a white chair next to a table \\
        \rowcolor{gray!10} chair12 & a green couch in front of a white wall \\
        \rowcolor{gray!10}  & an office chair with wooden legs \\
        chair13 & a green couch in front of a white wall \\
         & a white chair next to a table \\
        \rowcolor{gray!10} chair23 & an office chair with wooden legs \\
        \rowcolor{gray!10}  & a white chair next to a table \\
        dog & a dog sitting in front of a brown background \\
         & a black and white dog sitting in the grass \\
        \rowcolor{gray!10} dog\_wolf & a corgi puppy sitting on an orange background \\
        \rowcolor{gray!10}  & a gray wolf looking at the camera \\
        gandalf & an old man with a pipe in his mouth \\
         & santa claus with a beard and glasses \\
        \rowcolor{gray!10} girl & a woman in a black jacket \\
        \rowcolor{gray!10}  & a young woman with brown hair \\
        girl01 & a young girl with blue eyes \\
         & a young woman with green eyes \\
        \rowcolor{gray!10} girl02 & a young girl with blue eyes \\
        \rowcolor{gray!10}  & a young girl with blonde hair \\
        girl03 & a young girl with blue eyes \\
         & a young girl with her hair blowing in the wind \\
        \rowcolor{gray!10} girl12 & a young woman with green eyes \\
        \rowcolor{gray!10}  & a young girl with blonde hair \\
        girl13 & a young woman with green eyes \\
         & a young girl with her hair blowing in the wind \\
        \rowcolor{gray!10} girl23 & a young girl with blonde hair \\
        \rowcolor{gray!10}  & a young girl with her hair blowing in the wind \\
        \bottomrule
    \end{tabular}\\
    {\normalsize (continued on the next page)}
\end{table}

\begin{table}[p]
    \centering
    \fontsize{6}{7.2}\selectfont
    \setlength{\tabcolsep}{4pt}
    \begin{tabular}{p{2.2cm} p{6.7cm}}
        \toprule
        Name & Prompt \\
        \midrule
        house\_left & an old house in the middle of a field \\
         & a barn with a red roof \\
        \rowcolor{gray!10} house\_lr & an old house in the middle of a field \\
        \rowcolor{gray!10}  & an old wooden cabin in the middle of a grassy field \\
        house\_right & an old wooden cabin in the middle of a grassy field \\
         & a red house covered in snow \\
        \rowcolor{gray!10} jay & asian man with black hair \\
        \rowcolor{gray!10}  & an asian man in a suit and tie \\
        leo & a young man with short hair \\
         & a man in a suit and tie \\
        \rowcolor{gray!10} lion\_tiger & a lion resting on a rock \\
        \rowcolor{gray!10}  & a tiger with its mouth open \\
        man\_van & a man wearing a black suit and a black hat \\
         & a painting of a man wearing a hat \\
        \rowcolor{gray!10} mona\_pearl & the mona lisa by leonardo da vinci \\
        \rowcolor{gray!10}  & a girl with a pearl earring \\
        Musk\_Feifei & a man in a suit and tie \\
         & an asian woman in a blue shirt \\
        \rowcolor{gray!10} Musk\_Obama & a man in a suit and tie \\
        \rowcolor{gray!10}  & president barack obama \\
        Musk\_Trump & a man in a suit and tie \\
         & a man wearing a suit and tie \\
        \rowcolor{gray!10} obama\_putin & president barack obama \\
        \rowcolor{gray!10}  & russian president vladimir putin \\
        Obama\_Trump & president barack obama \\
         & donald trump \\
        \rowcolor{gray!10} pika & a pikachu on a white background \\
        \rowcolor{gray!10}  & a pikachu with a lightning bolt coming out of its mouth \\
        raccoon & a raccoon looking at the camera \\
         & a raccoon with blue eyes \\
        \rowcolor{gray!10} realdog\_cat & a golden retriever puppy \\
        \rowcolor{gray!10}  & an orange and white cat \\
        red\_car & a red maserati convertible parked on a pier \\
         & a small red car parked on the side of the road \\
        \rowcolor{gray!10} scream & home alone 2 double pack \\
        \rowcolor{gray!10}  & the scream by edvard munch \\
        sculp & a marble head with curly hair \\
         & a bust of a man with curly hair \\
        \rowcolor{gray!10} snow\_mountain & a mountain range covered in snow \\
        \rowcolor{gray!10}  & the milky way in the night sky over a mountain \\
        taylor\_yifei & taylor swift with red lipstick \\
         & an asian woman with long hair \\
        \rowcolor{gray!10} thanos & thanos in fortnite \\
        \rowcolor{gray!10}  & superman flying through the air \\
        thu\_mit & the entrance to a university building \\
         & people sitting on the grass in front of a large building \\
        \rowcolor{gray!10} Trump\_Biden & a man wearing a suit and tie \\
        \rowcolor{gray!10}  & joe biden in front of the american flag \\
        vangogh & a painting of vincent van gogh \\
         & a painting of vincent van gogh \\
        \rowcolor{gray!10} van\_jeep & a silver mini cooper on a gray floor \\
        \rowcolor{gray!10}  & the mercedes g - class pickup truck \\
        van\_mona & a painting of a man wearing a hat \\
         & the mona lisa by leonardo da vinci \\
        \rowcolor{gray!10} van\_pearl & a painting of vincent van gogh \\
        \rowcolor{gray!10}  & a painting of a girl with a pearl earring \\
        van\_self & a painting of vincent van gogh \\
         & a painting of a man wearing a hat \\
        \rowcolor{gray!10} wave & a large wave in the ocean \\
        \rowcolor{gray!10}  & the great wave off kanagawa \\
        wc & a restroom sign with a man and woman \\
         & a man and woman dancing in a yellow dress \\
        \rowcolor{gray!10} whitehouse\_church & the U.S. Capitol building in Washington, DC \\
        \rowcolor{gray!10}  & the cathedral in florence, italy \\
        wolf\_tiger & a gray wolf looking at the camera \\
         & a tiger with its mouth open \\
        \bottomrule
    \end{tabular}
    \\
    \scriptsize
    \raggedright Each endpoint has a distinct prompt. All prompts are prefixed with ``a photo of.''
    The first and second rows of each pair show the prompts for $x_0^\sss{0}$ and $x_0^\sss{1}$, respectively.
\end{table}

\section{Additional Results}\label{appendix:additional_results}
\subsection{Additional Qualitative Results for Image Interpolation}
\begin{figure}[p]
    \centering
    \scriptsize
    \setlength{\tabcolsep}{1pt}
    \renewcommand{\arraystretch}{0}
    \begin{tabular}{rc}
        \raisebox{2.9mm}{\textcolor{BLUE}{LERP}}      & {\includegraphics[width=10.4cm]{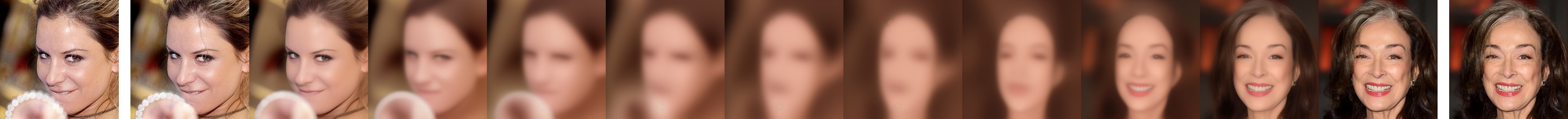}}              \\ \addlinespace[1.5pt]
        \raisebox{2.9mm}{\textcolor{GREEN}{SLERP}}    & {\includegraphics[width=10.4cm]{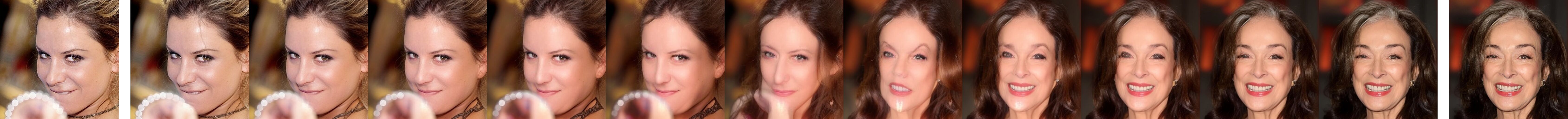}}             \\ \addlinespace[1.5pt]
        \raisebox{2.9mm}{\textcolor{PURPLE}{NAO}}     & {\includegraphics[width=10.4cm]{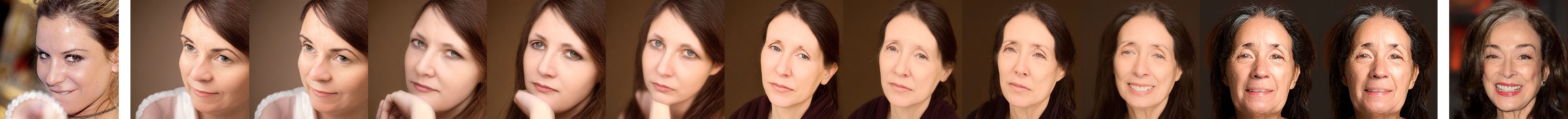}}               \\ \addlinespace[1.5pt]
        \raisebox{2.9mm}{\textcolor{ORANGE}{NoiseDiff}} & {\includegraphics[width=10.4cm]{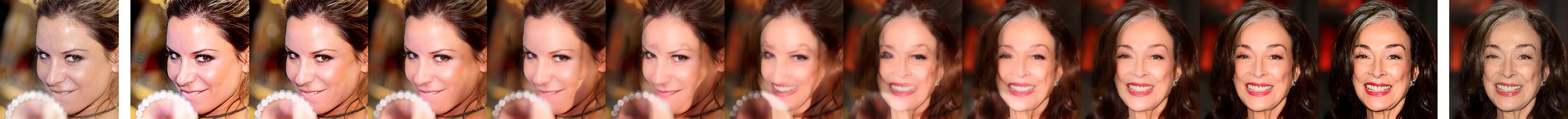}}         \\ \addlinespace[1.5pt]
        \raisebox{2.9mm}{\textcolor{VERMILION}{GeoDiff}} & {\includegraphics[width=10.4cm]{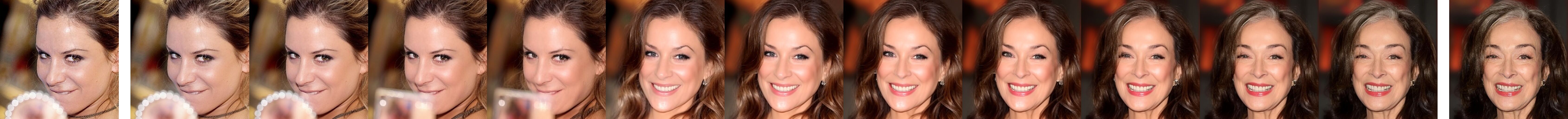}}           \\ \addlinespace[1.5pt]
        \raisebox{2.9mm}{\textcolor{GRAY}{FIM-based}} & {\includegraphics[width=10.4cm]{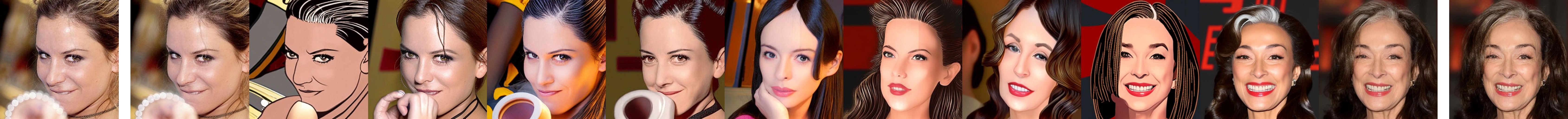}}               \\ \addlinespace[1.5pt]
        \raisebox{2.9mm}{\textcolor{BLACK}{Ours}}     & {\includegraphics[width=10.4cm]{assets/results/image/female_ours.jpg}}              \\ \addlinespace[1.5pt]
                                                      & \includegraphics[scale=0.946]{assets/utils/imagelabel.pdf}                             \\ \addlinespace[3.0pt]
                                                      & (c) CelebA-HQ (Female)                                                                                     \\ \addlinespace[6.0pt]
        \raisebox{2.9mm}{\textcolor{BLUE}{LERP}}      & {\includegraphics[width=10.4cm]{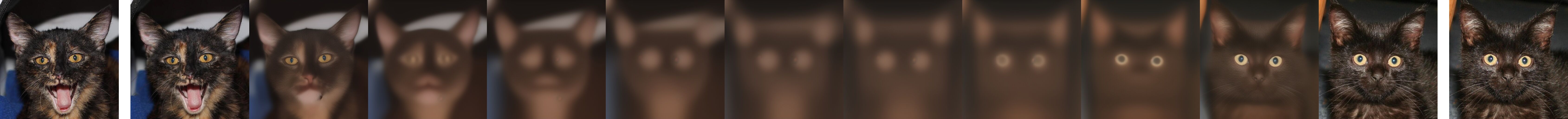}}          \\ \addlinespace[1.5pt]
        \raisebox{2.9mm}{\textcolor{GREEN}{SLERP}}    & {\includegraphics[width=10.4cm]{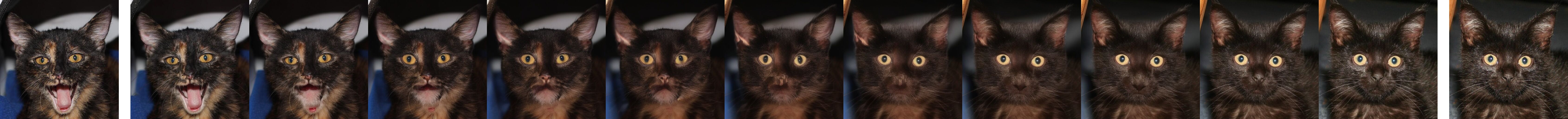}}         \\ \addlinespace[1.5pt]
        \raisebox{2.9mm}{\textcolor{PURPLE}{NAO}}     & {\includegraphics[width=10.4cm]{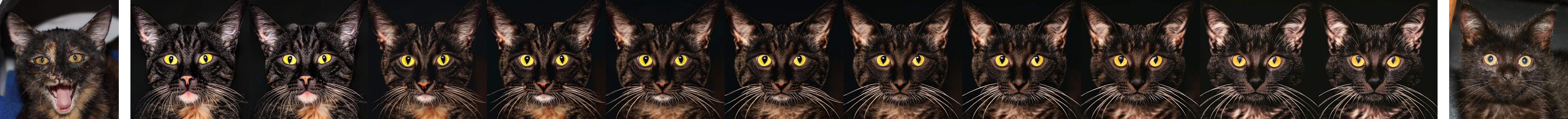}}           \\ \addlinespace[1.5pt]
        \raisebox{2.9mm}{\textcolor{ORANGE}{NoiseDiff}} & {\includegraphics[width=10.4cm]{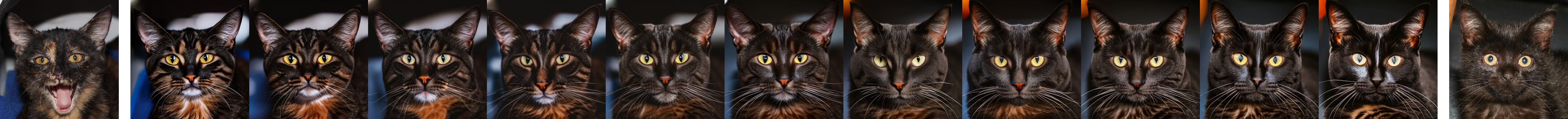}}     \\ \addlinespace[1.5pt]
        \raisebox{2.9mm}{\textcolor{VERMILION}{GeoDiff}} & {\includegraphics[width=10.4cm]{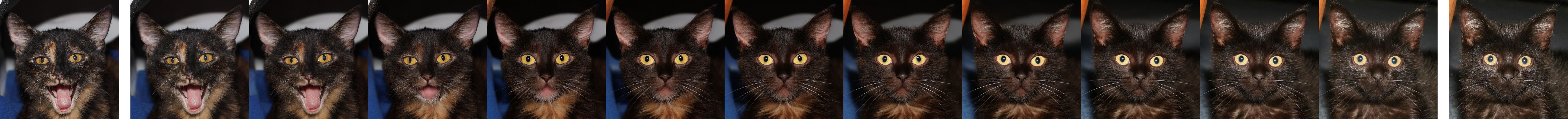}}       \\ \addlinespace[1.5pt]
        \raisebox{2.9mm}{\textcolor{GRAY}{FIM-based}} & {\includegraphics[width=10.4cm]{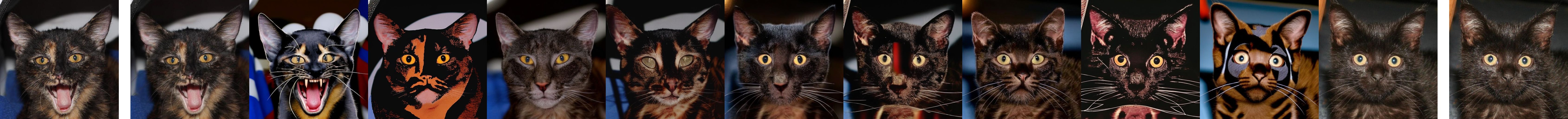}}           \\ \addlinespace[1.5pt]
        \raisebox{2.9mm}{\textcolor{BLACK}{Ours}}     & {\includegraphics[width=10.4cm]{assets/results/image/cat_ours.jpg}}          \\ \addlinespace[1.5pt]
                                                      & \includegraphics[scale=0.946]{assets/utils/imagelabel.pdf}                      \\ \addlinespace[3.0pt]
                                                      & (d) Animal Faces-HQ (Cat)
    \end{tabular}
    \caption{Examples of interpolated image sequences.
        The leftmost and rightmost images are the given endpoints $x_0^\sss{0}$ and $x_0^\sss{1}$, and the middle images are the interpolated results $\{\hat{x}_0^\sss{u}\}$ for $u\in[0,1]$.
        See also \cref{fig:results_qualitative}.
    }
    \label{fig:results_qualitative_appendix}
\end{figure}

\Cref{fig:results_qualitative_appendix} provides additional qualitative examples of image interpolation, which complement \cref{fig:results_qualitative} in the main body.

\begin{figure}[p]
    \centering
    \scriptsize
    \setlength{\tabcolsep}{1pt}
    \renewcommand{\arraystretch}{0}
    \begin{tabular}{rc}
        \raisebox{2.9mm}{\textcolor{BLUE}{LERP}}      & {\includegraphics[width=10.4cm]{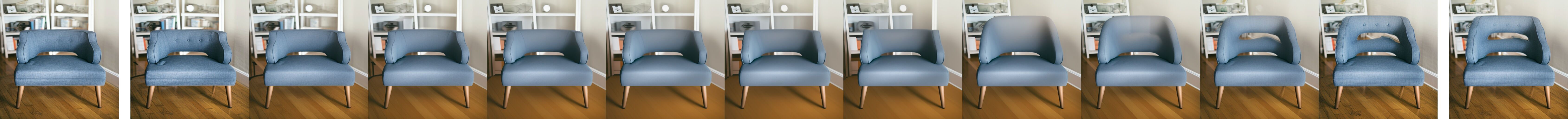}}              \\ \addlinespace[1.5pt]
        \raisebox{2.9mm}{\textcolor{GREEN}{SLERP}}    & {\includegraphics[width=10.4cm]{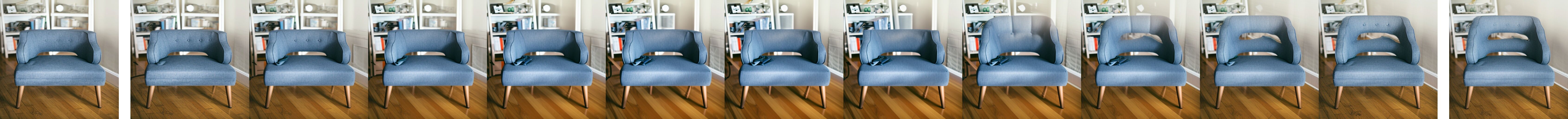}}             \\ \addlinespace[1.5pt]
        \raisebox{2.9mm}{\textcolor{PURPLE}{NAO}}     & {\includegraphics[width=10.4cm]{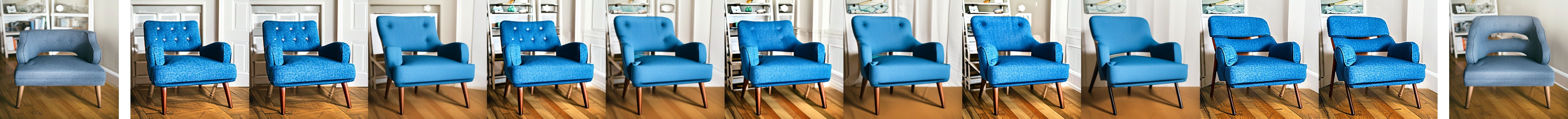}}               \\ \addlinespace[1.5pt]
        \raisebox{2.9mm}{\textcolor{ORANGE}{NoiseDiff}} & {\includegraphics[width=10.4cm]{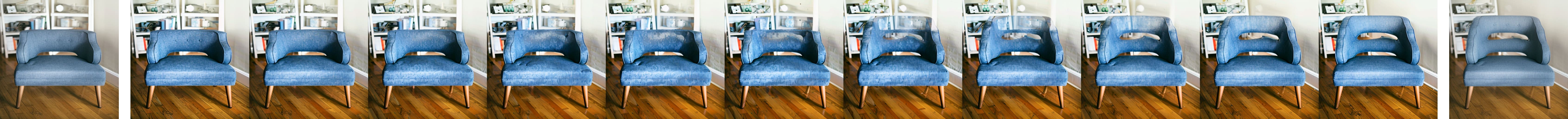}}         \\ \addlinespace[1.5pt]
        \raisebox{2.9mm}{\textcolor{VERMILION}{GeoDiff}} & {\includegraphics[width=10.4cm]{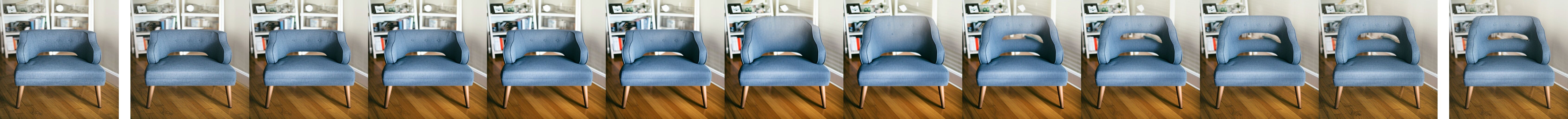}}           \\ \addlinespace[1.5pt]
        \raisebox{2.9mm}{\textcolor{GRAY}{FIM-based}} & {\includegraphics[width=10.4cm]{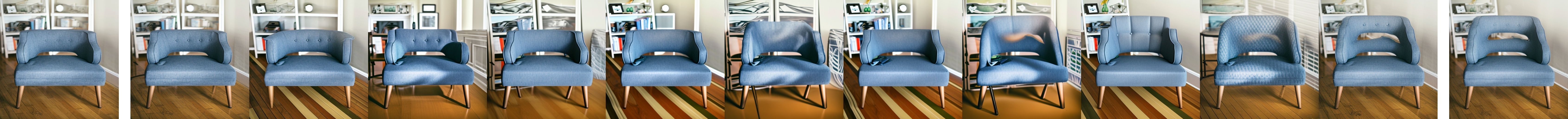}}               \\ \addlinespace[1.5pt]
        \raisebox{2.9mm}{\textcolor{BLACK}{Ours}}     & {\includegraphics[width=10.4cm]{assets/results/image/chair_ours.jpg}}              \\ \addlinespace[1.5pt]
                                                      & \includegraphics[scale=0.946]{assets/utils/imagelabel.pdf}                           \\ \addlinespace[3.0pt]
                                                      & (a) MorphBench (Animation)                                                                      \\ \addlinespace[6.0pt]
        \raisebox{2.9mm}{\textcolor{BLUE}{LERP}}      & {\includegraphics[width=10.4cm]{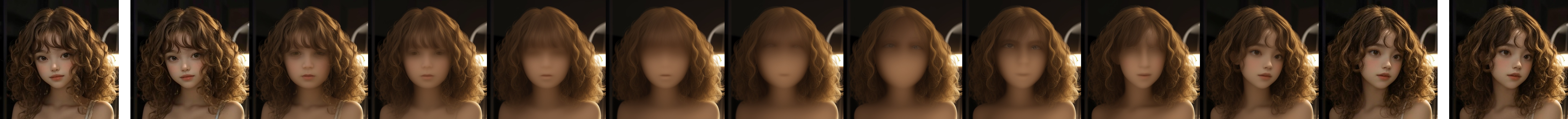}}      \\ \addlinespace[1.5pt]
        \raisebox{2.9mm}{\textcolor{GREEN}{SLERP}}    & {\includegraphics[width=10.4cm]{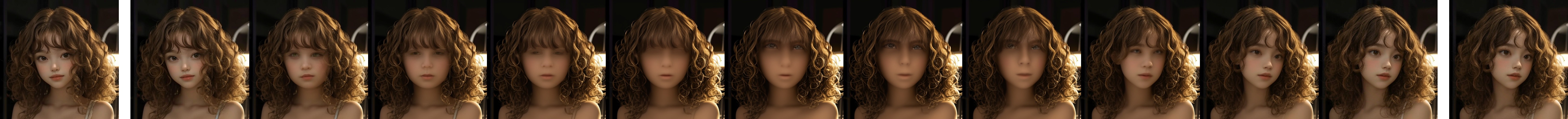}}     \\ \addlinespace[1.5pt]
        \raisebox{2.9mm}{\textcolor{PURPLE}{NAO}}     & {\includegraphics[width=10.4cm]{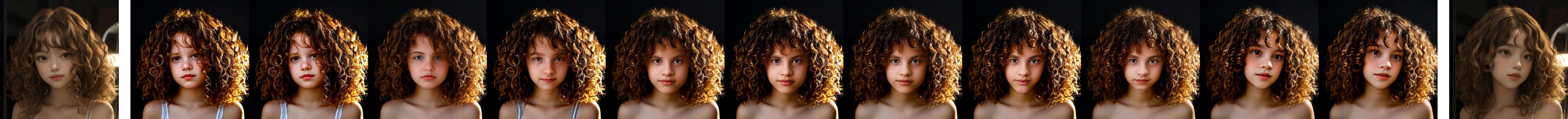}}       \\ \addlinespace[1.5pt]
        \raisebox{2.9mm}{\textcolor{ORANGE}{NoiseDiff}} & {\includegraphics[width=10.4cm]{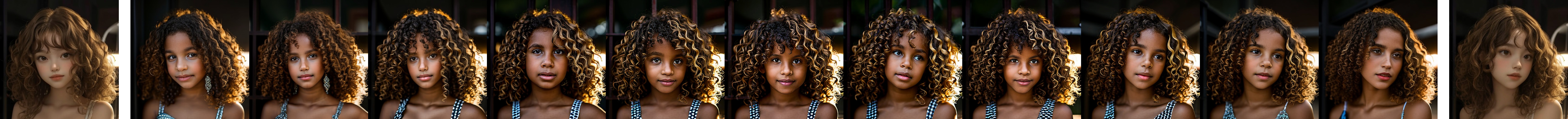}} \\ \addlinespace[1.5pt]
        \raisebox{2.9mm}{\textcolor{VERMILION}{GeoDiff}} & {\includegraphics[width=10.4cm]{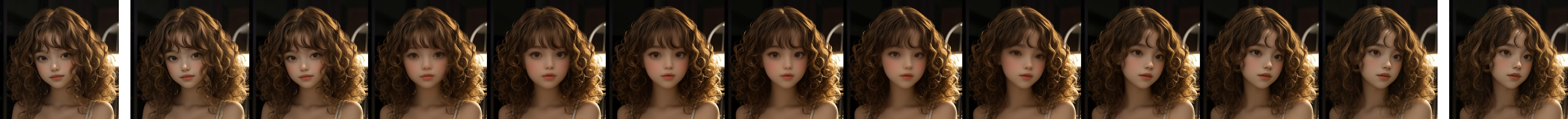}}   \\ \addlinespace[1.5pt]
        \raisebox{2.9mm}{\textcolor{GRAY}{FIM-based}} & {\includegraphics[width=10.4cm]{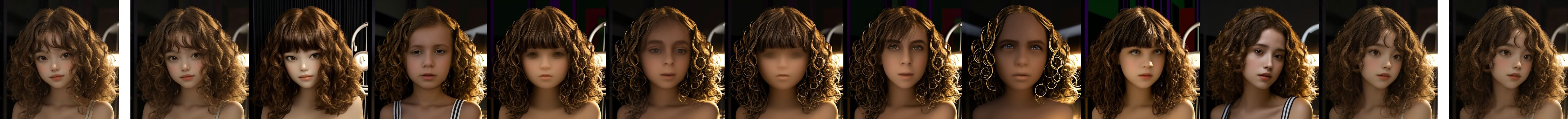}}       \\ \addlinespace[1.5pt]
        \raisebox{2.9mm}{\textcolor{BLACK}{Ours}}     & {\includegraphics[width=10.4cm]{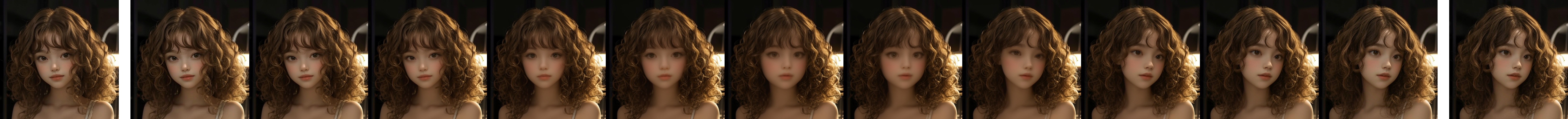}}      \\ \addlinespace[1.5pt]
                                                      & \includegraphics[scale=0.946]{assets/utils/imagelabel.pdf}                           \\ \addlinespace[3.0pt]
                                                      & (b) MorphBench (Animation)                                                                      \\ \addlinespace[6.0pt]
    \end{tabular}\\
    {\normalsize (continued on the next page)}
\end{figure}

\subsection{Evaluation with Different Pre-trained Weights}
In the main body, we examined Stable Diffusion v2.1-base \cite{Rombach2022} as the backbone.
To verify the generality of our method, we conducted additional experiments using Stable Diffusion v2.0-base \cite{Rombach2022}.
While the model architecture remains identical, the dataset and pipeline used for training are different, so the learned data manifold may also be different.
We applied all interpolation methods to the v2.0-base checkpoint without modifying any hyperparameters from the default settings reported in the main body.

The quantitative results are summarized in \cref{tab:image_interp_v2.0,tab:video_interp_v2.0}, where the performance scores for v2.0 are slightly worse than v2.1 due to the lack of the improvements introduced in the v2.1 release.
Nonetheless, we observe similar trends to those in \cref{tab:main_results} for v2.1-base; our method achieves consistent improvements across all metrics and datasets except for PDV.
This demonstrates that our approach effectively captures the geometric structure in the diffusion model, regardless of the specific model version.

\bigskip
\hfill(continued on page \pageref{appendix:ablation})

\clearpage

\begin{table}[p]
    \centering
    \footnotesize
    \caption{Results on image interpolation by Stable Diffusion v2.0-base}
    \label{tab:image_interp_v2.0}
    \setlength{\tabcolsep}{5.0pt}
    \begin{tabular}{@{}l rrrr rrrr@{}}
        \toprule
                        & \multicolumn{4}{c}{\textbf{PPL} $\downarrow$} & \multicolumn{4}{c}{\textbf{PDV} $\downarrow$} \\
        \cmidrule(lr){2-5} \cmidrule(lr){6-9}
        \textbf{Method} & \multicolumn{1}{c}{MB(A)} & \multicolumn{1}{c}{MB(M)} & \multicolumn{1}{c}{CA} & \multicolumn{1}{c}{AF} & \multicolumn{1}{c}{MB(A)} & \multicolumn{1}{c}{MB(M)} & \multicolumn{1}{c}{CA} & \multicolumn{1}{c}{AF} \\
        \midrule
        \textcolor{BLUE}{LERP}           & 0.876              & 1.820              & 1.463              & 1.895              & 0.056              & 0.135              & 0.100              & 0.166              \\
        \textcolor{GREEN}{SLERP}         & 0.642              & 1.082              & 0.702              & 0.867              & 0.030              & \fst{0.055\sigtwo} & \fst{0.032}        & \fst{0.023}        \\
        \textcolor{PURPLE}{NAO}          & 2.840              & 4.315              & 1.976              & 2.077              & 0.159              & 0.160              & 0.142              & 0.333              \\
        \textcolor{ORANGE}{NoiseDiff}    & 1.832              & 2.038              & 1.239              & 1.731              & 0.078              & 0.088              & 0.059              & 0.079              \\
        \textcolor{VERMILION}{GeoDiff}   & \snd{0.389}        & \snd{1.016}        & \snd{0.659}        & \snd{0.816}        & \snd{0.023}        & 0.068              & 0.038              & \snd{0.026}        \\
        \textcolor{GRAY}{FIM-based}      & 3.365              & 4.561              & 4.271              & 5.351              & 0.145              & 0.186              & 0.176              & 0.198              \\
        \midrule
        \textcolor{BLACK}{Ours}          & \fst{0.382}        & \fst{0.974\sigone} & \fst{0.632\sigtwo} & \fst{0.761\sigtwo} & \fst{0.020}        & \snd{0.070}        & \snd{0.034}        & \fst{0.023}        \\
        \bottomrule
        \toprule
                        & \multicolumn{4}{c}{\textbf{FID} $\downarrow$} & \multicolumn{4}{c}{\textbf{RE} $\downarrow$ $\scriptscriptstyle (\times 10^{-3})$} \\
        \cmidrule(lr){2-5} \cmidrule(lr){6-9}
        \textbf{Method} & \multicolumn{1}{c}{MB(A)} & \multicolumn{1}{c}{MB(M)} & \multicolumn{1}{c}{CA} & \multicolumn{1}{c}{AF} & \multicolumn{1}{c}{MB(A)} & \multicolumn{1}{c}{MB(M)} & \multicolumn{1}{c}{CA} & \multicolumn{1}{c}{AF} \\
        \midrule
        \textcolor{BLUE}{LERP}           & 85.31              & 123.83             & 97.42              & 127.02             & 0.404              & 0.403              & 1.030              & 2.086              \\
        \textcolor{GREEN}{SLERP}         & 63.44              & 50.32              & 37.70              & \snd{25.43}        & 0.404              & 0.403              & 1.030              & 2.086              \\
        \textcolor{PURPLE}{NAO}          & 130.46             & 101.06             & 82.78              & 126.38             & 39.391             & 45.087             & 26.683             & 5.736              \\
        \textcolor{ORANGE}{NoiseDiff}    & 105.45             & 73.25              & 59.08              & 55.76              & 7.792              & 7.819              & 4.875              & 10.022             \\
        \textcolor{VERMILION}{GeoDiff}   & \snd{29.06}        & \snd{38.60}        & \snd{36.00}        & 27.10              & \snd{0.211}        & \snd{0.290}        & \snd{0.962}        & \snd{2.005}        \\
        \textcolor{GRAY}{FIM-based}      & 93.10              & 82.23              & 72.81              & 64.24              & 0.404              & 0.403              & 1.030              & 2.086              \\
        \midrule
        \textcolor{BLACK}{Ours}          & \fst{28.99}        & \fst{37.18}        & \fst{33.63}        & \fst{21.16}        & \fst{0.183\sigtwo} & \fst{0.210\sigtwo} & \fst{0.902\sigtwo} & \fst{1.961\sigtwo} \\
        \bottomrule
    \end{tabular}\\
    \raggedright\textsuperscript{\tiny\!*} and \textsuperscript{\tiny\!*\!*} indicate that the improvement over the second-best method is statistically significant at the 0.01 and 0.001 levels, respectively, according to a one-sided exact binomial test ($H_0:\ p = 0.5$).
\end{table}

\begin{table}[p]
    \centering
    \footnotesize
    \setlength{\tabcolsep}{5.0pt}
    \caption{Results on video frame interpolation by Stable Diffusion v2.0-base}
    \label{tab:video_interp_v2.0}
    \begin{tabular}{l rrr rrr}
        \toprule
                                      & \multicolumn{3}{c}{\textbf{MSE} $\downarrow$ $\scriptscriptstyle (\times 10^{-3})$} & \multicolumn{3}{c}{\textbf{LPIPS} $\downarrow$} \\
        \cmidrule(lr){2-4} \cmidrule(lr){5-7}
        \textbf{Method}               & \multicolumn{1}{c}{DAVIS} & \multicolumn{1}{c}{Human} & \multicolumn{1}{c}{RE10K} & \multicolumn{1}{c}{DAVIS} & \multicolumn{1}{c}{Human} & \multicolumn{1}{c}{RE10K} \\
        \midrule
        \textcolor{BLUE}{LERP}        & \snd{11.915}              & 4.612                     & 6.373                     & 0.589                     & 0.380                     & 0.379                     \\
        \textcolor{GREEN}{SLERP}      & 15.859                    & 6.045                     & 5.972                     & 0.485                     & 0.319                     & 0.298                     \\
        \textcolor{PURPLE}{NAO}       & 72.883                    & 75.128                    & 80.230                    & 0.825                     & 0.719                     & 0.813                     \\
        \textcolor{ORANGE}{NoiseDiff} & 33.977                    & 33.055                    & 18.253                    & 0.518                     & 0.515                     & 0.418                     \\
        \textcolor{VERMILION}{GeoDiff} & 13.585                    & \snd{3.444}               & \snd{4.753}               & \snd{0.343}               & \snd{0.184}               & \snd{0.181}               \\
        \textcolor{GRAY}{FIM-based}   & 27.479                    & 12.320                    & 12.211                    & 0.516                     & 0.382                     & 0.363                     \\
        \midrule
        \textcolor{BLACK}{Ours}       & \fst{9.193\sigtwo}       & \fst{2.152\sigtwo}       & \fst{2.580\sigtwo}       & \fst{0.335}              & \fst{0.176\sigone}       & \fst{0.169\sigone}       \\
        \bottomrule
    \end{tabular}\\
    \textsuperscript{\tiny\!*} and \textsuperscript{\tiny\!*\!*} indicate the statistical significance in the same manner as \cref{tab:image_interp_v2.0}.
\end{table}

\clearpage

\subsection{Ablation Studies}
\label{appendix:ablation}

\subsubsection{Prompt Adjustment.}
We adopt the prompt adjustment of GeoDiff~\cite{Yu2025} (see Appendix~\ref{appendix:prompt_adjustment} for details) to better align the text embedding with the images.
\Cref{tab:ablation_prompt_adjustment} reports an ablation on video frame interpolation.
Because GeoDiff is designed to operate with this adjustment enabled, we do not report a GeoDiff variant without it.
With the adjustment, all of SLERP, FIM-based metric, and our proposed metric improve in MSE and LPIPS, and our proposed method places first in all cases.
The gains are larger for our metric than for SLERP: the adjustment enables the guided diffusion model to better capture the local data manifold, and our metric explicitly leverages such refined local information.
By contrast, SLERP focuses on the Gaussian prior and is less sensitive to refinements.

\subsubsection{Geodesic Optimization Iterations.}
We compare the metric-based geodesic methods in terms of wall-clock time and iteration budget.
\Cref{tab:walltime_interp} reports the end-to-end wall time per image pair on MB(A), measured on a single NVIDIA RTX A6000 GPU under the same settings as \cref{sec:image_interp}.
Although all three methods have the same asymptotic order (\cref{tab:comp_cost}), ours has a higher wall-clock time because each iteration backpropagates through $s_\theta$ at the $N-1$ interpolants to capture local Jacobian structure.
In this setting, ours takes $1.8\times$ the wall time of GeoDiff, while achieving the best FID on all image interpolation datasets and the lowest MSE/LPIPS on all video interpolation datasets (\cref{tab:main_results,tab:video_results}).

\begin{figure}[p]
    \centering
    \includegraphics[scale=0.55]{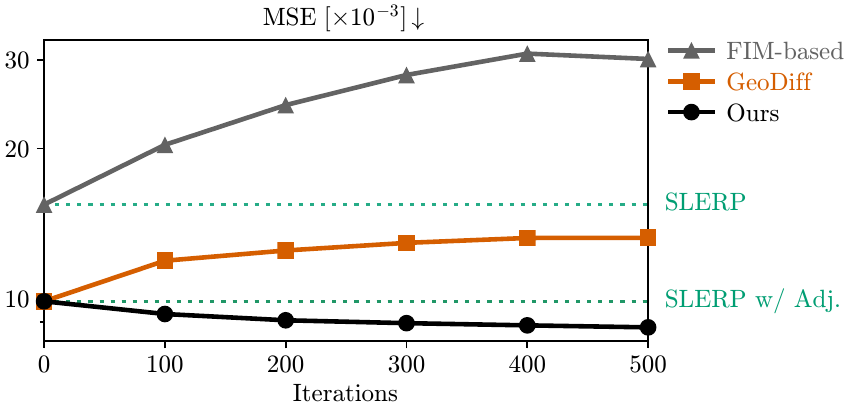}
    \caption{Ablation study on the number of geodesic optimization iterations for video frame interpolation on DAVIS}
    \label{fig:iteration_budget}
\end{figure}

\Cref{fig:iteration_budget} reports the effect of the iteration budget on video frame interpolation on DAVIS in 100-iteration increments.
Ours achieves the best MSE among these methods at 100 iterations, and improves further with more iterations.

\subsubsection{Noise Level $\tau$ and Spectral Gap.}
\label{appendix:ablation_tau}
We denote by $\tau$ the timestep at which we operate interpolations in the noise space and visualize results for varying $\tau$ in \cref{fig:ablation_tau_images}.
At $\tau=0$, from left to right, another face appears behind the main face and merges with the main face.
This is obviously awkward and undesirable.
With no injected noise, the data manifold is extremely thin, and finding a geodesic under our metric becomes ill-conditioned.
As $\tau$ increases, the interpolations become smoother and more globally coherent.
At $\tau=T$, however, the interpolations are no longer smooth: the noisy-sample distribution is close to an isotropic Gaussian, the data manifold is not well defined, and meaningful manifold-aware geodesics cannot be recovered.
Empirically, $\tau \in [0.4T,\,0.6T]$ yields the best visual quality.

\begin{figure}[p]
    \centering
    \includegraphics[scale=0.35]{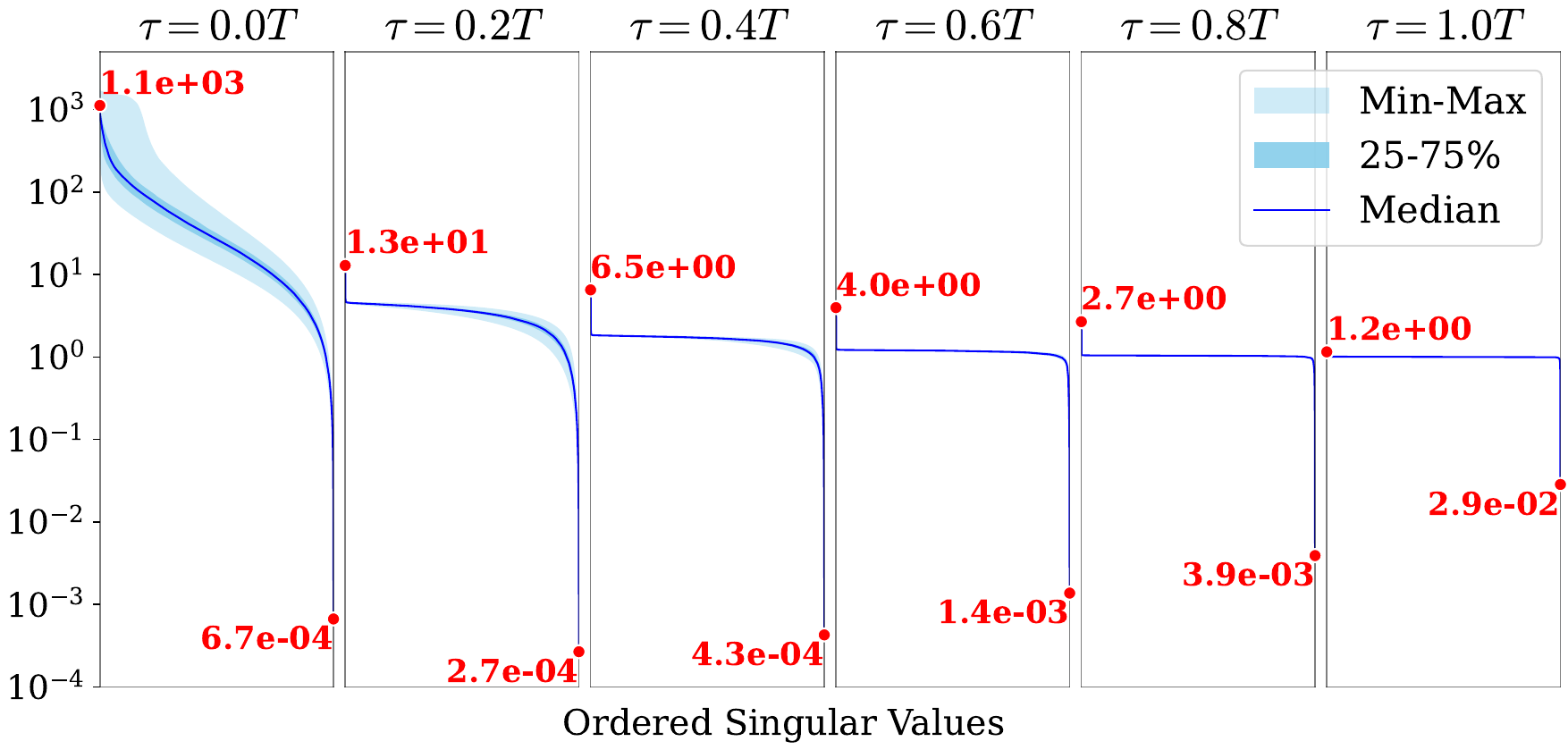}
    \caption{Distribution of singular values of the Jacobian $J_{x_\tau}$ of the score function $s_\theta$, aggregated over all images from the CelebA-HQ dataset with different $\tau$.
    The horizontal and vertical axes represent the index and the singular value (in log scale), respectively.
    The solid line, dark band, and light band indicate the median, 25--75\% range, and min--max range, respectively.}
    \label{fig:ablation_tau_spectrum}
\end{figure}

\Cref{fig:ablation_tau_spectrum} shows the distribution of singular values of the Jacobian $J_{x_\tau}$ of the score function $s_\theta$, aggregated over all images from the CelebA-HQ dataset used in our interpolation experiments, with the median, 25--75\% range, and min--max range shown.
Stable Diffusion v2.1-base \cite{Rombach2022} operates in a VAE latent space of $64\times 64\times 4=16{,}384$ dimensions.
At small $\tau$, hundreds of singular values are near zero, suggesting a local intrinsic dimensionality on the order of a few hundreds.
The spectral gap between the large and small singular values is the largest at small $\tau$ (e.g., $\tau=0.2T$), and it decreases as $\tau$ increases.
As $\tau\to T$, the noisy distribution approaches an isotropic Gaussian; in this ideal limit, the Jacobian is a scaled identity, so $G_{x_\tau}=J_{x_\tau}^{\top}J_{x_\tau}$ becomes a rescaled Euclidean metric and the tangent--normal spectral gap vanishes.
However, DDIM-inverted endpoints from real images remain input-dependent even near $\tau=T$, rather than being independent Gaussian-prior samples, which can leave a small residual spectral gap.
To balance the manifold's thickness and the spectral gap, moderate $\tau$ (e.g., $\tau=0.4T$ or $0.6T$) is preferred for interpolation.

\begin{figure}[p]
    \centering
    \includegraphics[scale=0.305]{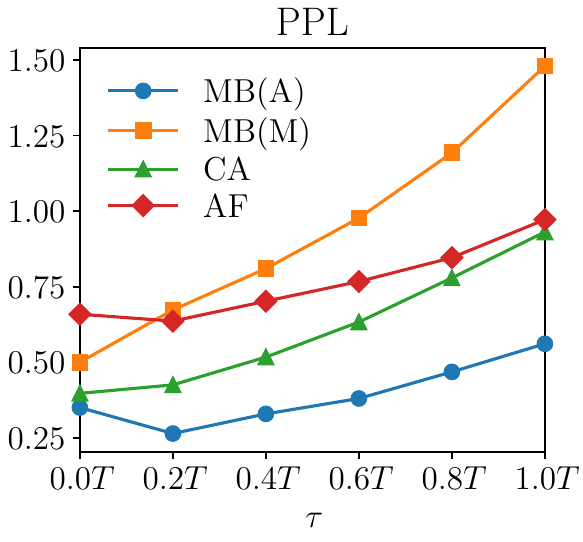}
    \includegraphics[scale=0.305]{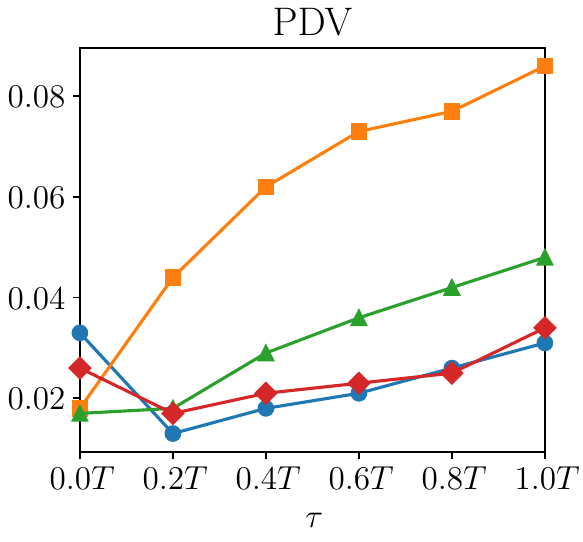}
    \includegraphics[scale=0.305]{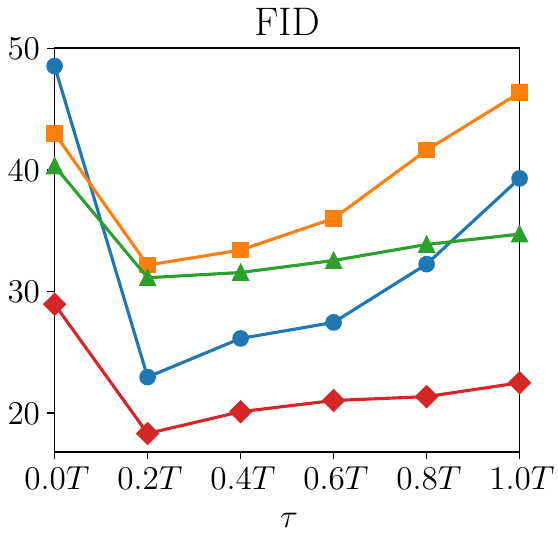}
    \includegraphics[scale=0.305]{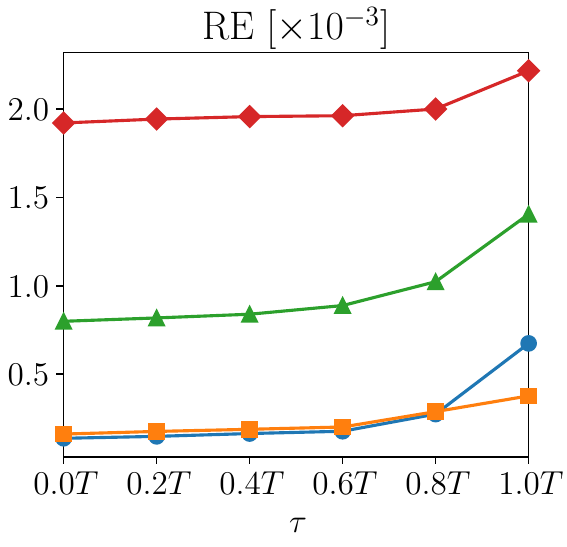}
    \caption{Ablation study on the choice of $\tau$ for image interpolation}
    \label{fig:ablation_tau_graph_image}
\end{figure}

\begin{figure}[p]
    \centering
    \includegraphics[scale=0.305]{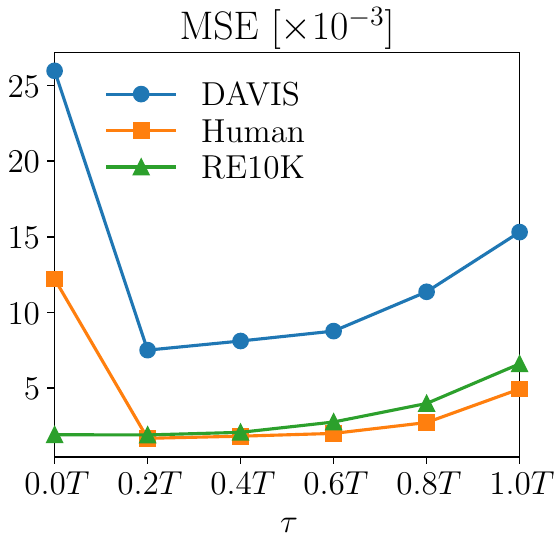}
    \includegraphics[scale=0.305]{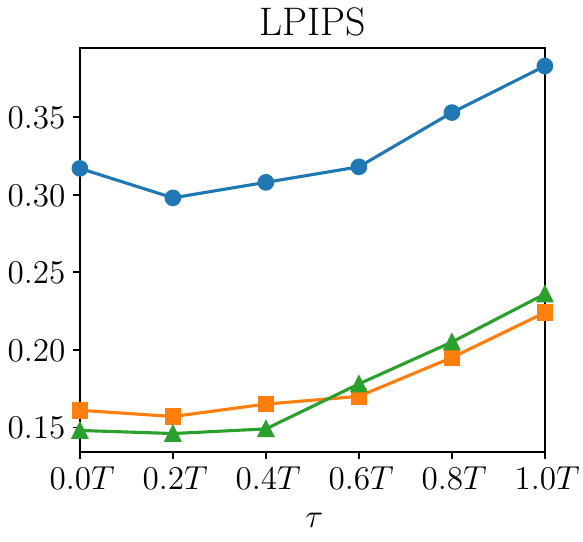}
    \caption{Ablation study on the choice of $\tau$ for video frame interpolation}
    \label{fig:ablation_tau_graph_video}
\end{figure}

We obtained the performance scores with varying $\tau$ and summarized results in \cref{fig:ablation_tau_graph_image} and \cref{fig:ablation_tau_graph_video}, where smaller $\tau$ values tend to yield better scores, provided that $\tau > 0$.
This quantitative trend does not align with the visual quality observed in \cref{fig:ablation_tau_images}.
A similar tendency was also reported by GeoDiff \cite{Yu2025}.
For small $\tau$, the interpolation behaves more like pixel-wise or patch-wise blending, which results in distorted images.
However, the errors measured by MSE and LPIPS are not particularly large.
This indicates the need for evaluation metrics that can properly measure image quality.
In any case, following prior work \cite{Yu2025}, we use $\tau = 0.6T$ in the image and video interpolation experiments for all applicable methods except NAO, which is designed to work at $\tau=1.0T$.
Using the same noise level for these methods ensures fairness in comparisons.

\subsubsection{Conjugate Gradient Weight $\lambda$ and Iterations.}
We ablate two hyperparameters of the conjugate gradient solver used in guidance correction (\cref{eq:corrected_guidance}): the regularization weight $\lambda$ and the number of CG iterations.
The evaluation follows \cref{sec:guidance}.
In each ablation, all other hyperparameters are held at their default values, and we fix the CFG scale to $w=7.5$.
\Cref{tab:ablation_cg_lambda,tab:ablation_cg_iters} show the results.

\Cref{tab:ablation_cg_lambda} shows that both FID and CLIP Score remain nearly constant as $\lambda$ varies from $10^{-2}$ to $10^{2}$.
This robustness can be understood from the spectral structure of $G_{x_t}=J_{x_t}^\top J_{x_t}$.
As discussed in Appendix~\ref{appendix:proposition_1}, the eigenvalues of $J_{x_t}$ exhibit a sharp spectral gap: eigenvalues associated with the normal space $\gN_{x_t}\gM_t$ are much larger than those associated with the tangent space $\gT_{x_t}\gM_t$.
Since our CG solver uses only a single update step (\cref{eq:corrected_guidance}), it does not converge to the exact solution of $(I+\lambda G_{x_t})\Delta\hat{s}^*=\Delta s$; instead, the single-step update suppresses the direction of the leading eigenvector of $G_{x_t}$, i.e., the normal-space direction, whose eigenvalue is dominant compared to $I$ because of the large spectral gap, making the correction robust to the choice of $\lambda$.

\Cref{tab:ablation_cg_iters} shows that increasing the number of CG iterations from one to two further lowers FID while nearly preserving CLIP Score.
The second iteration refines the correction along subdominant eigendirections of $G_{x_t}$ that the first step does not fully capture, further suppressing the residual normal component of $\Delta s$.
However, three or more iterations degrade performance.
Each CG iteration involves multiplication by $G_{x_t}$, which we approximate via finite differences and the assumption that $J_{x_t}$ is symmetric (\cref{sec:guidance}); these approximation errors accumulate over iterations, eventually outweighing the benefit of additional refinement.
Note that each additional CG iteration requires four extra evaluations of $s_\theta$, increasing the number of function evaluations (NFE) per denoising step.
Considering the trade-off between computational cost and performance, we adopt a single CG iteration as the default in all experiments.

\begin{table}[p]
    \centering
    \footnotesize
    \setlength{\tabcolsep}{4.5pt}
    \caption{Ablation study on prompt adjustment}
    \label{tab:ablation_prompt_adjustment}
    \begin{tabular}{l c rrr rrr}
        \toprule
               &      & \multicolumn{3}{c}{\textbf{MSE} $\downarrow$ $\scriptscriptstyle (\times 10^{-3})$} & \multicolumn{3}{c}{\textbf{LPIPS} $\downarrow$} \\
        \cmidrule(lr){3-5} \cmidrule(lr){6-8}
        \textbf{Method} & \textbf{Adj.} & \multicolumn{1}{c}{DAVIS} & \multicolumn{1}{c}{Human} & \multicolumn{1}{c}{RE10K} & \multicolumn{1}{c}{DAVIS} & \multicolumn{1}{c}{Human} & \multicolumn{1}{c}{RE10K} \\
        \midrule
        \textcolor{GREEN}{SLERP}       &            & 15.440             & 6.080              & 6.128              & 0.487              & 0.320              & 0.301              \\
        \textcolor{GREEN}{SLERP}       & \ding{51}  & 9.894              & 2.559              & 3.778              & 0.355              & 0.200              & 0.200              \\
        \textcolor{VERMILION}{GeoDiff} & \ding{51}  & 13.253             & 3.363              & 5.941              & \snd{0.334}        & \snd{0.184}        & 0.229              \\
        \textcolor{GRAY}{FIM-based}    &            & 30.172             & 11.638             & 12.679             & 0.535              & 0.388              & 0.373              \\
        \textcolor{GRAY}{FIM-based}    & \ding{51}  & \snd{9.757}        & \snd{2.506}        & \snd{3.001}        & 0.345              & 0.196              & \snd{0.194}        \\
        \midrule
        \textcolor{BLACK}{Ours}        &            & 13.517             & 5.008              & 6.016              & 0.500              & 0.350              & 0.325              \\
        \textcolor{BLACK}{Ours}        & \ding{51}  & \fst{8.777\sigtwo} & \fst{2.018\sigtwo} & \fst{2.771\sigtwo} & \fst{0.318\sigone} & \fst{0.170\sigtwo} & \fst{0.178\sigtwo} \\
        \bottomrule
    \end{tabular}\\
    \textsuperscript{\tiny\!*} and \textsuperscript{\tiny\!*\!*} indicate the statistical significance in the same manner as \cref{tab:image_interp_v2.0}.
\end{table}

\begin{table}[p]
    \centering
    \footnotesize
    \setlength{\tabcolsep}{4pt}
    \caption{Wall time per image interpolation pair on MB(A)}
    \label{tab:walltime_interp}
    \begin{tabular}{lc}
        \toprule
        \textbf{Method}                & \textbf{Time (s)} \\
        \midrule
        \textcolor{GRAY}{FIM-based}    & 178.29 \\
        \textcolor{VERMILION}{GeoDiff} & 216.79 \\
        \textcolor{BLACK}{Ours}        & 392.92 \\
        \bottomrule
    \end{tabular}
\end{table}

\clearpage

\begin{figure}[p]
    \centering
    \scriptsize
    \setlength{\tabcolsep}{1pt}
    \renewcommand{\arraystretch}{0}
    \begin{tabular}{rc}
        \raisebox{2.9mm}{$\tau=0.0T$} & {\includegraphics[width=10.4cm]{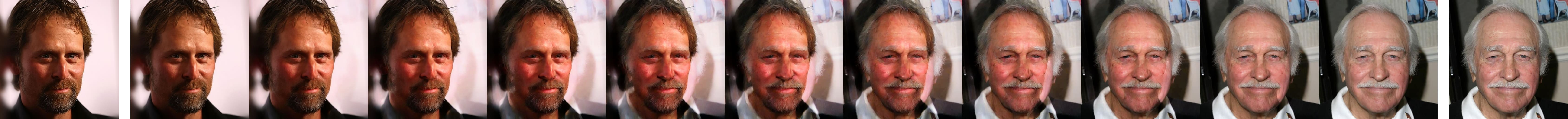}} \\ \addlinespace[1.5pt]
        \raisebox{2.9mm}{$\tau=0.2T$} & {\includegraphics[width=10.4cm]{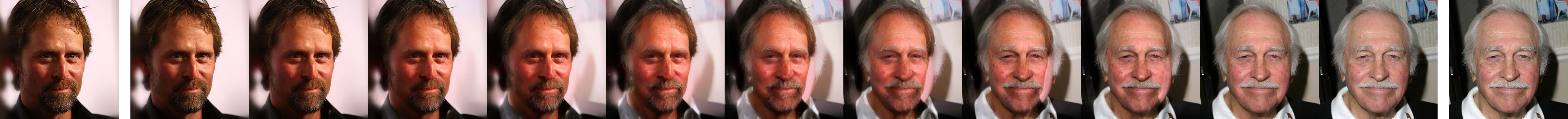}} \\ \addlinespace[1.5pt]
        \raisebox{2.9mm}{$\tau=0.4T$} & {\includegraphics[width=10.4cm]{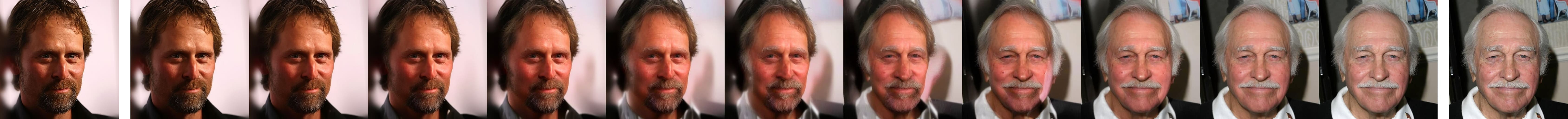}} \\ \addlinespace[1.5pt]
        \raisebox{2.9mm}{$\tau=0.6T$} & {\includegraphics[width=10.4cm]{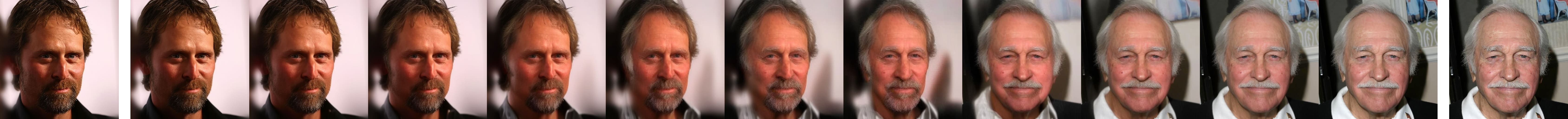}} \\ \addlinespace[1.5pt]
        \raisebox{2.9mm}{$\tau=0.8T$} & {\includegraphics[width=10.4cm]{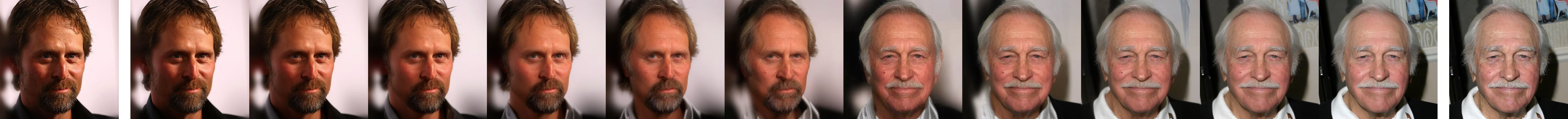}} \\ \addlinespace[1.5pt]
        \raisebox{2.9mm}{$\tau=1.0T$} & {\includegraphics[width=10.4cm]{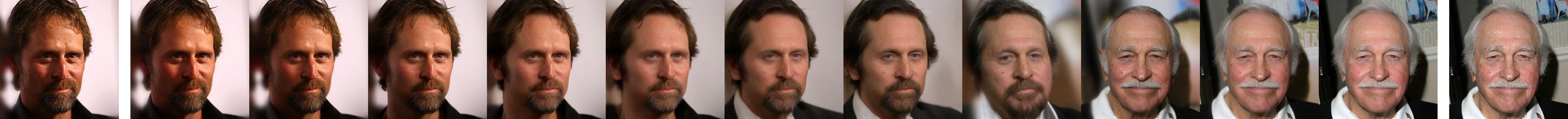}} \\ \addlinespace[1.5pt]
                                      & \includegraphics[scale=0.946]{assets/utils/imagelabel.pdf}                \\ \addlinespace[1.5pt]
    \end{tabular}
    \caption{Qualitative examples of interpolated image sequences with different $\tau$ using CelebA-HQ (Male).
        The leftmost and rightmost images are the given endpoints $x_0^\sss{0}$ and $x_0^\sss{1}$, and the middle images are the interpolated results $\{\hat{x}_0^\sss{u}\}$ for $u\in[0,1]$.}
    \label{fig:ablation_tau_images}
\end{figure}

\clearpage

\begin{table}[p]
    \centering
    \footnotesize
    \setlength{\tabcolsep}{4pt}
    \caption{Ablation study on the regularization weight $\lambda$ in guidance correction}
    \label{tab:ablation_cg_lambda}
    \begin{tabular}{c c c}
        \toprule
        $\lambda$ & FID $\downarrow$ & CLIP $\uparrow$ \\
        \midrule
        100   & 13.72 & 0.314 \\
        10    & 13.80 & 0.314 \\
        1     & 13.80 & 0.314 \\
        0.1   & 13.81 & 0.314 \\
        0.01  & 13.74 & 0.314 \\
        \bottomrule
    \end{tabular}
\end{table}

\begin{table}[p]
    \centering
    \footnotesize
    \setlength{\tabcolsep}{4pt}
    \caption{Ablation study on the number of CG iterations in guidance correction}
    \label{tab:ablation_cg_iters}
    \begin{tabular}{c c c c}
        \toprule
        Iterations & FID $\downarrow$ & CLIP $\uparrow$ & NFE $\downarrow$ \\
        \midrule
        0 & 14.29 & 0.314 & 2 \\
        \midrule
        1 & 13.81 & 0.314 & 6 \\
        2 & 13.12 & 0.313 & 10 \\
        3 & 23.03 & 0.308 & 14 \\
        \bottomrule
    \end{tabular}
\end{table}


\begin{thebibliography}{100}
\providecommand{\url}[1]{\texttt{#1}}
\providecommand{\urlprefix}{URL }
\providecommand{\doi}[1]{https://doi.org/#1}

\bibitem{Abdal2021}
Abdal, R., Zhu, P., Mitra, N.J., Wonka, P.: {StyleFlow: Attribute-Conditioned Exploration of StyleGAN-Generated Images Using Conditional Continuous Normalizing Flows}. ACM Transactions on Graphics  (2021)

\bibitem{Aoshima2023}
Aoshima, T., Matsubara, T.: {Deep Curvilinear Editing: Commutative and Nonlinear Image Manipulation for Pretrained Deep Generative Model}. In: IEEE/CVF Conference on Computer Vision and Pattern Recognition (CVPR) (2023)

\bibitem{Arjovsky2017}
Arjovsky, M., Bottou, L.: {Towards Principled Methods for Training Generative Adversarial Networks}. In: International Conference on Learning Representations (ICLR) (2017)

\bibitem{Arvanitidis2022}
Arvanitidis, G., Georgiev, B.M., Sch\"olkopf, B.: {A Prior-Based Approximate Latent Riemannian Metric}. In: International Conference on Artificial Intelligence and Statistics (AISTATS) (2022)

\bibitem{Arvanitidis2018}
Arvanitidis, G., Hansen, L.K., Hauberg, S.: {Latent Space Oddity: on the Curvature of Deep Generative Models}. In: International Conference on Learning Representations (ICLR) (2018)

\bibitem{Arvanitidis2021}
Arvanitidis, G., Hauberg, S., Sch{\"o}lkopf, B.: {Geometrically Enriched Latent Spaces}. In: International Conference on Artificial Intelligence and Statistics (AISTATS) (2021)

\bibitem{Azeglio2025}
Azeglio, S., Bernardo, A.D.: {What's Inside Your Diffusion Model? A Score-Based Riemannian Metric to Explore the Data Manifold}. arXiv  (2025)

\bibitem{Bengio2012}
Bengio, Y., Courville, A.C., Vincent, P.: {Representation Learning: A Review and New Perspectives}. IEEE Transactions on Pattern Analysis and Machine Intelligence  (2013)

\bibitem{Bethune2025}
B{\'e}thune, L., Vigouroux, D., Du, Y., VanRullen, R., Serre, T., Boutin, V.: {Follow the Energy, Find the Path: Riemannian Metrics from Energy-Based Models}. arXiv  (2025)

\bibitem{Bodin2025}
Bodin, E., Stere, A.I., Margineantu, D.D., Ek, C.H., Moss, H.: {Linear Combinations of Latents in Generative Models: Subspaces and Beyond}. In: International Conference on Learning Representations (ICLR) (2025)

\bibitem{Charpiat2019}
Charpiat, G., Girard, N., Felardos, L., Tarabalka, Y.: Input {{Similarity}} from the {{Neural Network Perspective}}. In: Advances in {{Neural Information Processing Systems}} ({{NeurIPS}}) (2019)

\bibitem{Chefer2023}
Chefer, H., Alaluf, Y., Vinker, Y., Wolf, L., {Cohen-Or}, D.: Attend-and-{{Excite}}: {{Attention-Based Semantic Guidance}} for {{Text-to-Image Diffusion Models}}. In: {{ACM Transactions on Graphics (TOG)}} (2023)

\bibitem{Chen2018}
Chen, N., Klushyn, A., Kurle, R., Jiang, X., Bayer, J., Smagt, P.: {Metrics for Deep Generative Models}. In: International Conference on Artificial Intelligence and Statistics (AISTATS) (2018)

\bibitem{Chen2022}
Chen, Z., Jiang, R., Duke, B., Zhao, H., Aarabi, P.: {{Exploring Gradient-Based Multi-directional Controls in GANs}}. In: European Conference on Computer Vision (ECCV) (2022)

\bibitem{Choi2022}
Choi, J., Lee, J., Yoon, C., Park, J.H., Hwang, G., Kang, M.: {Do Not Escape From the Manifold: Discovering the Local Coordinates on the Latent Space of {GAN}s}. In: International Conference on Learning Representations (ICLR) (2022)

\bibitem{Choi2022b}
Choi, Y., Uh, Y., Yoo, J., Ha, J.W.: {StarGAN v2: Diverse Image Synthesis for Multiple Domains}. In: IEEE Conference on Computer Vision and Pattern Recognition (CVPR) (2020)

\bibitem{Chung2025}
Chung, H., Kim, J., Park, G.Y., Nam, H., Ye, J.C.: {{CFG}}++: {{Manifold-constrained Classifier Free Guidance}} for {{Diffusion Models}}. In: International {{Conference}} on {{Learning Representations}} ({{ICLR}}) (2025)

\bibitem{Diepeveen2025}
Diepeveen, W., Batzolis, G., Shumaylov, Z., Sch{\"o}nlieb, C.B.: Score-based pullback riemannian geometry: Extracting the data manifold geometry using anisotropic flows. In: International Conference on Machine Learning (ICML) (2025)

\bibitem{Elfwing2017}
Elfwing, S., Uchibe, E., Doya, K.: {Sigmoid-Weighted Linear Units for Neural Network Function Approximation in Reinforcement Learning}. arXiv  (2017)

\bibitem{Fefferman2016}
Fefferman, C., Mitter, S.K., Narayanan, H.: Testing the manifold hypothesis. Journal of the American Mathematical Society  (2016)

\bibitem{Gal2023}
Gal, R., Alaluf, Y., Atzmon, Y., Patashnik, O., Bermano, A.H., Chechik, G., Cohen-Or, D.: {An Image is Worth One Word: Personalizing Text-to-Image Generation using Textual Inversion}. In: International Conference on Learning Representations (ICLR) (2023)

\bibitem{George2025}
George, A.J., Veiga, R., Macris, N.: {Analysis of Diffusion Models for Manifold Data}. arXiv  (2025)

\bibitem{Goetschalckx2019}
Goetschalckx, L., Andonian, A., Oliva, A., Isola, P.: {GANalyze: Toward Visual Definitions of Cognitive Image Properties}. In: IEEE/CVF International Conference on Computer Vision (ICCV) (2019)

\bibitem{Goodfellow2014}
Goodfellow, I., Pouget-Abadie, J., Mirza, M., Xu, B., Warde-Farley, D., Ozair, S., Courville, A., Bengio, Y.: {Generative Adversarial Nets}. In: Advances in Neural Information Processing Systems (NeurIPS) (2014)

\bibitem{Gruffaz2025}
Gruffaz, S., Sassen, J.: {Riemannian Metric Learning: Closer to You than You Imagine}. arXiv  (2025)

\bibitem{Guo2024}
Guo, J., Xu, X., Pu, Y., Ni, Z., Wang, C., Vasu, M., Song, S., Huang, G., Shi, H.: {Smooth Diffusion: Crafting Smooth Latent Spaces in Diffusion Models}. In: IEEE/CVF Conference on Computer Vision and Pattern Recognition (CVPR) (2024)

\bibitem{Haas2022}
Haas, R., Gra{\ss}hof, S., Brandt, S.S.: {Tensor-based Emotion Editing in the StyleGAN Latent Space}. In: CVPR 2022 Workshop on AI for Content Creation Workshop (2022)

\bibitem{Hahm2024}
Hahm, J., Lee, J., Kim, S., Lee, J.: {Isometric representation learning for disentangled latent space of diffusion models}. In: International Conference on Machine Learning (ICML) (2024)

\bibitem{Hanawa2021}
Hanawa, K., Yokoi, S., Hara, S., Inui, K.: Evaluation of {{Similarity-based Explanations}}. In: {{International Conference}} on {{Learning Representations}} ({{ICLR}}) (2021)

\bibitem{Harkonen2020}
H{\"a}rk{\"o}nen, E., Hertzmann, A., Lehtinen, J., Paris, S.: {GANSpace: Discovering Interpretable GAN Controls}. In: Advances in Neural Information Processing Systems (NeurIPS) (2020)

\bibitem{He2024}
He, Q., Wang, J., Liu, Z., Yao, A.: {AID: Attention Interpolation of Text-to-Image Diffusion}. In: Advances in Neural Information Processing Systems (NeurIPS) (2024)

\bibitem{Martin2017}
Heusel, M., Ramsauer, H., Unterthiner, T., Nessler, B., Hochreiter, S.: {GANs trained by a two time-scale update rule converge to a local Nash equilibrium}. In: Advances in Neural Information Processing Systems (NeurIPS) (2017)

\bibitem{Ho2020}
Ho, J., Jain, A., Abbeel, P.: {Denoising Diffusion Probabilistic Models}. In: Advances in Neural Information Processing Systems (NeurIPS) (2020)

\bibitem{Ho2021}
Ho, J., Salimans, T.: {Classifier-Free Diffusion Guidance}. In: NeurIPS 2021 Workshop on Deep Generative Models and Downstream Applications (2021)

\bibitem{Horvat2024}
Horvat, C., Pfister, J.P.: {On Gauge Freedom, Conservativity and Intrinsic Dimensionality Estimation in Diffusion models}. In: International Conference on Learning Representations (ICLR) (2024)

\bibitem{Humayun2025}
Humayun, A.I., Amara, I., Vasconcelos, C.N., Ramachandran, D., Schumann, C., He, J., Heller, K.A., Farnadi, G., Rostamzadeh, N., Havaei, M.: {What Secrets Do Your Manifolds Hold? Understanding the Local Geometry of Generative Models}. In: International Conference on Learning Representations (ICLR) (2025)

\bibitem{Jahanian2020}
Jahanian, A., Chai, L., Isola, P.: {On the "steerability" of generative adversarial networks}. In: International Conference on Learning Representations, {(ICLR)} (2020)

\bibitem{Kamkari2024}
Kamkari, H., Ross, B.L., Hosseinzadeh, R., Cresswell, J.C., Loaiza-Ganem, G.: {A Geometric View of Data Complexity: Efficient Local Intrinsic Dimension Estimation with Diffusion Models}. In: Advances in Neural Information Processing Systems (NeurIPS) (2024)

\bibitem{Karczewski2025a}
Karczewski, R., Heinonen, M., Garg, V.K.: {Devil is in the Details: Density Guidance for Detail-Aware Generation with Flow Models}. In: International Conference on Machine Learning (ICML) (2025)

\bibitem{Karczewski2025b}
Karczewski, R., Heinonen, M., Pouplin, A., Hauberg, S., Garg, V.: {The Spacetime of Diffusion Models: An Information Geometry Perspective}. In: International Conference on Learning Representations (ICLR) (2026)

\bibitem{Karras2018}
Karras, T., Aila, T., Laine, S., Lehtinen, J.: {Progressive Growing of GANs for Improved Quality, Stability, and Variation}. In: International Conference on Learning Representations (ICLR) (2018)

\bibitem{Karras2018b}
Karras, T., Laine, S., Aila, T.: A style-based generator architecture for generative adversarial networks. In: IEEE/CVF Conference on Computer Vision and Pattern Recognition (CVPR) (2019)

\bibitem{Khrulkov2021}
Khrulkov, V., Mirvakhabova, L., Oseledets, I., Babenko, A.: {Latent Transformations via NeuralODEs for GAN-based Image Editing}. In: IEEE/CVF International Conference on Computer Vision (ICCV) (2021)

\bibitem{Kim2025}
Kim, Y., Lee, K., Park, M., Na, B., chul Moon, I.: {Diffusion Bridge AutoEncoders for Unsupervised Representation Learning}. In: International Conference on Learning Representations (ICLR) (2025)

\bibitem{Kingma2015}
Kingma, D.P., Ba, J.: {Adam: A Method for Stochastic Optimization}. In: {International Conference on Learning Representations (ICLR)} (2015)

\bibitem{Kingma2014}
Kingma, D.P., Welling, M.: {Auto-Encoding Variational Bayes}. In: International Conference on Learning Representations (ICLR) (2014)

\bibitem{Kwon2023}
Kwon, M., Jeong, J., Uh, Y.: {Diffusion Models Already Have A Semantic Latent Space}. In: International Conference on Learning Representations (ICLR) (2023)

\bibitem{Kwon2025}
Kwon, M., seong Kim, S., Jeong, J., Hsiao, Y.T., Uh, Y.: {{TCFG}}: {{Tangential Damping Classifier-free Guidance}}. In: {{IEEE}}/{{CVF Conference}} on {{Computer Vision}} and {{Pattern Recognition}} ({{CVPR}}) (2025)

\bibitem{Lee2019}
Lee, J.M.: {Introduction to Riemannian Manifolds}. Springer (2019)

\bibitem{Lee2022}
Lee, Y., Kim, S., Choi, J., Park, F.: {A Statistical Manifold Framework for Point Cloud Data}. In: International Conference on Machine Learning (ICML) (2022)

\bibitem{Li2023}
Li, J., Li, D., Savarese, S., Hoi, S.: {BLIP-2: bootstrapping language-image pre-training with frozen image encoders and large language models}. In: International Conference on Machine Learning (ICML) (2023)

\bibitem{Liang2021}
Liang, H., Hou, X., Shen, L.: {SSFlow: Style-guided Neural Spline Flows for Face Image Manipulation}. In: ACM International Conference on Multimedia (2021)

\bibitem{Lin2014}
Lin, T.Y., Maire, M., Belongie, S.J., Hays, J., Perona, P., Ramanan, D., Doll{\'a}r, P., Zitnick, C.L.: Microsoft {{COCO}}: {{Common Objects}} in {{Context}}. In: European Conference on Computer Vision (ECCV) (2014)

\bibitem{Loaiza-ganem2024}
Loaiza-Ganem, G., Ross, B.L., Hosseinzadeh, R., Caterini, A.L., Cresswell, J.C.: {Deep Generative Models through the Lens of the Manifold Hypothesis: A Survey and New Connections}. Transactions on Machine Learning Research  (2024)

\bibitem{Lobashev2025}
Lobashev, A., Guskov, D., Larchenko, M., Tamm, M.: {Hessian Geometry of Latent Space in Generative Models}. In: International Conference on Machine Learning (ICML) (2025)

\bibitem{Loshchilov2017}
Loshchilov, I., Hutter, F.: {SGDR: Stochastic Gradient Descent with Warm Restarts}. In: International Conference on Learning Representations (ICLR) (2017)

\bibitem{Loshchilov2019}
Loshchilov, I., Hutter, F.: {Decoupled Weight Decay Regularization}. In: International Conference on Learning Representations (ICLR) (2019)

\bibitem{Lu2024}
Lu, Z., Wu, C., Chen, X., Wang, Y., Bai, L., Qiao, Y., Liu, X.: {Hierarchical Diffusion Autoencoders and Disentangled Image Manipulation}. In: IEEE/CVF Winter Conference on Applications of Computer Vision (WACV) (2024)

\bibitem{Luo2023}
Luo, S., Tan, Y., Huang, L., Li, J., Zhao, H.: Latent consistency models: Synthesizing high-resolution images with few-step inference. arXiv  (2023)

\bibitem{Mokady2023}
Mokady, R., Hertz, A., Aberman, K., Pritch, Y., Cohen-Or, D.: {Null-text Inversion for Editing Real Images using Guided Diffusion Models}. In: IEEE/CVF Conference on Computer Vision and Pattern Recognition, CVPR 2023 (2023)

\bibitem{Oldfield2021}
Oldfield, J., Georgopoulos, M., Panagakis, Y., Nicolaou, M.A., Patras, I.: {Tensor Component Analysis for Interpreting the Latent Space of GANs}. In: British Machine Vision Conference (BMVC) (2021)

\bibitem{Park2023a}
Park, Y.H., Kwon, M., Choi, J., Jo, J., Uh, Y.: {Understanding the Latent Space of Diffusion Models through the Lens of Riemannian Geometry}. In: Advances in Neural Information Processing Systems (NeurIPS) (2023)

\bibitem{Park2023b}
Park, Y.H., Kwon, M., Jo, J., Uh, Y.: {Unsupervised Discovery of Semantic Latent Directions in Diffusion Models}. arXiv  (2023)

\bibitem{Perazzi2016}
Perazzi, F., Pont-Tuset, J., McWilliams, B., Van~Gool, L., Gross, M., Sorkine-Hornung, A.: A benchmark dataset and evaluation methodology for video object segmentation. In: IEEE/CVF Conference on Computer Vision and Pattern Recognition (CVPR) (2016)

\bibitem{Pidstrigach2022}
Pidstrigach, J.: {Score-Based Generative Models Detect Manifolds}. In: Advances in Neural Information Processing Systems (NeurIPS) (2022)

\bibitem{von-platen-etal-2022-diffusers}
von Platen, P., Patil, S., Lozhkov, A., Cuenca, P., Lambert, N., Rasul, K., Davaadorj, M., Nair, D., Paul, S., Berman, W., Xu, Y., Liu, S., Wolf, T.: Diffusers: State-of-the-art diffusion models. \url{https://github.com/huggingface/diffusers} (2022)

\bibitem{Plumerault2020}
Plumerault, A., Borgne, H.L., Hudelot, C.: {Controlling generative models with continuous factors of variations}. In: International Conference on Learning Representations (ICLR) (2020)

\bibitem{Potaptchik2025}
Potaptchik, P., Azangulov, I., Deligiannidis, G.: {Linear Convergence of Diffusion Models Under the Manifold Hypothesis}. arXiv  (2025)

\bibitem{Konpat2022}
Preechakul, K., Chatthee, N., Wizadwongsa, S., Suwajanakorn, S.: {Diffusion Autoencoders: Toward a Meaningful and Decodable Representation}. In: IEEE/CVF Conference on Computer Vision and Pattern Recognition (CVPR) (2022)

\bibitem{Radford2021}
Radford, A., Kim, J.W., Hallacy, C., Ramesh, A., Goh, G., Agarwal, S., Sastry, G., Askell, A., Mishkin, P., Clark, J., et~al.: {Learning transferable visual models from natural language supervision}. In: International Conference on Machine Learning (ICML) (2021)

\bibitem{Ramesh2019}
Ramesh, A., Choi, Y., LeCun, Y.: {A Spectral Regularizer for Unsupervised Disentanglement}. arXiv  (2019)

\bibitem{Rassin2023}
Rassin, R., Hirsch, E., Glickman, D., Ravfogel, S., Goldberg, Y., Chechik, G.: Linguistic {{Binding}} in {{Diffusion Models}}: {{Enhancing Attribute Correspondence}} through {{Attention Map Alignment}}. In: Advances in Neural Information Processing Systems (NeurIPS) (2023)

\bibitem{Rombach2022}
Rombach, R., Blattmann, A., Lorenz, D., Esser, P., Ommer, B.: {High-resolution Image Synthesis with Latent Diffusion Models}. In: IEEE/CVF Conference on Computer Vision and Pattern Recognition (CVPR) (2022)

\bibitem{Samuel2023}
Samuel, D., Ben-Ari, R., Darshan, N., Maron, H., Chechik, G.: {Norm-guided Latent Space Exploration for Text-to-image Generation}. In: Advances in Neural Information Processing Systems (NeurIPS) (2023)

\bibitem{Shao2017}
Shao, H., Kumar, A., Fletcher, P.T.: {The Riemannian Geometry of Deep Generative Models}. In: IEEE/CVF Conference on Computer Vision and Pattern Recognition Workshops (CVPRW) (2018)

\bibitem{Shen2024}
Shen, L., Liu, T., Sun, H., Ye, X., Li, B., Zhang, J., Cao, Z.: {DreamMover: Leveraging the Prior of Diffusion Models for Image Interpolation with Large Motion}. In: European Conference on Computer Vision (ECCV) (2024)

\bibitem{Shen2020}
Shen, Y., Gu, J., Tang, X., Zhou, B.: {Interpreting the Latent Space of GANs for Semantic Face Editing}. In: IEEE/CVF Conference on Computer Vision and Pattern Recognition (CVPR) (2020)

\bibitem{Shen2021}
Shen, Y., Zhou, B.: {Closed-form Factorization of Latent Semantics in GANs}. In: IEEE/CVF Conference on Computer Vision and Pattern Recognition (CVPR) (2021)

\bibitem{Shoemake1985}
Shoemake, K.: {Animating Rotation with Quaternion Curves}. International Conference on Computer Graphics and Interactive Techniques  (1985)

\bibitem{Sohl-Dickstein2015}
Sohl-Dickstein, J., Weiss, E., Maheswaranathan, N., Ganguli, S.: {Deep Unsupervised Learning using Nonequilibrium Thermodynamics}. In: International Conference on Machine Learning (ICML) (2015)

\bibitem{Song2021a}
Song, J., Meng, C., Ermon, S.: {Denoising Diffusion Implicit Models}. In: International Conference on Learning Representations (ICLR) (2021)

\bibitem{Song2021b}
Song, Y., Sohl-Dickstein, J., Kingma, D.P., Kumar, A., Ermon, S., Poole, B.: {Score-Based Generative Modeling through Stochastic Differential Equations}. In: International Conference on Learning Representations (ICLR) (2021)

\bibitem{Sorrenson2025}
Sorrenson, P., Behrend-Uriarte, D., Schn{\"o}rr, C., K{\"o}the, U.: Learning distances from data with normalizing flows and score matching. In: International Conference on Machine Learning (ICML) (2025)

\bibitem{Spingarn2021}
Spingarn, N., Banner, R., Michaeli, T.: {GAN} ''steerability'' without optimization. In: International Conference on Learning Representations (ICLR) (2021)

\bibitem{Stanczuk2024}
Stanczuk, J.P., Batzolis, G., Deveney, T., Sch{\"o}nlieb, C.B.: {Diffusion Models Encode the Intrinsic Dimension of Data Manifolds}. In: International Conference on Machine Learning (ICML) (2024)

\bibitem{Sueyoshi2024}
Sueyoshi, K., Matsubara, T.: Predicated {{Diffusion}}: {{Predicate Logic-Based Attention Guidance}} for {{Text-to-Image Diffusion Models}}. In: {{IEEE}}/{{CVF Conference}} on {{Computer Vision}} and {{Pattern Recognition}} ({{CVPR}}) (2024)

\bibitem{Szegedy2015}
Szegedy, C., Vanhoucke, V., Ioffe, S., Shlens, J., Wojna, Z.: {Rethinking the Inception Architecture for Computer Vision}. In: IEEE/CVF Conference on Computer Vision and Pattern Recognition (CVPR) (2016)

\bibitem{Tang2024}
Tang, R., Yang, Y.: {Adaptivity of Diffusion Models to Manifold Structures}. In: International Conference on Artificial Intelligence and Statistics (AISTATS) (2024)

\bibitem{Tewari2020}
Tewari, A.K., Elgharib, M.A., Bharaj, G., Bernard, F., Seidel, H.P., P{\'e}rez, P., Zollh{\"o}fer, M., Theobalt, C.: {StyleRig: Rigging StyleGAN for 3D Control Over Portrait Images}. In: IEEE/CVF Conference on Computer Vision and Pattern Recognition (CVPR) (2020)

\bibitem{Tzelepis2021}
Tzelepis, C., Tzimiropoulos, G., Patras, I.: {WarpedGANSpace: Finding non-linear RBF paths in GAN latent space}. In: IEEE/CVF International Conference on Computer Vision (ICCV) (2021)

\bibitem{Ventura2025}
Ventura, E., Achilli, B., Silvestri, G., Lucibello, C., Ambrogioni, L.: {Manifolds, Random Matrices and Spectral Gaps: The Geometric Phases of Generative Diffusion}. In: International Conference on Learning Representations (ICLR) (2025)

\bibitem{Voynov2020}
Voynov, A., Babenko, A.: {Unsupervised discovery of interpretable directions in the GAN latent space}. In: International Conference on Machine Learning (ICML) (2020)

\bibitem{Wang2023}
Wang, C.J., Golland, P.: {Interpolating between Images with Diffusion Models}. In: ICML 2023 Workshop on Challenges of Deploying Generative AI (2023)

\bibitem{Wenliang2022}
Wenliang, L.K., Moran, B.: {Score-based generative models learn manifold-like structures with constrained mixing}. In: NeurIPS 2022 Workshop on Score-Based Methods (2022)

\bibitem{Yang2018}
Yang, T., Arvanitidis, G., Fu, D., Li, X., Hauberg, S.: {Geodesic Clustering in Deep Generative Models}. arXiv  (2018)

\bibitem{Yang2024}
Yang, Z., Yu, Z., Xu, Z., Singh, J., Zhang, J., Campbell, D., Tu, P., Hartley, R.: {IMPUS: Image Morphing with Perceptually-Uniform Sampling Using Diffusion Models}. In: International Conference on Learning Representations (ICLR) (2024)

\bibitem{Yeh2018}
Yeh, C.K., Kim, J.S., Yen, I.E.H., Ravikumar, P.: Representer {{Point Selection}} for {{Explaining Deep Neural Networks}}. In: Advances in {{Neural Information Processing Systems}} ({{NeurIPS}}) (2018)

\bibitem{Yu2025}
Yu, Q., Singh, J., Yang, Z., Tu, P.H., Zhang, J., Li, H., Hartley, R., Campbell, D.: {Probability Density Geodesics in Image Diffusion Latent Space}. In: IEEE/CVF Conference on Computer Vision and Pattern Recognition (CVPR) (2025)

\bibitem{Yun2024}
Yun, Z., Chuang, G., Dong, D., Chen, Y.: {Denoising} for {Manifold Extrapolation}. In: NeurIPS 2024 Workshop on Scientific Methods for Understanding Deep Learning (SciForDL) (2024)

\bibitem{Kaiwen2023}
Zhang, K., Zhou, Y., Xu, X., Dai, B., Pan, X.: {DiffMorpher: Unleashing the Capability of Diffusion Models for Image Morphing}. In: IEEE/CVF Conference on Computer Vision and Pattern Recognition (CVPR) (2024)

\bibitem{Zhang2018}
Zhang, R., Isola, P., Efros, A.A., Shechtman, E., Wang, O.: {The Unreasonable Effectiveness of Deep Features as a Perceptual Metric}. In: IEEE/CVF Conference on Computer Vision and Pattern Recognition (CVPR) (2018)

\bibitem{Zheng2024}
Zheng, P., Zhang, Y., Fang, Z., Liu, T., Lian, D., Han, B.: {NoiseDiffusion: Correcting Noise for Image Interpolation with Diffusion Models beyond Spherical Linear Interpolation}. In: International Conference on Learning Representations (ICLR) (2024)

\bibitem{Zhou2018}
Zhou, T., Tucker, R., Flynn, J., Fyffe, G., Snavely, N.: Stereo magnification: Learning view synthesis using multiplane images. ACM Transactions on Graphics (TOG)  (2018)

\bibitem{Zhu2024}
Zhu, T., Ren, D., Wang, Q., Wu, X., Zuo, W.: Generative inbetweening through frame-wise conditions-driven video generation. arXiv  (2024)

\bibitem{Zhuang2021}
Zhuang, P., Koyejo, O.O., Schwing, A.: {Enjoy Your Editing: Controllable GANs} for image editing via latent space navigation. In: International Conference on Learning Representations (ICLR) (2021)

\end{thebibliography}
\end{document}